%% file: attendlight.tex
\documentclass{article}

\usepackage[final, nonatbib]{neurips_2020}

\usepackage[utf8]{inputenc} 
\usepackage[T1]{fontenc}    
\usepackage{hyperref}       
\usepackage{url}            
\usepackage{booktabs}       
\usepackage{amsfonts}       
\usepackage{nicefrac}       
\usepackage{microtype}      
\usepackage{todonotes}
\usepackage{multirow}

\usepackage{tikz}
\usepackage{xcolor}
\usetikzlibrary{arrows,positioning,backgrounds,decorations.markings}
\usepackage{subcaption}

\usepackage{amsmath}
\usepackage{enumerate,verbatim}
\usepackage{hyperref}       
\usepackage{url}            
\usepackage{multirow} 
\usepackage{wrapfig}
\usepackage{comment}
\usepackage{soul} 
\usepackage{adjustbox}
\usepackage{float}
\usepackage{mathtools}
\usepackage{keyval}
\usepackage{ifthen}

\usepackage{psfrag,cite,amsmath}

\usepackage{algorithmic}
\usepackage[ruled,vlined]{algorithm2e}

\usepackage{pdfpages}

\newcommand{\adddavood}[1]{{\color{orange} {#1}}}

\title{AttendLight: Universal Attention-Based Reinforcement Learning Model for Traffic Signal Control}

\author{%
  Afshin Oroojlooy \\
   SAS Institute Inc.\\
   Cary, NC 27513 \\
  \texttt{afshin.oroojlooy@sas.com} \\
   \And
   Mohammadreza Nazari \\
   SAS Institute Inc.\\
   Cary, NC 27513 \\
   \texttt{reza.nazari@sas.com} \\
   \AND
   Davood Hajinezhad \\
   SAS Institute Inc.\\
   Cary, NC 27513 \\
   \texttt{davood.hajinezhad@sas.com} \\
   \And
   Jorge Silva \\
   SAS Institute Inc.\\
   Cary, NC 27513 \\
   \texttt{jorge.silva@sas.com} \\
}

\begin{document}

\newpage
\maketitle

\begin{abstract}
We propose AttendLight, an end-to-end Reinforcement Learning (RL) algorithm for the problem of traffic signal control. Previous approaches for this problem have the shortcoming that they require training for each new intersection with a different structure or traffic flow distribution. AttendLight solves this issue by training a single, universal model for intersections with any number of roads, lanes, phases (possible signals), and traffic flow. To this end, we propose a deep RL model which incorporates two attention models. The first attention model is introduced to  handle different numbers of roads-lanes; and the second attention model is intended for enabling decision-making with any number of phases in an intersection. As a result, our
proposed model works for any intersection configuration, as long as a similar configuration is represented in the training set. 
Experiments were conducted with both synthetic and real-world standard benchmark data-sets. The results we show cover intersections with three or four approaching roads; one-directional/bi-directional roads with one, two, and three lanes; different number of phases; and different traffic flows. We consider two regimes: (i) single-environment training, single-deployment, and (ii) multi-environment training, multi-deployment. AttendLight outperforms both classical and other RL-based approaches on all cases in both regimes.
\end{abstract}

\section{Introduction}
  With the emergence of urbanization and the increase in household car ownership, traffic congestion has been one of the major challenges in many highly-populated cities. Traffic congestion can be mitigated by road expansion/correction, sophisticated road allowance rules, or improved traffic signal controlling. Although either of these solutions could decrease the travel times and fuel costs, optimizing the traffic signals is more convenient due to the limited funding resources and opportunity of finding more effective strategies. This paper introduces a framework for learning a general traffic control policy that can be deployed in an intersection of interest and ease the traffic flow.
  
  Approaches for controlling signalized intersections could be categorized into two main classes, namely conventional methods and adaptive methods. In the former, customarily rule-based fixed cycles and phase times are determined a priori and offline based on historical measurements as well as some assumptions about the underlying problem structure. However, since traffic behavior is dynamically changing, that makes most conventional methods highly inefficient. In adaptive methods, decisions are made based on the current state of the intersection. Self-organizing Traffic Light Control (SOTL) \cite{gershenson2004self} and {\it Max-pressure} \cite{varaiya2013max} are among the most popular adaptive methods that consider the number of approaching vehicles to the intersection in their traffic control algorithm (See, e.g., \cite{koonce2008traffic} for more details). These methods bring considerable improvements in traffic signal control; nonetheless, they are short-sighted and do not consider the long-term effects of the decisions on the traffic. Besides, these methods do not use the feedback from previous actions toward making more efficient decisions. In response, more sophisticated algorithms have been proposed. Using artificial intelligence (AI) for controlling traffic signals has recently attracted a lot of attention, due to the major potential benefits that it can bring toward having less-congested cities. Reinforcement Learning (RL) \cite{sutton2018reinforcement}, which has been flourished in recent years in the AI community, has shown superior performance for a wide range of problems such as games \cite{silver2016mastering}, robotics \cite{gu2017deep}, finance \cite {fischer2018reinforcement}, and operations research \cite{bengio2018machine}, only to name a few. This coincides with growing applications of RL in traffic signal control problem (TSCP) \cite{van2016coordinated,li2016traffic,zheng2019learning,liang2018deep,velivckovic2017graph}.

  In spite of immense improvements achieved by RL methods for a broad domain of intersections, the main limitation for the majority of the methods is that the proposed model needs to be re-designed and re-trained from scratch whenever it faces a different intersection either with different topology or traffic distribution. Learning specialized policies for each individual intersection can be problematic, as not only do RL agents have to store a distinct policy for each intersection but in practice data collection resources and preparation impose costs. These costs include the burden on human-experts' time to setup a new model, and computational resources to train and tune a new model. Thus, it is not clear whether such a cumbersome procedure is feasible for a city with thousands of distinct intersections. There exists some prior work on partially alleviating such issues  \cite{zheng2019learning, wei2019deep} using transfer learning; however, the trained models still need to be manipulated for different intersection structures and require retraining to achieve reasonable performance. 
  

  To address these issues, we bring ideas from attentional models \cite{bahdanau2014neural} into TSCP. Rather than specializing on a single intersection, our goal is to design a mechanism with satisfactory performance across a group of intersections. Attentional mechanisms are a natural choice, since they allow unified system representations by handling variable-length inputs. We propose the AttendLight framework, a reinforcement learning algorithm, to train a ``universal'' model which can be used for any intersection, with any number of roads, lanes, phase, traffic distribution, and type of sensory data to measure the traffic. In other words, once the model is trained under a comprehensive set of phases, roads, lanes, and traffic distribution, our trained model can be used for new unseen intersection, and it provides reasonable performance as long as the intersection configuration follows a pattern present in the training set. We find that AttendLight architecture can extract an abstract representation of the intersection status, without any extra grounding or redefinition, and reuse this information for fast deployment. We show that our approach substantially outperforms purely conventional controls and FRAP \cite{zheng2019learning}, one of the state-of-the-art RL-based methods.

\section{Related Work}
  The selection of RL components in traffic light control is quite challenging. The most common action set for the traffic problem is the set of all possible phases. In \cite{van2016coordinated} an image-like representation is used as the state and a combination of vehicle delay and waiting time is considered as the reward. A deep Q-Network algorithm was proposed in \cite{li2016traffic}, where the queue length of the last four samples is defined as the state, and reward is defined as the absolute value of the difference between queue length of approaching lanes. In \cite{liang2018deep} the intersection was divided into multiple chunks building a matrix such that each chunk contains a binary indicator for existence of a car and its speed. Using this matrix as the state and defining reward as the difference of the cumulative waiting time for two cycles, they proposed to learn the duration of each phase in a fixed cycle by a Double Dueling DQN algorithm with a prioritized experience replay.
  Likewise, \cite{shabestary2018deep} defined a similar state by dividing each lane into a few chucks and the reward is the reduction of cumulative delay in the intersection. A DQN approach was proposed to train a policy to choose the next phase. Ault et al. \cite{ault2019learning} proposed three DQN-based algorithms to obtain an interpretable policy. A simple function approximator with 256 variables was used and is showed that it obtains slightly worse result compared to the DQN algorithm with an uninterpretable approximator. 
  The IntelliLight algorithm was proposed in \cite{wei2018intellilight}. The state and reward are a combination of several components. 
 A multi-intersection problem was considered in \cite{wei2019presslight}, where an RL agent was trained for every individual intersection. The main idea is to use the pressure as the reward function, thus the algorithm is called PressLight. To efficiently model the influence of the neighbor intersections, a graph attentional network \cite{velivckovic2017graph} is utilized in CoLight \cite{wei2019colight}. See \cite{wei2019survey} for a detailed review of conventional and RL-based methods for TSCP. 

  There have been attempts to transfer the learned experiences between different intersections. For example, in FRAP algorithm \cite{zheng2019learning} a trained model for a 4-way intersection needs to be changed through reducing the neurons as well as zero-padding modifications to make it compatible for a 3-way intersection. A modification of FRAP was proposed on \cite{zangmetalight}. This new algorithm is called MetaLight, where the key idea is to use meta-learning strategy proposed in \cite{finn2017model} to make a more universal model. However, MetaLight still needs to re-train its model parameter for any new intersection. In \cite{wei2019deep}, a multi-agent RL algorithm is proposed to control the traffic signals for multiple intersections with arterial streets. The key idea is to train a single agent for an individual intersection and then apply {\it Transfer Learning} to obtain another model for a 2-intersection structure. Similarly, a model for a 3-intersection structure can be obtained from 2-intersection and etc. Besides the natural drawbacks of Transfer Learning, such as the assumption on the data distribution, the proposed model in this work is not robust in terms of the number of approaching roads, lanes, and phases. For example, a trained model for a single 4-way intersection cannot be transferred to a 3-way one.

\section{Traffic Signal Control Problem}

We consider a single-intersection traffic signal control problem (TSCP). An \textit{intersection} is defined as a junction of a few roads, where it can be in the form of 3-way, 4-way or it can have a more complex structure with five or more approaching roads. Each road might have one or two direction(s) and each direction includes one lane or more. Let $\mathcal{M}$ be a set of intersections, where each intersection $m\in\mathcal{M}$ is associated with a known intersection topology and traffic-data. Throughout this paper, we simplify notations by omitting their dependency on $m$.

Let us define $s_l^t$ as the \emph{traffic characteristics} of lane $l\in \mathcal{L}$ at time $t$ , where we use $\mathcal{L}$ to denote the set of all approaching (i.e., either entering or leaving) lanes to the intersection. We denote by $\mathcal{L}^{in}$ and $\mathcal{L}^{out}$ the set of entering and leaving lanes, respectively. Clearly, $\mathcal{L}^{in}\cup \mathcal{L}^{out}=\mathcal{L}$ holds. In TSCP, different traffic characteristics for each lane are proposed in the literature, e.g., queue length, waiting time, delay, the number of moving vehicles, etc. See \cite{wei2019survey} for more details. The model we propose in this work is not restricted to any of the mentioned characteristics and assigning $s_l^t$ to any of these is completely up to the practitioner. There is a finite number of ways that the cars can transition from the entering lanes to leaving lanes. We define a \emph{traffic movement} $\upsilon_l$ as a set that maps the entering traffic of lane $l\in\mathcal{L}^{in}$ to possible leaving lanes $l' \in\mathcal{L}^{out}$. 

The set of the traffic movements which are valid during a green light is called a \emph{phase}, and it represented by $p$ and we denote by $\mathcal{P}$ the set of all phases. An intersection has at least two phases, and they may have some shared traffic movement(s). We define the \emph{participating lanes} $\mathcal{L}_p$ as the set of lanes that have appeared in at least one traffic movement of phase $p$.  Each phase runs for a given minimum amount of time and after that, a decision about the next phase should be taken. 

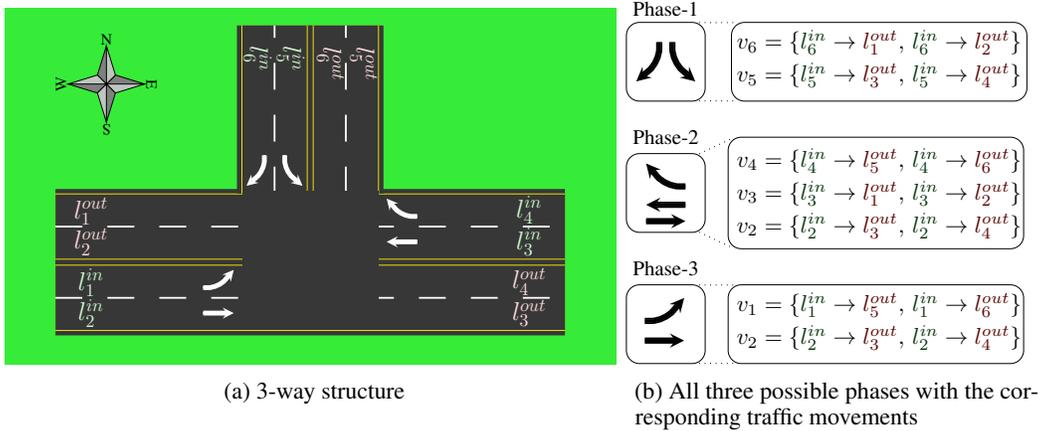
\begin{figure}[htbp]
\captionsetup[subfigure]{width=0.95\textwidth}
\centering
\begin{subfigure}[t]{.59\textwidth}
    \centering
    \resizebox{1\textwidth}{!}{
        \rule[2cm]{-1.5cm}{.6cm}\input{intersection1}
    }
    \caption{3-way structure}
    \label{fig:3_way_intersection_roads}
\end{subfigure}%
\begin{subfigure}[t]{.4\textwidth}
    \centering
    \resizebox{.98\textwidth}{!}{
        \rule[2cm]{-1.2cm}{.6cm}\input{intersection1_traffic}
    }
    \caption{All three possible phases with the corresponding traffic movements}
    \label{fig:3_way_intersection_phases}
\end{subfigure}
\caption{A 3-way intersection topology with its available phases and traffic movements} 
\label{fig:3_way_intersection}
\end{figure}

To further clarify our problem definition and the terminologies, we  use a 3-way intersection example, depicted in Figure \ref{fig:3_way_intersection}. There are six lanes entering the intersection and six leaving lanes. We have labeled these lanes with $l^{in}_{k}$ and $l^{out}_{k}$, $k\in\{1,\cdots,6\}$ in Figure \ref{fig:3_way_intersection_roads}. Accordingly, there are six traffic movements denoted by $\upsilon_1, \cdots, \upsilon_6$ as illustrated in Figure~\ref{fig:3_way_intersection_phases}. Also, three possible phases for this intersection along with the valid traffic movements of each phase are listed in Figure~\ref{fig:3_way_intersection}. For example, phase-1 involves two traffic movements $\upsilon_6= \left\{l^{in}_{6} \to l^{out}_{1}, l^{in}_{6} \to l^{out}_{2} \right\}$ and $\upsilon_5 = \left\{l^{in}_{5} \to l^{out}_{3}, l^{in}_{5} \to l^{out}_{4}\right\}$, hence the set of participating lanes associated with this phase is $\mathcal{L}_1 = \{l^{in}_{5}, l^{in}_{6}, l^{out}_{1}, l^{out}_{2}, l^{out}_{3}, l^{out}_{4}\}$. As one may notice, the number of traffic movements as well as the number of  participating lanes are not necessarily the same for different phases. For example, in Figure \ref{fig:3_way_intersection} phase-1 and phase-3 involve two traffic movements, while there are three traffic movements in phase-2. Further, phase-1 and phase-3 involve six participating lanes while phase-2 includes nine lanes. 
This results in different size of the input/output of the model among different intersection instances. Therefore, building a universal model which handles such complexity is not straightforward using conventional deep RL algorithms. To address this issue, we design AttendLight which uses a special attention mechanism as described in the next section.

The main goal of TSCP is to minimize the average travel time of all vehicles within a finite time frame. However, the travel time is not a direct function of state and action so in practice it is not possible to optimize it directly. Zheng et al. \cite{wei2019presslight} use an alternative reward measure, namely \emph{pressure}, which is defined as the absolute value of the total number of waiting vehicles in entering lanes minus the total number of leaving vehicles. They show that minimizing pressure is equivalent to minimizing the average travel time. 
Similarly, in this work we also choose to minimize the pressure of the intersection, though we report the average travel times in the numerical experiments.

\section{Reinforcement Learning Design}

In this section we present our end-to-end RL framework for solving the TSCP. To formulate this problem into an RL context, we first require to identify \emph{state}, \emph{action}, and \emph{reward}. \\
{\bf State}. The state at time $t$ is the traffic characteristics $s^t_l$ for all lanes $l\in\mathcal{L}$, i.e., $s^t = \left\{s^t_l, \,l \in \mathcal{L} \right\}$. \\
{\bf Action}. At each time step $t$, we define the action as the active phase at time $t+1$.\\
{\bf Reward}. Following the discussion in \cite{wei2019presslight}, the reward in each time step is set to be the negative of intersection pressure.

In TSCP, we are interested in learning a policy $\pi$, which for a given state $s^t$ of an intersection suggests the phase for the next time-step in order to optimize the long-term cumulative rewards. We design a unified model for approximating $\pi$ that fits to every intersection configuration. The AttendLight model that we present in Section \ref{sec:attendlight} instantiates such a policy $\pi$ that achieves this universality by appropriate use of two attention mechanisms. 

\subsection{Attention Mechanism}\label{sec:attention}
Attention mechanism introduced for natural language processing \cite{bahdanau2014neural,luong2015effective}, but it has been proved to be effective in other domains such as healthcare \cite{choi2016retain}, combinatorial optimization \cite{bello2016neural, nazari2018reinforcement}, and recommender systems \cite{seo2017interpretable}. By providing a mechanism to learn the importance of each element in a problem with variable number of inputs, attention allows having a fixed length encoding of inputs. The attention mechanism that we use in this work is the key for achieving the universal capability of the AttendLight.
 We adjust the attention mechanism of \cite{bahdanau2014neural} for solving  TSCP. In this setting, we have two types of inputs to $\texttt{attention}(\cdot,\cdot)$, namely a set of references $\{r_i\}$ and a {query} $q$. The attention computes an \textit{alignment} $a \coloneqq \{a_i\}$, where
\begin{align}
    a_i = u_a\texttt{tanh}(U_r \bar{r}_i+U_q \bar{q}),
\end{align}
in which $\bar{q}$ and $\bar{r}_i$ are trainable linear mappings of $q$ and $r_i$, respectively; $U_q,U_r,u_a$ are trainable variables. Note that $a$ has the same size as the input reference set $\{r_i\}$. Finally, the attention returns $w$, which is a probability distribution computed as $\texttt{softmax}(a)$. 

\begin{figure}[t!]
    \centering
    \resizebox{0.97\textwidth}{!}{
        \rule[-0cm]{-9cm}{-0cm}\input{att_model_2}
    }
    \caption{AttendLight model}
    \label{fig:att_model}
\end{figure}
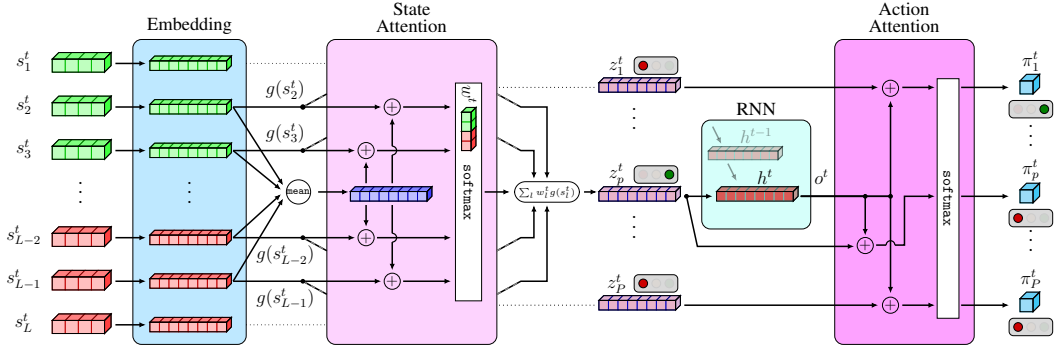

\subsection{AttendLight Algorithm} \label{sec:attendlight}
AttendLight is our proposed algorithm and has two major responsibilities: \textit{i}) extracting meaningful \textit{phase representations} $z^t_p$ for every phase $p$, and \textit{ii}) deciding on the next active phase. To add universality to these responsibilities, the input and output dimension of the model needs to be independent of intersection configuration. To this order, we propose two attention mechanisms---as introduced in Section \ref{sec:attention}---called {\tt state-attention} and {\tt action-attention} for handling the phase representation from the raw-state and for choosing the next phase, respectively.
The policy model that we used in AttendLight is visualized in Figure \ref{fig:att_model}. Next, we explain how AttendLight achieves these goals.

A key part of AttendLight is to construct meaningful phase representations. Although identifying the phase representations for TSCP is not a trivial process, we expect that it could be extracted from the traffic characteristic of the participating lanes $\left\{s^t_l, \,l \in \mathcal{L}_p \right\}$. For example, the number of cars in participating lanes can be utilized to infer how congested the traffic movements are, how close the cars are with respect to the intersection, and whether or not activating a phase will ease the traffic. AttendLight uses an embedding followed by an attention mechanism to learn the phase representation.

In order to allow more capacity in feature extraction, we embed $s_l^t$ into a higher dimensional space. In this work, we use a single \texttt{Conv1D} transformation as our embedding function $g$, but one may use more intricate embedding functions as well. The {\tt state-attention} uses these embeddings and returns the importance weights $w_p^t \coloneqq \{w^t_l,\, l \in\mathcal{L}_p\}$ for attending on $g(s^t_l)$
of each participating lane in phase $p$, i.e.,
\vspace{-3mm}
\begin{align}
        & w^t_p= \texttt{state-attention} \left(r_p^t,  q_p^t \right) , \,\forall p\in\mathcal{P}, 
\end{align}
where the references $r_p^t$ are the embedded participating lane characteristics and the query $q_p^t$ is their average, i.e.,
\vspace{-10pt}
\begin{align*}
    r_p^t \coloneqq \left\{g(s^t_l), \, l\in \mathcal{L}_p \right\} \qquad  q_p^t \coloneqq \sum_{l\in \mathcal{L}_p} \frac{g(s^t_l)}{|\mathcal{L}_p|}.
\end{align*}
Then, the phase representation $z^t_p$ is computed as $z^t_p = \sum_{l\in\mathcal{L}_p} w_l^t \times g(s^t_l)$. From this formulation, we can see that the phase representation is independent of the number of participating lanes in each phase since $\mathcal{L}_p$ can have any arbitrary cardinality. It worth mentioning that using the average of $g(s_l^t)$ as the query $q_p^t$ in the state-attention, allows the model to learn the importance of each lane-traffic compared to the average traffic. For example, whenever all lanes are either full or mostly empty, the query detects the situation and in turn, the attention model assigns appropriate weights to each lane.

To capture the sequential information of active phases, we incorporate an LSTM cell. At every time step, the phase representation $z_p^t$ of the active phases, i.e., the phase associated with the current green light is fed into the cell. LSTM uses a hidden memory $h^t$ to store encoded sequence of active phase in previous times and produces an output $o^t$. Then, the {\tt action-attention} model provides the probability $\pi^t\coloneqq\{\pi^t_p,\,p\in\mathcal{P}\}$ of switching to any of the phases in the next time step using
\begin{align} 
   \pi^t = \texttt{action-attention} \left(\{z^t_p,\, p\in\mathcal{P}\}, o^t \right),
    \label{eq:output_attention}
\end{align}
Hence, AttendLight provides an appropriate mapping of the states to probability of taking actions for any intersection configuration, regardless of the number of roads, lanes, traffic movements, and number of phases. We would like to emphasize that AttendLight is invariant to the order of lanes or phases, so how to enumerate these components will result in the same control decisions.

\subsection{RL Training}

We train the AttendLight with a variance-reduce variant of REINFORCE \cite{sutton2000policy}. This training algorithm is quite standard, so we leave its detailed description to Appendix \ref{appsec:rl-train}. 
%
%
To train the AttendLight, we follow two regimes for the intersection sampling process: (i) train for a single environment and deploy on the same environment, which we refer to it as ``\textit{single-env}'' regime, and (ii) train on multiple environments and deploy on multiple environments, which we call is as ``\textit{multi-env}'' regime. In the first regime, we use a particular environment instance $m$ to train a policy and deploy the model on the same environment instance $m$. This is the common practice in all of the current RL algorithms for TSCP \cite{zheng2019learning, zangmetalight, wei2019presslight}, in which the trained model only works on the intended problem. While in the second regime, in each episode we sample $n$ environments from $\mathcal{M}$ and run the train-step based on all those environments. In experiments of this paper, we let $n=|\mathcal{M}|$. However, in Appendix \ref{appsec:stoch-regime}, we present alternative multi-env regimes to deal with a large number of intersections. 

\section{Numerical Experiments}
Many variants of the intersections have been studied in the literature and we claim that a single model can provide reasonable phase decisions for all of them. We consider 11 intersection topologies, where they vary in terms of the number of approaching roads (i.e., 3-way or 4-way), and the number of lanes in each road. Further, each of  these 11 intersections may have a different number of phases and traffic-data. Table~\ref{tb:definition_of_all_intersection} summarizes the properties of all intersections.

To train and test AttendLight, a combination of real-world and synthetic traffic-data is utilized. For 4-way intersections with two lanes, we use the real-world traffic-data of intersections in Hangzhou and Atlanta \cite{zheng2019learning, wei2019survey}. For notation simplicity, we denote these data by \texttt{H1}, $\cdots$, \texttt{H5} and \texttt{A1}, $\cdots$, \texttt{A5} for Hangzhou and Atlanta, respectively. For the rest of the intersections with two lanes (e.g., 3-way intersections), slightly adapted version of these data-sets are used. Due to lack of real-world data for the intersection with three lanes, we created synthetic traffic-data denoted by \texttt{S1}, $\cdots$, \texttt{S6}. The combination of intersection topologies, their available phases, and traffic-data allows us to construct the set $\mathcal{M}$ with 112 unique intersection instances. We label each intersection instance by {\tt {INT\#-dataID}-\#phase}. For example, the problem which runs intersection \texttt{INT5} with the \texttt{H1} traffic-data through 8-phase is denoted by {\tt INT5-H1-8}. Table \ref{tb:definition_of_all_intersection} summarizes the properties of all 11 intersections used in this study.\vspace{-15pt}
\begin{table}[ht]
\centering
\caption{All intersection configurations}
\label{tb:definition_of_all_intersection}
\begin{adjustbox}{width=1\textwidth}
\begin{tabular}{lccccccccccc}
\toprule
intersection ID           & \texttt{INT1}  & \texttt{INT2}       & \texttt{INT3}              & \texttt{INT4}             & \texttt{INT5}             & \texttt{INT6}          & \texttt{INT7}          & \texttt{INT8}                      & \texttt{INT9}           & INT10             & INT11                       \\
\#road                                                         & 3     & 3          & 3                 & 3                & 3                & 3             & 4             & 4                         & 4              & 4                 & 4                           \\
\#lanes per road                                                         & 2     & 2          & 1,2               & 1,3              & 1,3              & 1,3           & 2             & 2                         & 3              & 2,3               & 2,3                         \\
\#phase                                                        & 3     & 3          & 2                 & 3                & 3                & 3             & 4,8           & 3                         & 4,8            & 4,8               & 3,5                         \\
($\min_p |\mathcal{L}_p|, \max_p |\mathcal{L}_p|)$ & (1,2) & (1,2)      & (1,2)             & (2,4)            & (2,4)            & (2,4)         & (2,2)         & (2,2)                     & (6,6)          & (6,6)             & (6,6)    \\\bottomrule                  
\end{tabular}
\end{adjustbox}
\end{table}

In all experiments, we choose the number of moving and waiting vehicles to represent traffic characteristic $s_l^t$. To this order, first, for lane $l$ we consider a segment of 300 meters from the intersection and split it into three chunks of 100 meters. Then, $\alpha_{l,c}^t$ for $c=1,2,3$ is the number of moving vehicles in chunk $c$ of the lane $l$ at time $t$. Also, we define $\beta_l^t$ as the number of waiting vehicles at lane $l$ at time $t$. Now, we represent the traffic characteristic of lane $l$ by $s_l^t \coloneqq [\alpha_{l,1}^{t},\alpha_{l,2}^{t},\alpha_{l,3}^{t}, \beta_l^t]$.

It is assumed that the traffic always can turn to the right unless there is conflicting traffic or there is a "no turn on red" signal. To clear the intersection, the green light is followed by 5 seconds of yellow light. For each intersection, we run the planning for the next 600 seconds with a minimum active time of 10 seconds for each phase. To simulate the environment, we used CityFlow \cite{zhang2019cityflow}. For more details on data construction, simulator, and intersections configurations, see Appendix.

\subsection{Results of the Single-Environment Regime}\label{sec:results-single-intersection}

\begin{figure}
	\centering
	\centering
        {	\includegraphics[trim=5 100 5 110,clip, width=0.75\textwidth]{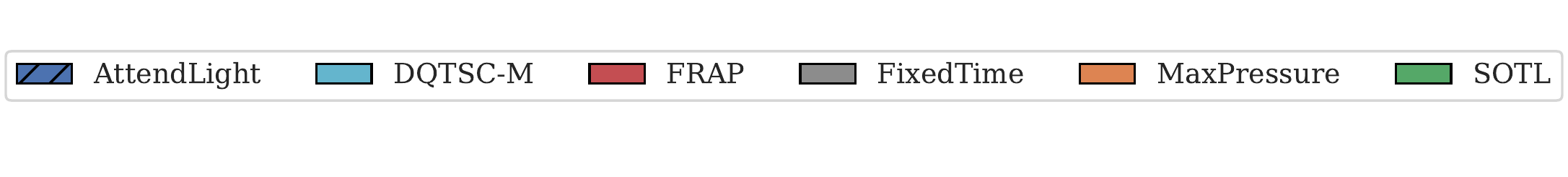}\vspace{-12pt}
}   
	\begin{subfigure}{0.245\textwidth}
		\centerline{ \includegraphics[width=\textwidth]{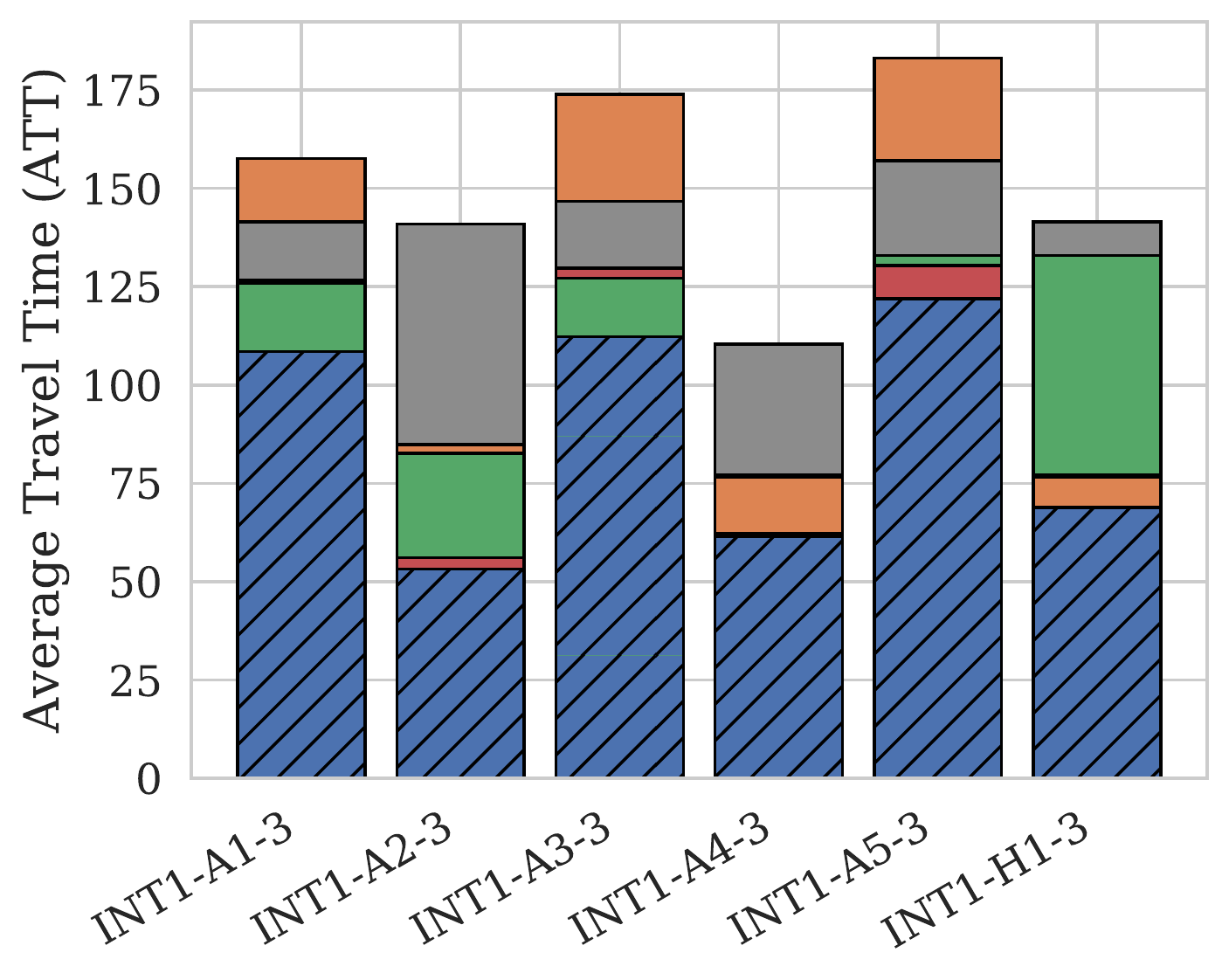}}
		\vspace{-5pt}
	\end{subfigure}
	\begin{subfigure}{0.245\textwidth}
		\centerline{ \includegraphics[width=\textwidth]{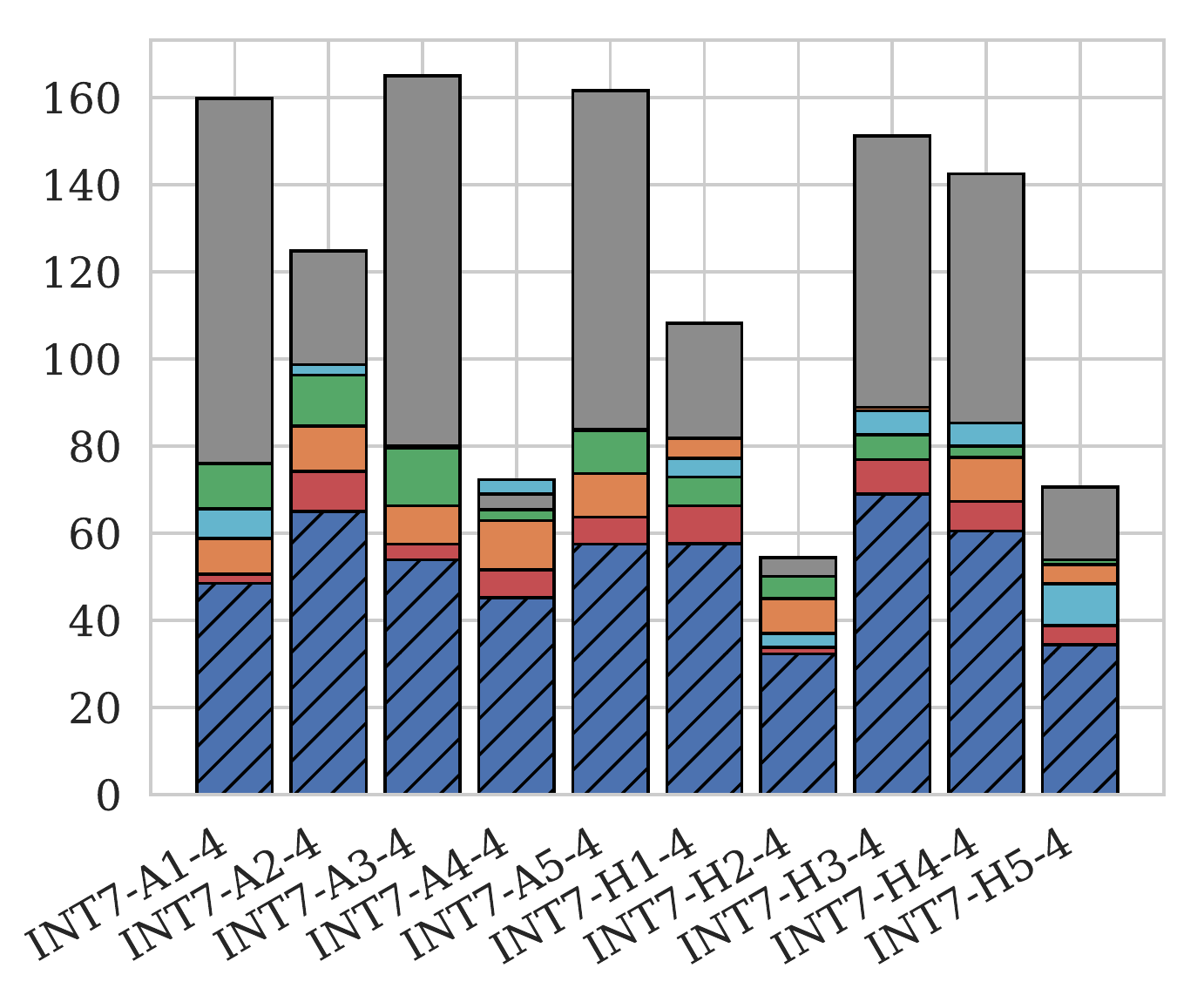}}
		\vspace{-5pt}
	\end{subfigure}
	\begin{subfigure}{0.245\textwidth}
		\centerline{ \includegraphics[width=\textwidth]{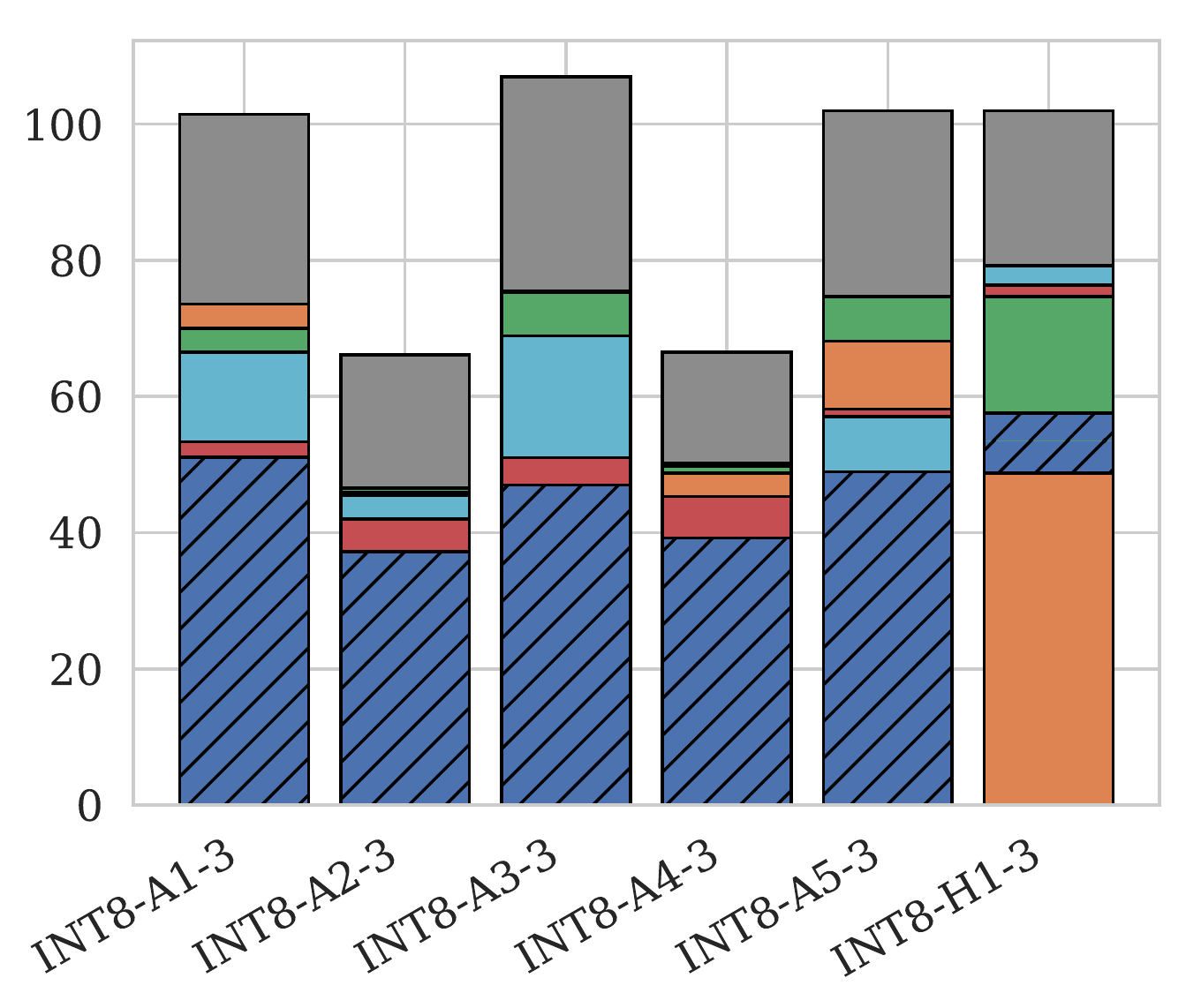}}
		\vspace{-5pt}
	\end{subfigure}
	\begin{subfigure}{0.245\textwidth}
		\centerline{ \includegraphics[width=\textwidth]{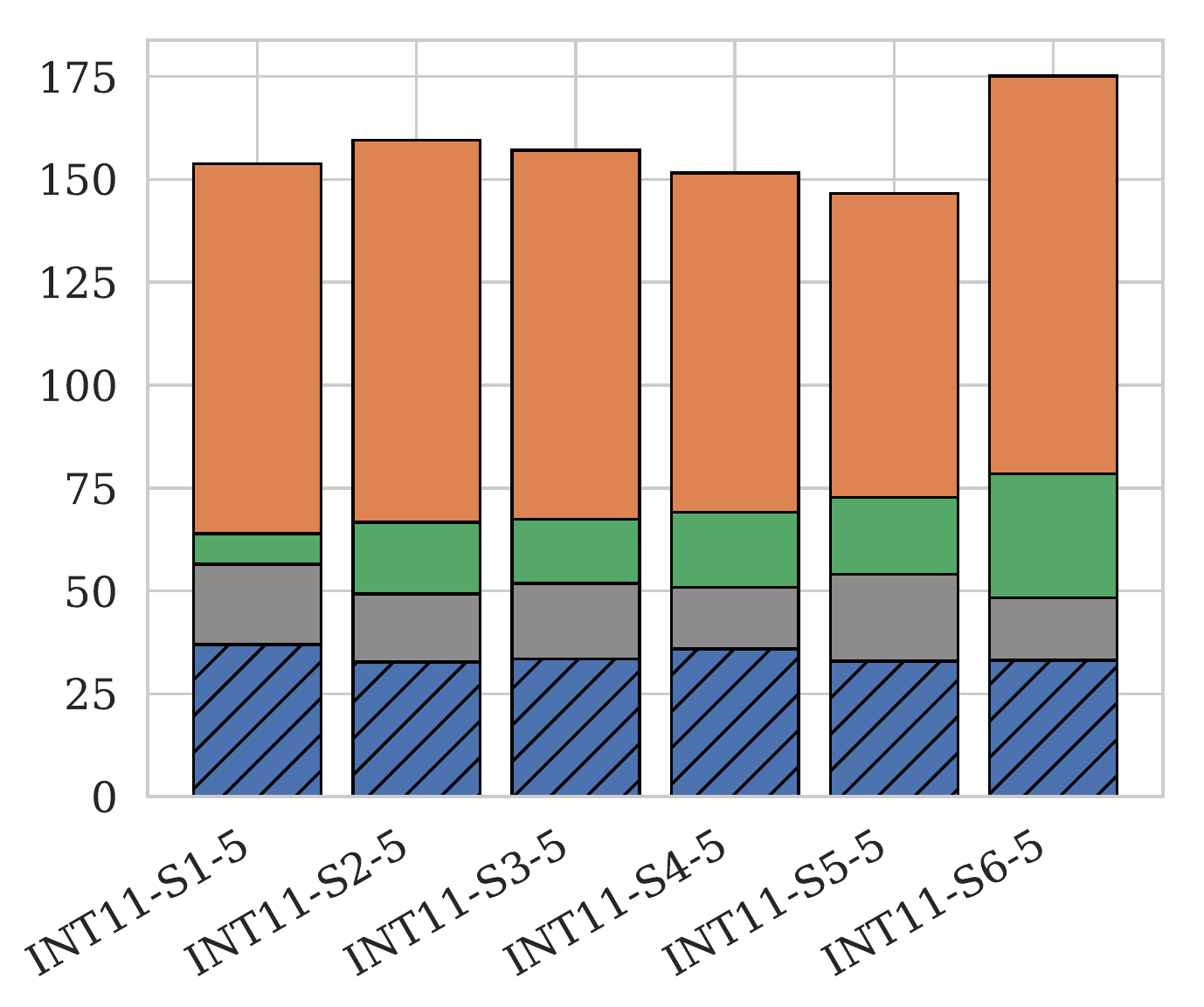}}
		\vspace{-5pt}
	\end{subfigure}

\vspace{3pt}

    \begin{subfigure}[t]{.245\textwidth}
        \centering
        \resizebox{0.7\textwidth}{!}{
            \rule[2cm]{-1.5cm}{.6cm}\input{intersection1_noinout}
        }
		\caption{INT1, 3-phase} \label{fig:sample_result_int1}
    \end{subfigure}
%
    \begin{subfigure}[t]{.245\textwidth}
        \centering
        \resizebox{.7\textwidth}{!}{
            \rule[2cm]{-1.5cm}{.6cm}\input{intersection2}
        }
		\caption{INT7, 4-phase} \label{fig:sample_result_int7}
    \end{subfigure}
%
    \begin{subfigure}[t]{.245\textwidth}
        \centering
        \resizebox{.7\textwidth}{!}{
            \rule[2cm]{-1.5cm}{.6cm}\input{intersection5}
        }
		\caption{INT8, 3-phase} \label{fig:sample_result_int8} 
    \end{subfigure}
%
    \begin{subfigure}[t]{.245\textwidth}
        \centering
        \resizebox{.7\textwidth}{!}{
            \rule[2cm]{-1.5cm}{.6cm}\input{intersection6}
        }
		\caption{INT11, 5-phase} \label{fig:sample_result_int11} 
	\end{subfigure}    
	\caption{The comparison of AttendLight in single-env regime with baseline algorithms. The corresponding intersection configuration is depicted below the ATT results.  }
	\label{fig:sample_result_single_env}
\end{figure}

In this experiment, we train the AttendLight for a particular intersection instance and then test it for the same intersection configuration. We compare AttendLight with SOTL \cite{gershenson2004self}, Max-pressure \cite{varaiya2013max}, Fixed-time policies, DQTSC-M \cite{shabestary2018deep}, and FRAP \cite{zheng2019learning}. The FRAP by design does not handle all the intersections that we consider in this work, so we compare against it when possible. 
For the purpose of comparison, we measure the Average Travel Time (ATT) for all algorithms. 
We consider a variety of common real-world intersections configurations with \{2,3,4,5,8\} phases, \{1,2,3\} lanes, \{3,4\} ways, and one- and bi-directional intersections. For example, Figure \ref{fig:sample_result_single_env} presents the results for four different cases. As it is shown, in most cases AttendLight outperforms benchmarks algorithms in terms of ATT, and FRAP is the second-best working algorithm. In Figure~\ref{fig:sample_result_int11}, FRAP is not applicable. When considering all 112 cases, AttendLight yields 46\%, 39\%, 34\%, 16\%, and 9\% improvement over FixedTime, MaxPressure, SOTL, DQTSC-M, and FRAP, respectively. 
Thus, when AttendLight is used solely to train a single environment, it works well for all the available number of roads, lanes, phases, and all traffic-data. Further results for other intersections are available in Appendix~\ref{apdx:details_results_single_multi_environment}.

\subsection{Results of Multi-Environment Regime}\label{sec:results-all-intersection}
We evaluate the key feature of AttendLight that enables it to be utilized for multiple TSCPs. We divide the set of intersection instances into two segments: training and testing sets, each with 42 and 70 instances, respectively. We train AttendLight by running all 42 training intersection instances in parallel to obtain data for the training and use the REINFORCE algorithm to optimize the trainable parameters. Once the trained model is available, we test the performance on training and testing instances. The testing set includes new intersection topologies as well as new traffic-data that has not been observed during training. Appendix~\ref{sec:apdx:description_of_all_cases} summarizes the details of both sets.

To the best of our knowledge, there is no RL-based algorithm in the literature that works on more than one intersection instance without any transfer learning or retraining. Hence, to evaluate the performance of AttendLight in the multi-env regime, we have no choice other than comparing it against the single-env regime and the previously explained baselines, e.g., SOTL, MaxPressure, FixedTime, DQTSC-M, and FRAP. Recall that for the AttendLight in single-env regime, DQTSC-M, and FRAP, a separate model is trained for every instance. In contrast, the multi-env regime trains a single model to make the traffic decisions of multiple intersections. Clearly, having such a ``general'' model will lead to an inevitable performance loss, which makes the comparison unfair. Nevertheless, we embrace such unfairness in our experiments. Comparing AttendLight in multi-env versus single-env also allows measuring the performance degradation as a result of training a general model. Finally, to make sure that the results are robust against the randomness, we trained five models with different random seeds and always report the average statistics.

\begin{figure}
    \centering
\begin{subfigure}[b]{0.41\textwidth}
    \centering
    \includegraphics[width=\textwidth]{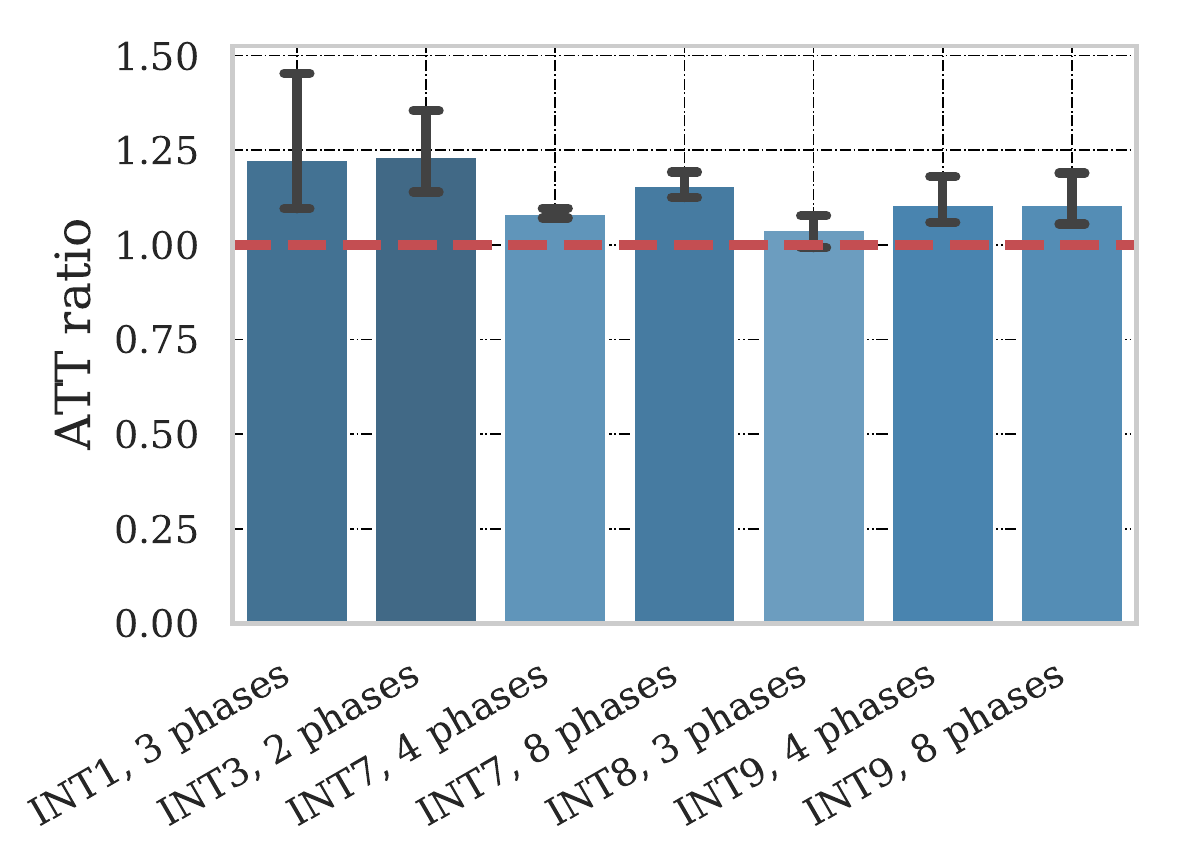}
    \caption{Training intersections instances}
    \label{fig:result-all-env-over-best-attendlight-training}
\end{subfigure}
\begin{subfigure}[b]{0.534\textwidth}
    \centering
    \includegraphics[width=\textwidth]{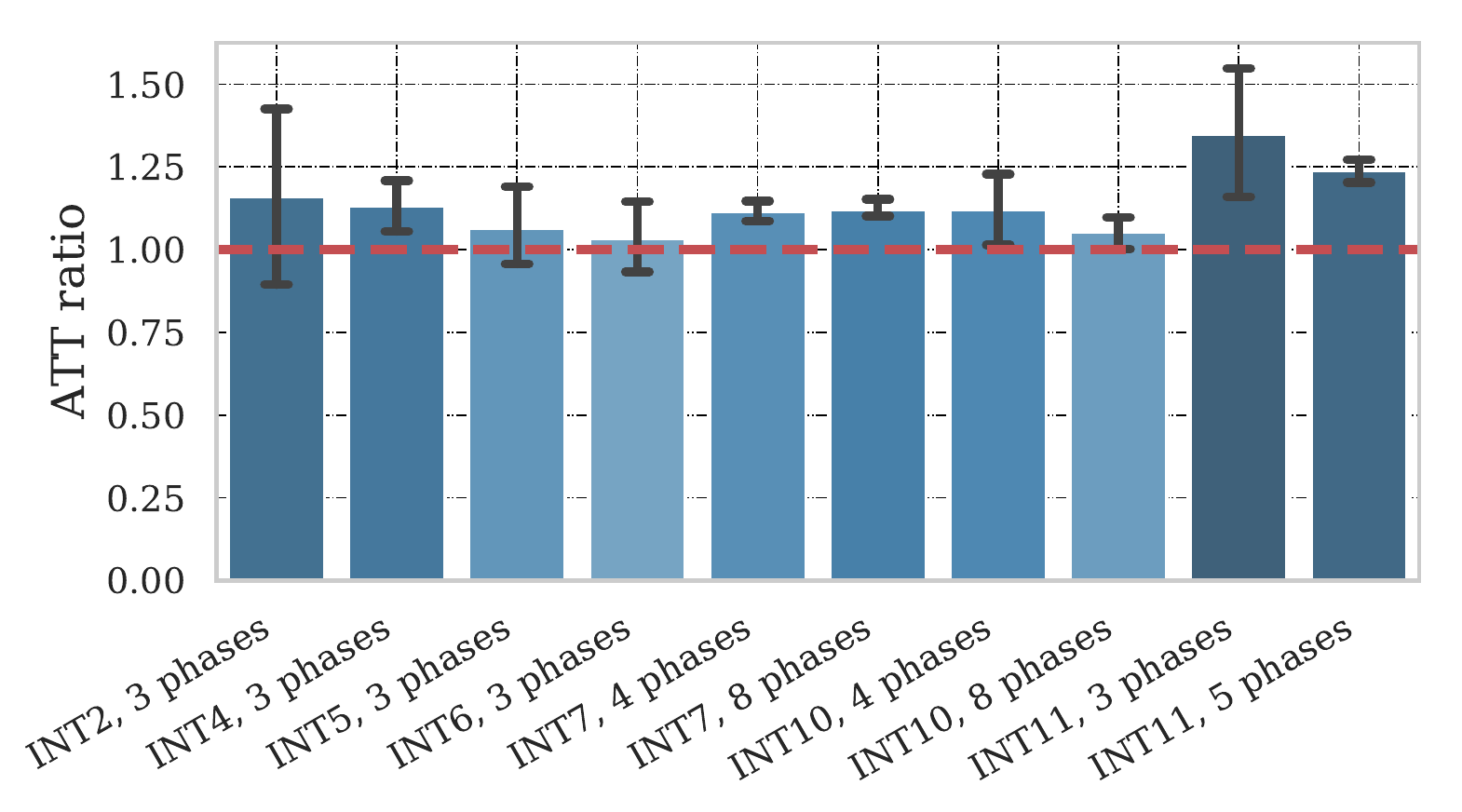}
    \caption{Testing intersections instances}
    \label{fig:result-all-env-over-best-attendlight-testing}
\end{subfigure}
    \centering
    \caption{The ATT of multi-env policy divided by ATT of single-env policy. The error bars represent 95\% confidence interval for ATT ratio. The closer ATT ratio is to one (red dashed line), the less degradation is caused as a result of using a multi-env model.}
    \label{fig:result-all-env-over-best-attendlight}
\end{figure}

Now, we would like to evaluate the performance of the multi-env regime versus single-env. Our goal in this analysis is twofold: In case (I) we evaluate how introducing a universal model trained in multi-env regime exacerbates the ATT in the training set, and case (II) demonstrates how the multi-env model generalizes to testing instances which are not visited during training. To measure this, we calculate \textit{ATT ratio}, which is ATT in multi-env divided by ATT of the single-env regime. The closer ATT ratio is to one, the less degradation is caused as a result of using a multi-env model.  Figure~\ref{fig:result-all-env-over-best-attendlight} summarizes the ATT ratio for all 112 intersection instances. Each bar in this plot illustrates the average ATT ratio for a number of traffic-data associated with an intersection configuration and the error bars represent the 95\% confidence interval. To analyze case (I), we use Figure~\ref{fig:result-all-env-over-best-attendlight-training}. We observe that in most cases the results of the multi-env regime are close to that of the single-env regime. In particular, we have an average 15\% ATT degradation with a standard deviation of 0.15. Case (II) analysis is  demonstrated in Figure~\ref{fig:result-all-env-over-best-attendlight-testing}.  We observe that the trained policy works well in the majority of intersection/traffic-data such that on average it has 13\% ATT gap with a standard deviation of 0.19. Furthermore, from this figure, one may notice that there are several intersection instances that the model-trained in the multi-env regime has lower ATT. We conjecture that such behavior is due to knowledge sharing between intersections, meaning that environments are sharing successful phase controls that are not necessarily explored in single-env training cases. 

\begin{figure}
\centering
\begin{subfigure}{.19\textwidth}
    \centering
    \includegraphics[width=\textwidth]{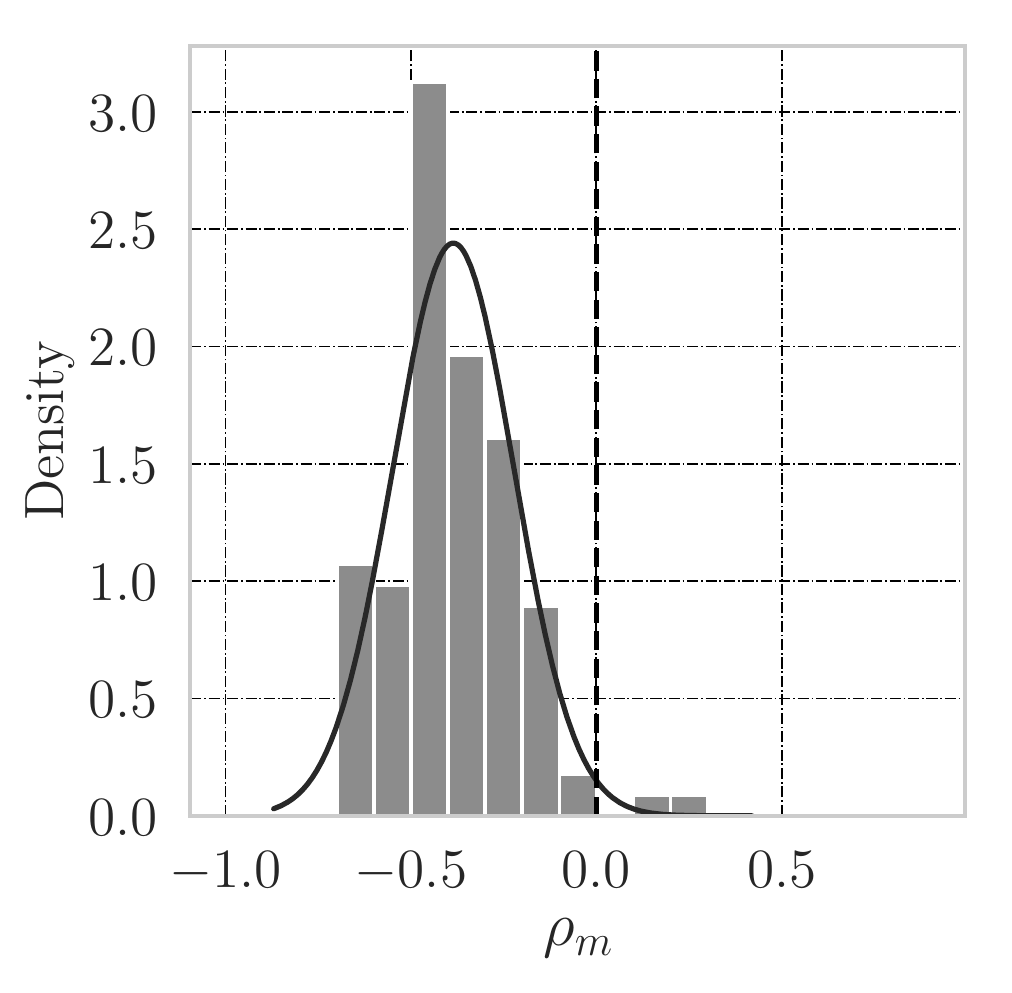}
    \caption{FixedTime}
    \label{fig:all-env-normal-dist-of-results-fixedtime}
\end{subfigure}
\begin{subfigure}{.19\textwidth}
    \centering
    \includegraphics[width=\textwidth]{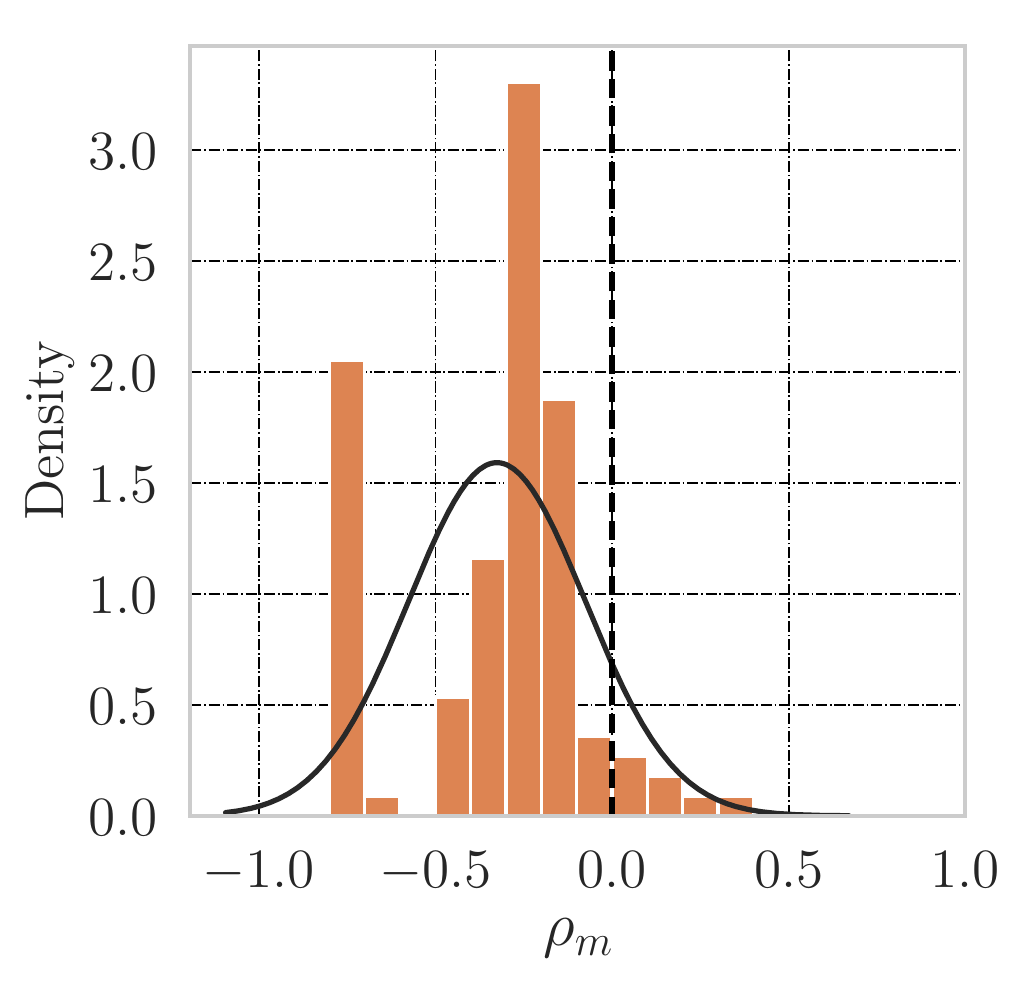}
    \caption{MaxPressure}
    \label{fig:all-env-normal-dist-of-results-maxpressure}
\end{subfigure}
\begin{subfigure}{.19\textwidth}
    \centering
    \includegraphics[width=\textwidth]{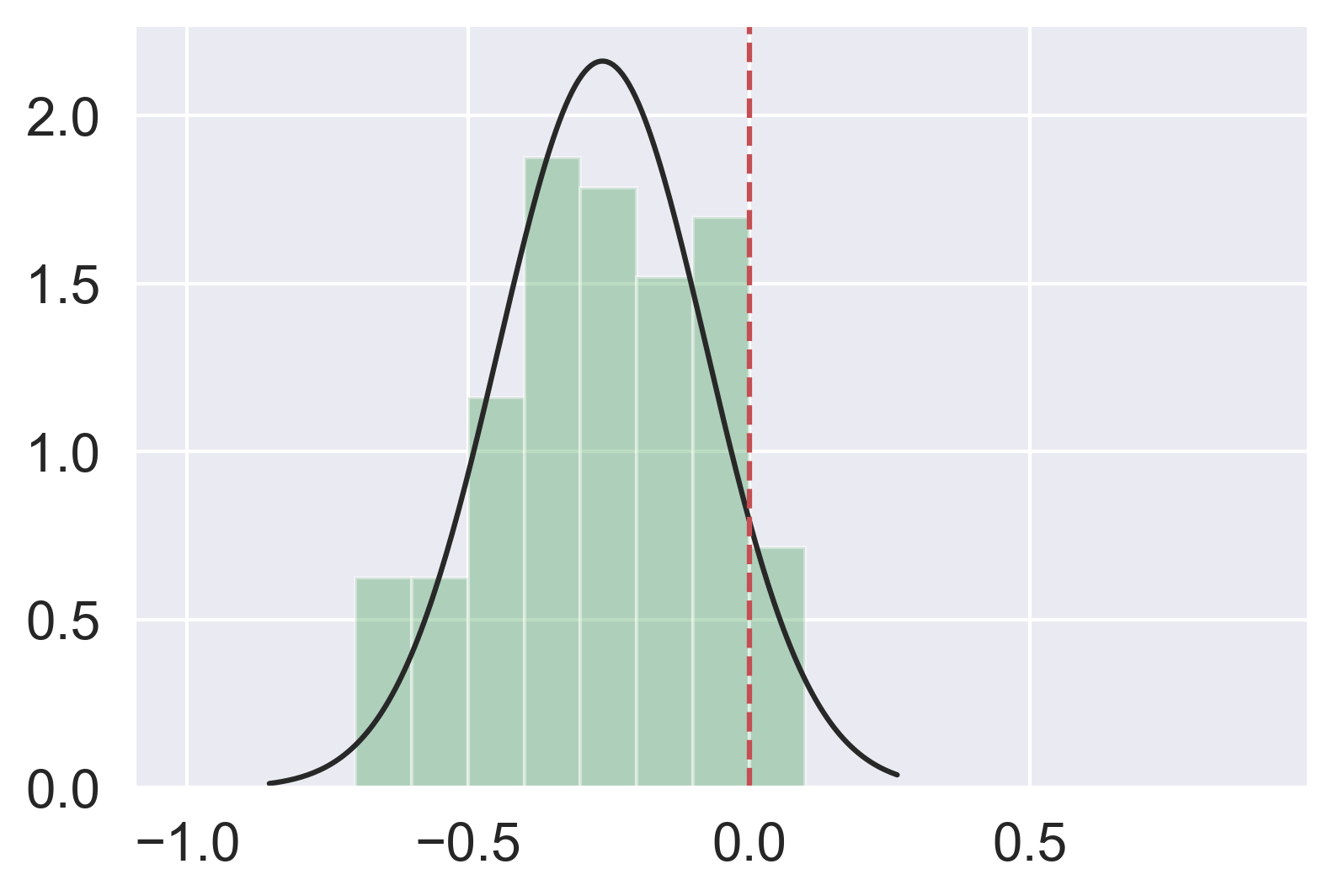}
    \caption{SOTL}
    \label{fig:all-env-normal-dist-of-results-sotl}
\end{subfigure}
\begin{subfigure}{.19\textwidth}
    \centering
    \includegraphics[width=\textwidth]{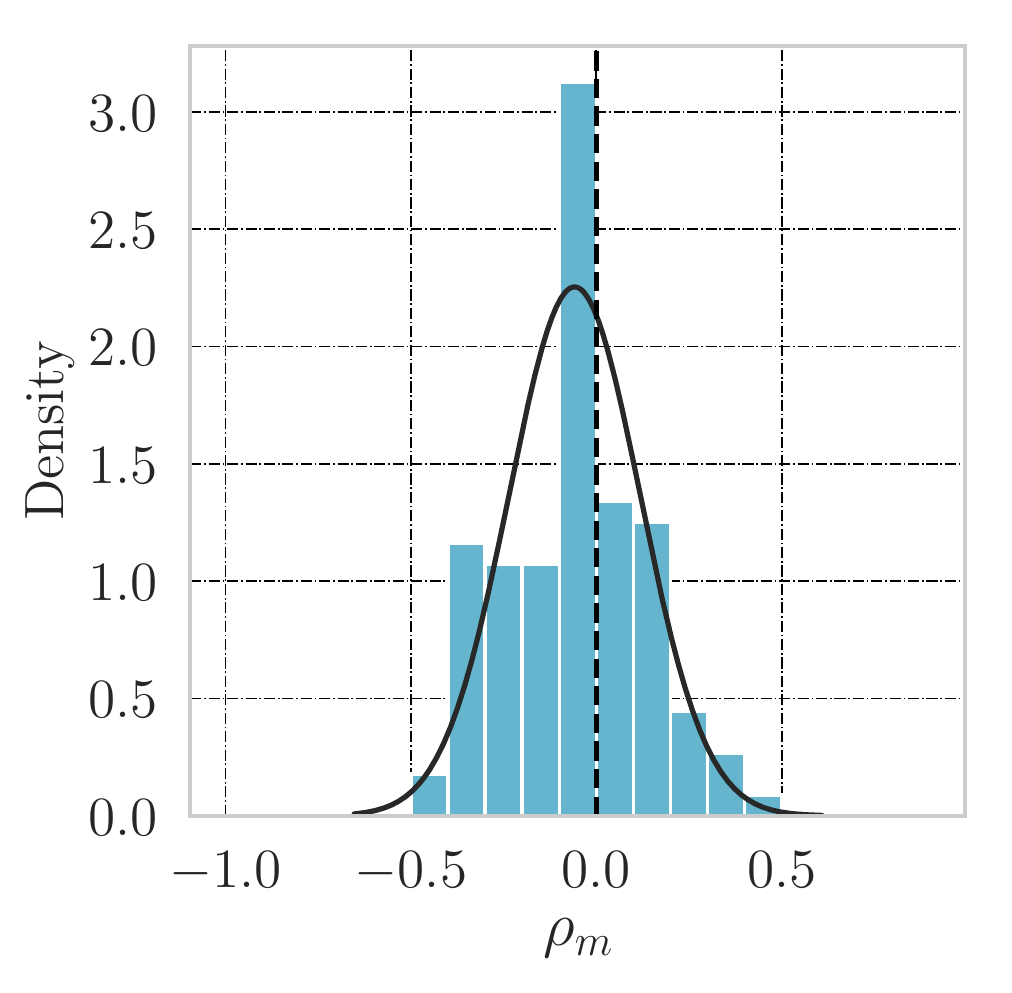}
    \caption{DQTSC-M}
    \label{fig:all-env-normal-dist-of-results-dqtscm}
\end{subfigure}
\begin{subfigure}{.19\textwidth}
    \centering
    \includegraphics[width=\textwidth]{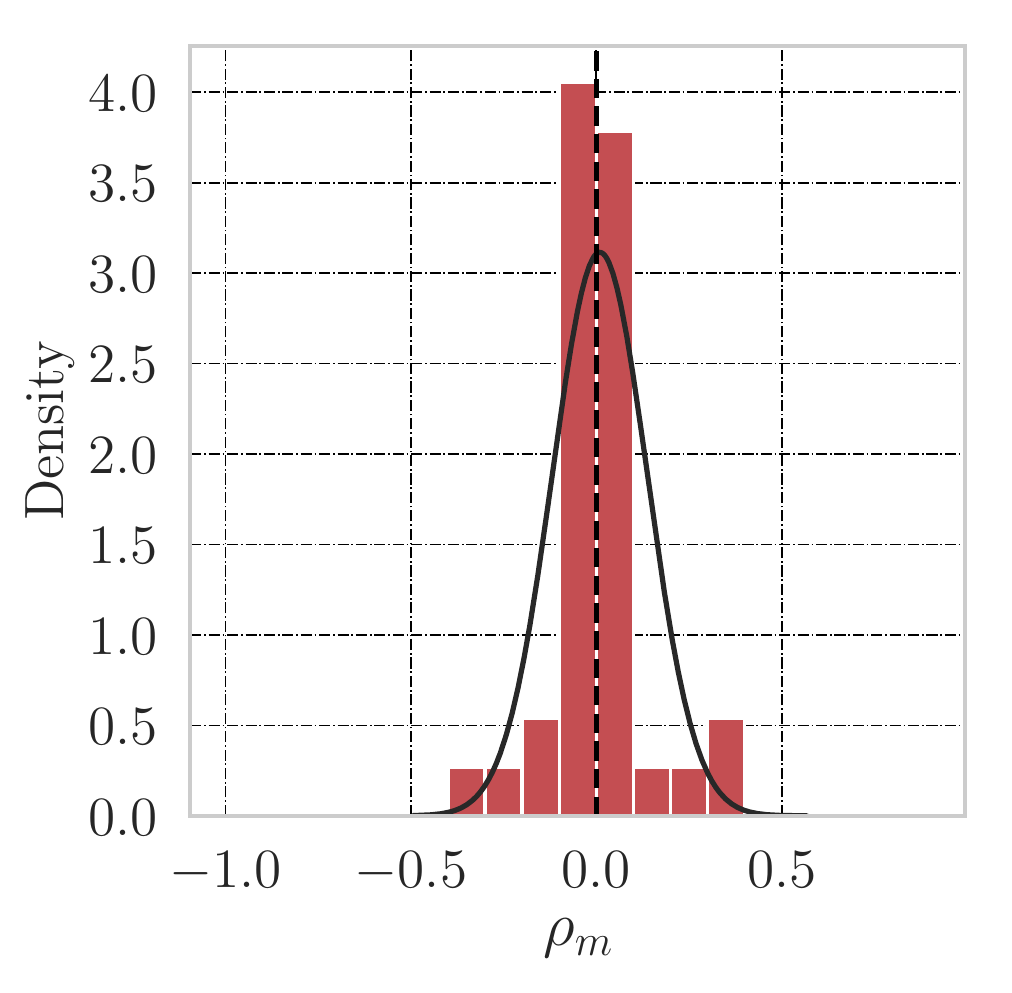}
    \caption{FRAP}
    \label{fig:all-env-normal-dist-of-results-frap}
\end{subfigure}
\caption{These plots illustrate the density of $\rho_m$ over all intersections. Here $\rho_m<0$ means that the multi-env outperforms the baseline algorithm. }
\label{fig:all-env-normal-dist-of-results}
\end{figure}

Next, we compare the multi-env regime with all the benchmark algorithms. To this end, we define $\rho_m = \frac{u_m - b_m}{\max(u_m, b_m)} \in[-1,1], \forall m\in\mathcal{M}$, in which $u_m$ is the ATT of intersection $m$ when the trained model of multi-env is scored greedily (i.e., we let the phase with the highest probability to be the next active phase), and $b_m$ is the ATT of the corresponding intersection of a certain benchmark algorithm. Having $\rho_m<0$ means that the multi-env model outperforms the baseline algorithm. The density plots of  Figure~\ref{fig:all-env-normal-dist-of-results} show the distribution of $\rho_m$ for all $m\in\mathcal{M}$ and for different baseline algorithms.
Figures~\ref{fig:all-env-normal-dist-of-results-fixedtime}-\ref{fig:all-env-normal-dist-of-results-sotl} indicate that multi-env obtains smaller ATT when compared to MaxPressure, FixedTime, and SOTL. Noting that the fitted Normal distribution is centered at a negative value and only a small tail of the distribution lies on the positive side suggests that the multi-env model almost always has a smaller ATT. 
As Figure~\ref{fig:all-env-normal-dist-of-results-dqtscm} suggests, there are few cases where DQTSC-M obtains smaller ATT; however, in average AttendLight multi-env regime achieves smaller ATT than DQTSC-M.
Figure~\ref{fig:all-env-normal-dist-of-results-frap} shows that multi-env AttendLight provides competitive performance with respect to FRAP, which is trained on single-env setting.

In summary, we can conclude that the multi-env algorithm works well in the training set and also obtains quite a small gap in unseen environment instances. So, the model can be used without any retraining on new intersection instances similar to those observed during training.  

\subsection{Few-Shot Training for Calibration}

Although the multi-env regime works well in practice, one might want to get a specialized policy for an important intersection. 
For this purpose, we start with the universal policy and after a few training steps, we obtain an improved policy for that intersection. 
Recall that AttendLight is designed to handle any intersection, so we do not need to modify the structure of the policy network (the number of inputs and outputs).
Following this strategy, the maximum and average of the multi-env gap compared to single-env decreases significantly after 200 training episodes (instead of 100,000 episodes when trained from scratch) such that we got to 5\% gap on average with respect to the single-env regime. After 1000 training steps this gap decreases to 3\%. See Appendix~\ref{apdx:few_shot_multi_env_regime} for more details.

\section{Conclusion and Discussion}

In this paper, we consider the traffic signal control problem, and for the first time, we propose a universal RL model, called AttendLight, which is capable of providing efficient control for any type of intersections. To provide such capability to the model, we propose a framework including two attention mechanisms to make the input and output of the model, independent of the intersection structure. The experimental results on a variety of scenarios verify the effectiveness of the AttendLight. First, we consider the single-environment regime. In this case, AttendLight outperforms existing methods in the literature. Next, we consider AttendLight in the multi-environment regime in which we train it over a set of distinct intersections. The trained model is tested in new intersections verifying the generalizability of AttendLight.

A future line of research could be extending AttendLight to control multiple intersections in a connected network of intersections. Another research direction would be applying different RL algorithms such as Actor-Critic, A2C, and A3C to  improve the numerical results. In a similar approach, considering other embedding functions and attention mechanisms would be of interest. 
In addition, the AttendLight framework is of independent interest and can be applied to a wide range of applications such as \emph{Assemble-to-Order Systems, Dynamic Matching Problem,} and  \emph{Wireless Resource Allocation} problems. Similar to TSCP, each of these problems has to deal with the varying number of inputs and outputs, and AttendLight can be applied with slight modifications.

\newpage
\section*{Broader Impact}
In this paper, the authors propose AttendLight, a Deep Reinforcement Learning algorithm to control the traffic signals efficiently and autonomously. Utilizing AttendLight 
for controlling traffic signals brings several benefits to society. 

First, this algorithm is responsive to the dynamic behavior of traffic movement and provides a control policy to an intersection to minimize the travel time. This has several societal implications: 
\begin{itemize}
    \item {\it Less traffic}:  according to information gathered in \cite{abcnews} in 2015, drivers in the United States wasted 6.9 billion hours annually in traffic. With AttendLight people will spend less time in traffic jams.
    \item \textit{Lower fuel consumption}: it has been shown that about 3 billion gallons of gas wasted in 2014 due to traffic congestion \cite{abcnews}. AttendLight will help to reduce fuel consumption by easing traffic flows.
    \item {\it Cleaner environment}: it is predicted that air pollution will cause around 2.5 million cases of non-communicable disease by 2035, should the air quality stay the same as in 2018 \cite{pimpin2018estimating}. People will breathe higher quality air if we have smarter traffic signal controllers.
\end{itemize}

Second, in contrast to previous RL models, AttendLight does not need to be trained for every new intersection. Thanks to the attention mechanism, indeed AttendLight is a universal model that can be simply deployed for any type of intersection after it is trained over a collection of distinct intersections. As long as the structure for the new intersection follows a similar distribution as the training set, AttendLight provides accurate results. This capability is the key advantage of this model because, designing a new model imposes several costs such as (i)  human expertise, (ii) the required computational power, and (iii) data collection resources. Therefore, sparing experts from repetitive work as well as saving computational resources are other societal impacts of AttendLight. 

One limitation of AttendLight is that it may fail to come up with an efficient policy whenever the intersection topology is very complex and unusual, such that the training set does not involve a similar structure. We hope that such a limitation can be addressed by increasing the training set diversity and incorporating more complex policy models. 

Future researchers are encouraged to consider extending AttendLight to control multiple intersections in a coordinated manner. Given a network of intersections, traffic signals have significant impacts on each other. Thus, controlling every individual signal without considering this incorporation may exacerbate the whole traffic. Reaching the point that AttendLight could control a network of intersections while it respects the association effects, we expect to achieve too many other valuable societal impacts. However, the extension of AttendLight to the multi-intersection scenario is not quite straightforward. Specifically, one needs to consider several challenges such as scalability of the proposed model, and how to involve the coordination in decision-making procedure. See more details in \cite{oroojlooyjadid2019review}.   

Further motivation to pursue this research would be improving the RL algorithm used to train the AttendLight model. In the current research, we utilized a policy-gradient RL algorithm called REINFORCE. Despite the superior
numerical results of AttendLight, there is definitely room for improvement. As a low-hanging fruit, other state-of-the-art policy-based RL algorithms such as Actor-Critic, A2C, and A3C can substitute REINFORCE in AttendLight.

\newpage
\bibliographystyle{spmpsci}
\bibliography{references}


\newpage
\appendix

\section{Numerical Experiments Details}\label{sec:apdx:numerical_experiments_details}
Appendix \ref{sec:apdx:numerical_experiments_details} includes the details of all numerical experiments. In Appendix~\ref{apdx:synthetic_data_details}, we describe our traffic data and how we generate the synthetic data. Appendices~\ref{appsec:rl-model}  and ~\ref{appsec:rl-train} explains the AttendLight model as well as the RL training that we consider throughout the paper. Appendix~\ref{apdx:details_of_baselines} gives a brief explanation on our baseline algorithms. Finally, in Appendix~\ref{apdx:details_results_single_multi_environment}, we provide extensive results of AttendLight for single-env and multi-env regimes.


\subsection{Details of Real-World Data and the Synthetic Data Generation}\label{apdx:synthetic_data_details}
In each intersection, we consider three traffic movement sets, namely straight, turn-left, and turn-right.These sets are used in defining the traffic data of an intersection in both real-world and synthetic cases. Let us reconsider intersection of Figure~\ref{fig:3_way_intersection} and the corresponding traffic movements in Figure \ref{fig:3_way_intersection_roads}. For this intersection, the straight, turn-right, and turn-left sets are $\{\nu_2, \nu_3\}$, $\{\nu_4,\nu_6\}$, and $\{\nu_1,\nu_5\}$, respectively. Next, we elaborate two sources of traffic that we use in our experiments.

{\bf Real-world traffic data}\\
For the real-world data, we use the public data of intersections in Hangzhou and Atlanta \cite{wei2019survey, zheng2019learning}, which are available to download in \cite{hangzhou_data} and \cite{atlanta_data}, respectively. These data-sets include the traffic on 4-way intersections with straight and turn-left traffic movements. For 3-way intersections, we slightly modify the data of 4-way intersections from Hangzhou and Atlanta to adapt them for 3-way structures. In particular, for each vehicle arrival, we keep the vehicle properties and the arrival time, but we change the entering and leaving lanes if they are not available in the new intersection. To do this, we allocate the traffic of unavailable movement to an available one from the same set (e.g., straight or turn-left) uniformly at random. 
For example, consider intersection of Figure~\ref{fig:3_way_intersection}. Compared to a 4-way intersection, this intersection does not have north-to-south and south-to-north traffic movements. So, to use traffic data {\tt A1-A5} and {\tt H1-H5}, we substitute the traffic movements north-to-south and south-to-north with the traffic movements east-to-west and west-to-east. We follow the same process for the left-turn traffic movements.

{\bf Synthetic traffic data}\\
For the intersections where there exist an entering/leaving roads with three-lanes, we could not find any real-world data of a single intersection with the right-turn data. So, we generate traffic-data with a different rate of incoming vehicles. For this purpose, we consider two Poisson arrival processes with a rate $\lambda$ from $\{3, 4\}$ (seconds). Incoming traffic with the probability of 70\%, 20\%, and 10\% is selected to go straight, turn-left, or turn-right, respectively. Inside each of three sets (i.e., straight, turn-left, and turn-right sets), the traffic movements are selected uniformly. For example, consider the intersection of Figure~\ref{fig:3_way_intersection}, and assume a vehicle is chosen to move straight. Then, with the probability of 50\% either $\nu_2$ or $\nu_3$ will be selected. 
In addition, to analyze the performance of each algorithm under the different volumes of traffic, we increase the number of arriving vehicles in three settings: at each arrival time, an additional vehicle is added with a probability $p$ from $\{0.05, 0.1, 0.3\}$. Overall, there are six cases of synthetic traffic data, which are summarized in Table~\ref{tb:synthetic_data_details}. 

\begin{table}[htbp]
    \centering
    \caption{Parameters of the synthetic data. Each parenthesis shows $\lambda$  of the Poisson distribution and the probability of having two vehicle arrival at each time.}
    \label{tb:synthetic_data_details}    
    \begin{tabular}{lcccccc}
    \toprule
        traffic-data & S1 & S2 & S3 & S4 & S5 & S6 \\ \midrule
        ($\lambda, p$) & (4, 0.3) & (4, 0.1) & (3, 0.1) & (3, 0.05) & (3, 0.3) & (4, 0.05) \\
    \bottomrule
    \end{tabular}
\end{table}

\subsection{Details of AttendLight Policy Model}\label{appsec:rl-model}

The AttendLight policy model---visualized in Figure \ref{fig:att_model}---contains five trainable components: embedding layer, {\tt state-attention}, Recurrent Neural Network (RNN), fully-connected layer, and {\tt action-attention}. In this work, we use a 1-dimensional convolution layer with the input-channel of size four, kernel-size and stride of size one, and the out-channel of size $d$. This embedding is an affine transformation that maps all lane characteristics to a $d$-dimensional space for allowing more effective feature representations. The {\tt state-attention} uses linear transformations of query and references with output size of $d$. For the RNN, we use a single layer of LSTM cell with a hidden dimension $d$. The output of the LSTM is passed into a linear layer with output size $d$ followed by a ReLU activation function. The {\tt action-attention} model also uses the same linear transformations for the query and references with $d$ output dimension and returns the probability of selecting each action. To train the single-env models, we set $d=128$ and the multi-env model use $d=256$. We would like to emphasize that we have not done any structured hyper-parameter tuning of AttendLight. For training the AttendLight policy model, we use a policy-gradient based method as explained next. 

\subsection{Details of the RL Training}\label{appsec:rl-train}
We use a variance-reduced variant of the REINFORCE algorithm \cite{sutton2000policy} to train the AttendLight. The details of this algorithm are presented in Algorithm \ref{alg-reinfore}. In REINFORCE, there are two neural networks, called the actor (with parameters $\theta$) and critic (with parameter $\phi$). The actor network is responsible for learning the optimal action, while the critic is used for variance reduction. The critic uses the output of the \texttt{state-attention} which is followed by two fully-connected layers with ReLU activation function and returns the value of being in a state.

We train the single-env models for 100000 episodes and the multi-env models for 30000 episodes. To train these models, we employ the Monte-Carlo simulation using the current policy $\pi_{\theta}$ to produce a valid sequence of state, action, and rewards. Given the sequence of observations, we run a single train-step of the REINFORCE algorithm and update the weights of the networks. We use Adam optimizer with the learning rate of 0.005 for the single-env regime and 0.0005 for the multi-env regime. To stabilize the training of the single-env regime, we consider $n=3$ instances of the intersection in parallel. For this problem, every 1100 episodes take approximately one hour on a single V100 GPU. On the multi-env regime, we choose $n=42$ environment instances (without replacement) and run them simultaneously. Not surprisingly, the multi-env regime takes a longer time such that it accomplishes 85 episodes in one hour on the same GPU.

\begin{algorithm}[h!]
	\caption{REINFORCE Algorithm}
	\label{alg-reinfore}
	\begin{algorithmic}[1]
		\STATE initialize the actor and critic networks with random weights $\theta$ and $\phi$, respectively
		\FOR {$iteration = 1,2,\cdots$} 
		\STATE reset gradients: $d\theta\leftarrow 0$, $d\phi\leftarrow 0$
		\STATE if single-env regime, get $n$ copy of the same environment instances; in multi-env regime choose randomly $n$ environment instances
		\FOR {$i=1,\cdots,n$}
		\STATE initialize step counter $t\leftarrow 0$
		\REPEAT
		\STATE choose {\it next-action} $a_i^t$ according to the distribution $\pi(\cdot|{s}^{t}_i;\theta)$
		\STATE observe new state $s^{t+1}_i$ and reward $r_i^t$ which is the negative of intersection pressure
		\STATE $t \leftarrow t+1$
		\UNTIL{termination condition $t\leq T$ is not satisfied}
		\ENDFOR
		\STATE compute the cumulative reward $R_i^{t}=\sum_{t'=t}^{T}r_i^{t'}$ for all $i=1,\cdots,n$ and $t=1,\cdots,T$
		\STATE $d\theta \leftarrow \frac{1}{nT} \sum_{i=1}^n \sum_{t=1}^T \left(R_i^t - V(s^t_i;\phi)\right) \nabla_\theta \log \pi(a_i^t|s^t_i;\theta)$\label{eq:vrp:alg:policy_update}
		\STATE $d\phi \leftarrow \frac{1}{nT} \sum_{i=1}^n \sum_{t=1}^T \nabla_\phi \left(R_i^t - V(s^t_i;\phi)\right)^2$\label{eq:vrp:alg:value}
		
		\STATE update $\theta$ using $d\theta$ and $\phi$ using $d\phi$.
		\ENDFOR
	\end{algorithmic}
\end{algorithm}

\subsection{Details of Baseline Algorithms}\label{apdx:details_of_baselines}

\subsubsection{SOTL}
Let's define counters $\alpha$ and $\beta$ to count the number of cars behind the phases with the red signal and the number of cars behind the phase with the green signal, respectively. In SOTL algorithm \cite{gershenson2004self}, if the active time of the current phase is more than a threshold $\delta$, and if $\alpha > \texttt{max-red-count}$ the $\beta < \texttt{min-green-count}$, then the active phase switches in a cyclic manner. To obtain the best values for the parameters of the SOTL, we try a grid-search over set \{2, 7, \dots, 62\} for {\tt max-red-count} and {\tt min-green-count}, \{2, 7, \dots, 33\} for $\delta$. This results in 13$\times$13$\times$7=1183 cases, in which we report the minimum ATT obtained among these cases.  

\subsubsection{Fixed-Time}
In the Fixed-Time algorithm, the phases change in a cyclic order. For each phase, an active time is determined in advance, and once the active phase meets that active time, the phase switches to the next one. This is the most common approach in real-world intersections. We set the active time of each phase to 15 seconds. 

\subsubsection{Max-Pressure}
The Max-Pressure algorithm \cite{varaiya2013max} considers the pressure of participating lanes for each phase, and it activates the phase with the highest pressure for the next time-step.

\subsubsection{FRAP}
The get the result of the FRAP algorithm \cite{zheng2019learning}, we run the publicly available code of the algorithm for our intersection instances. 
FRAP has two main limitations which make it inapplicable to some of the intersections that we consider: (i) it assumes that each phase involves exactly two traffic movements; and (ii) it is designed based on the assumption that the number of traffic movements in all phases are the same. The first limitation is due to the implementation of algorithm \cite{frap_code}. We adjusted the implementation by generating dummy traffic movements to handle intersections such as {\tt INT1} and {\tt INT2}. 
However, the second limitation is due to the design of FRAP. This is because, FRAP obtains an embedding of the traffic characteristic ($f_p^\nu$) and the phase-id ($f_p^s$) for phase $p$ by:
\begin{equation}
\begin{array}{c}
    h_p^\nu = \text{RELU}(W^\nu f_p^\nu + b^\nu), \\
    h_p^s = \text{RELU}(W^s f_p^s + b^s),
\end{array}
\end{equation}
in which $W^\nu$, $W^s$, $b^\nu$, and $b^s$ and trainable variables, $h_p^\nu$ and $h_p^s$ are the embedded traffic characteristic and the phase. Then, these are combined with adding another layer of affine transformation to obtain the state definition for each phase:
\begin{equation}
    h_p^\nu = \text{RELU}(W^h [h_p^\nu, h_p^s] + b^h).
\end{equation}
Since $W^\nu$ is shared among all phases, it is not applicable when phases have a different number of lanes. So, it is not possible to adjust the algorithm to work for some of the intersections such as {\tt INT3} and {\tt INT11}. 

\subsubsection{DQTSC-M}
Since there was not any publicly available code of DQTSC-M \cite{shabestary2018deep}, we implemented the algorithm with some slight modifications. We used CityFlow \cite{zhang2019cityflow} to get the state and reward of each time-step. We implemented the DQN \cite{mnih2015human} algorithm as is described in the paper. 
Three convolutional layers with filter sizes 2$\times$4, 2$\times$4, and 2$\times$2, with stride of 1$\times$2, 1$\times$2, and 1$\times$3 are used in the layer one, two, and three, respectively. The output of the convolution layers is passed into a fully connected layer with 128 nodes, followed by another fully connected layer of 64 nodes and the last layer provides the Q-value of each possible action.


\subsection{Extended Comparison Results}\label{apdx:details_results_single_multi_environment}

\begin{figure}[htbp]
	\centering
    {	\includegraphics[trim=5 100 5 110,clip, width=0.75\textwidth]{leg_single_6.pdf}\vspace{-12pt}
}   
	\begin{subfigure}{0.31\textwidth}
		\centerline{ \includegraphics[width=\textwidth]{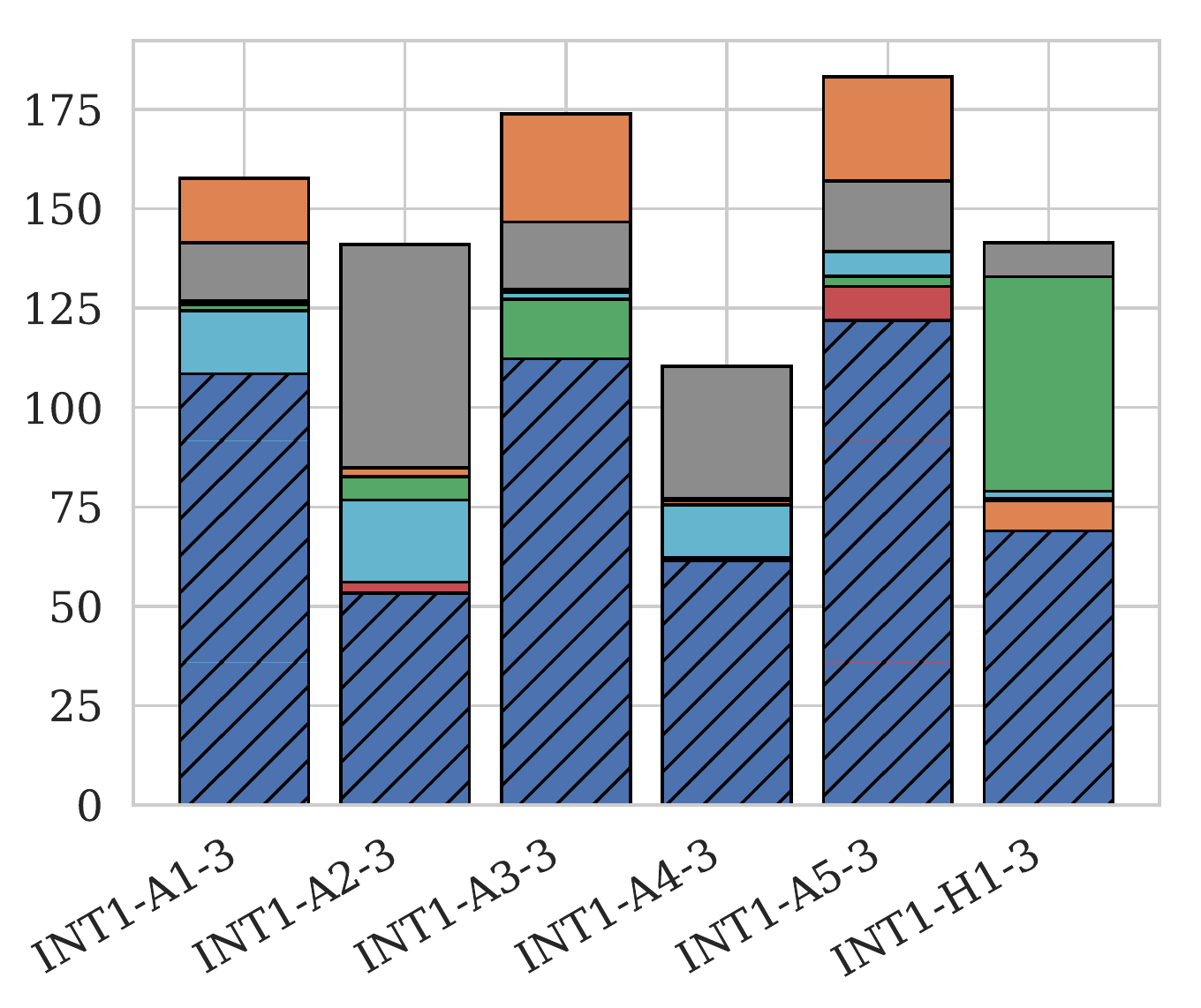}}
		\vspace{-5pt}
		\caption{INT1, 3-phase} \label{fig:result-INT1}
	\end{subfigure}
	\begin{subfigure}{0.31\textwidth}
		\centerline{ \includegraphics[width=\textwidth]{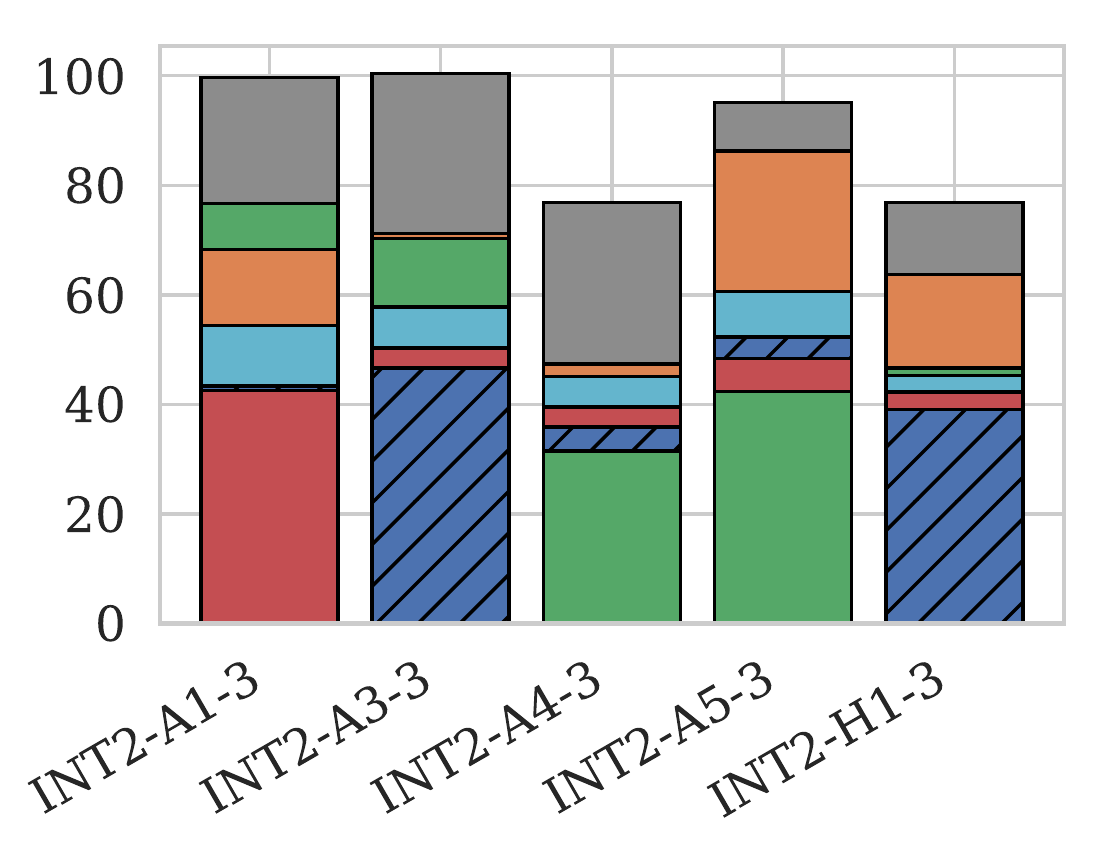}}
		\vspace{-5pt}
		\caption{INT2, 3-phase} \label{fig:result-int2}
	\end{subfigure}
	\begin{subfigure}{0.31\textwidth}
		\centerline{ \includegraphics[width=\textwidth]{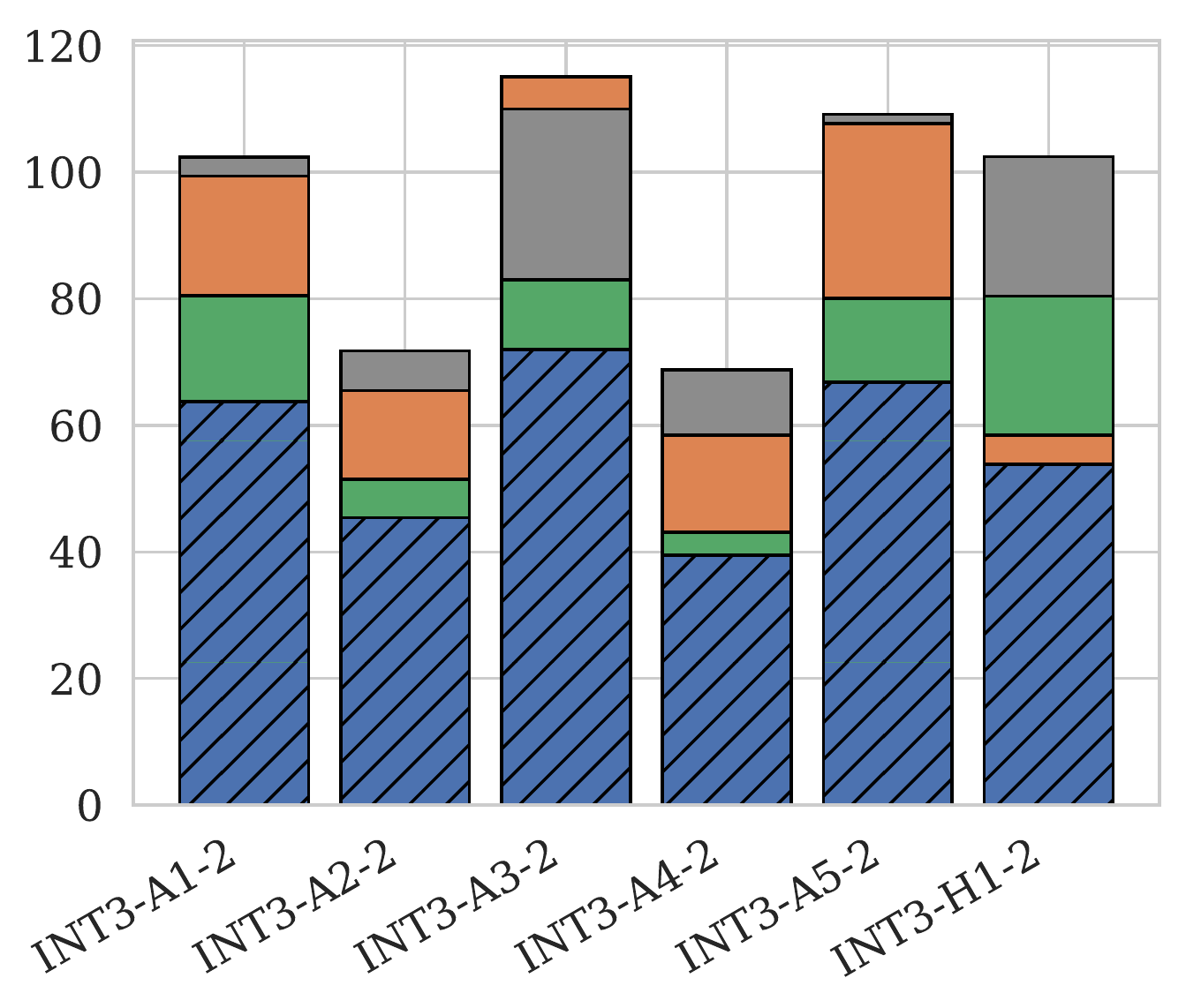}}
		\vspace{-5pt}
		\caption{INT3, 2-phase} \label{fig:result-int3} 
	\end{subfigure}

	\begin{subfigure}{0.31\textwidth}
		\centerline{ \includegraphics[width=\textwidth]{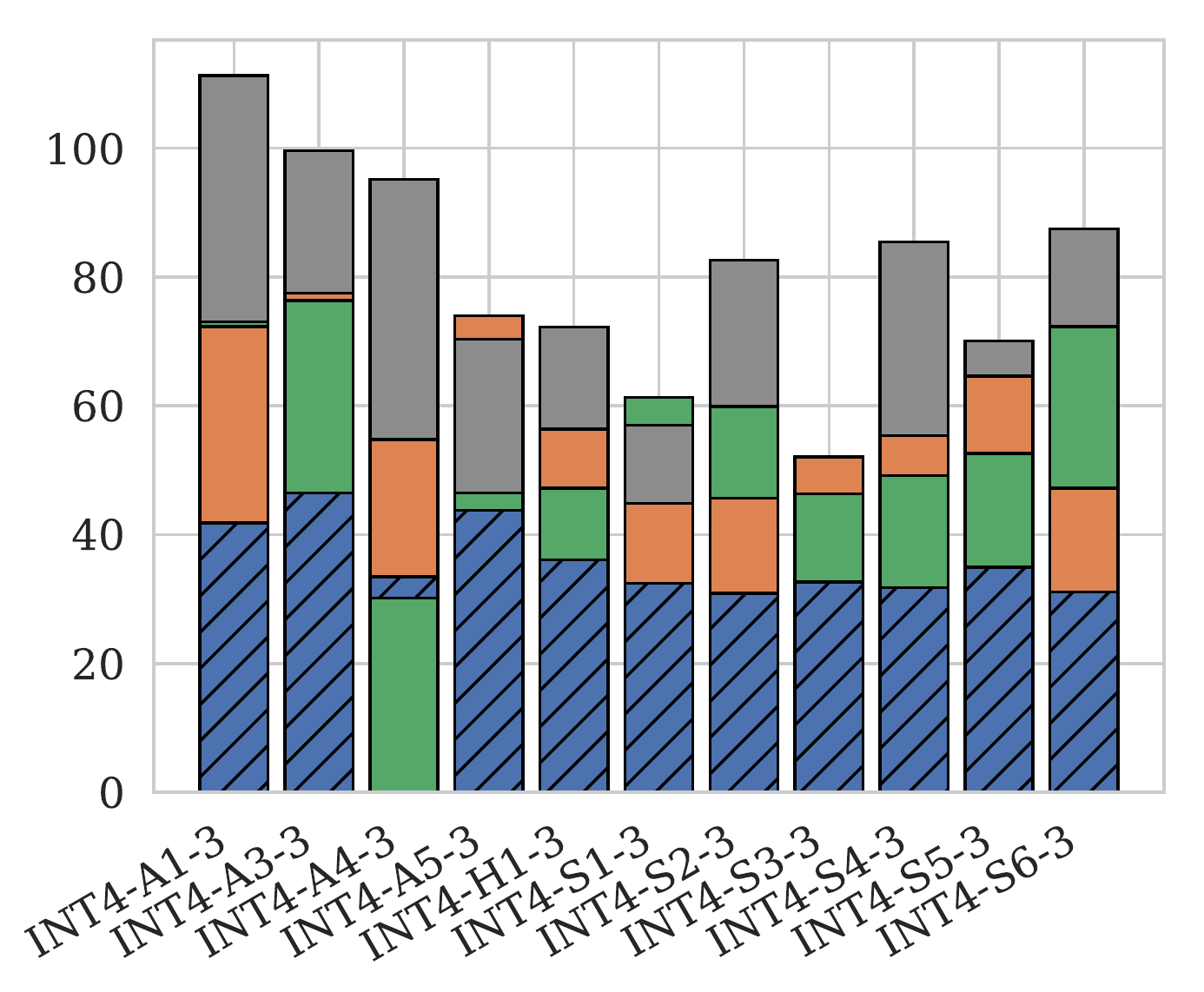}}
		\vspace{-5pt}
		\caption{INT4, 3-phase} \label{fig:result-int4} 
	\end{subfigure}
	\begin{subfigure}{0.31\textwidth}
		\centerline{ \includegraphics[width=\textwidth]{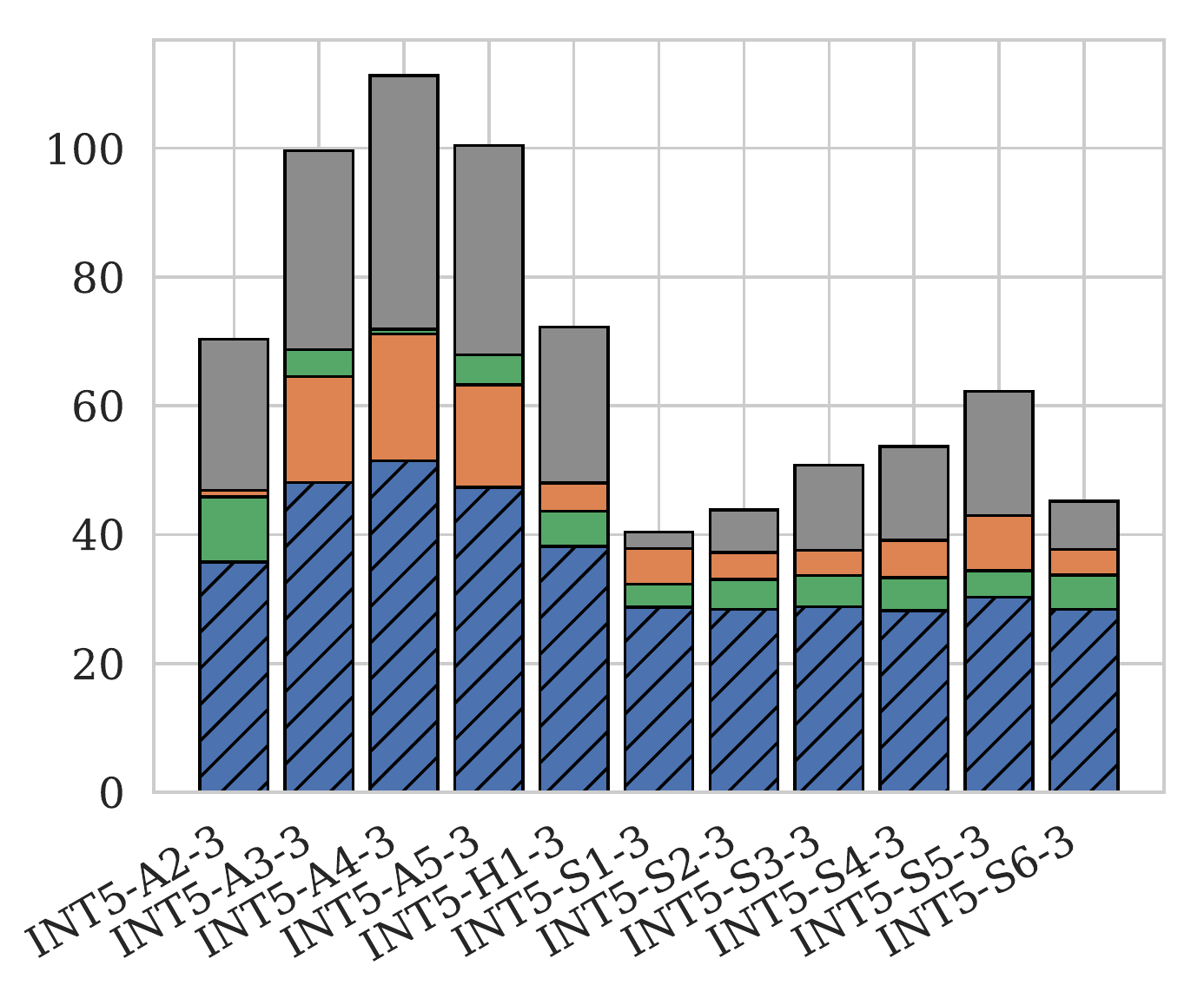}}
		\vspace{-5pt}
		\caption{INT5, 3-phase} \label{fig:result-int5} 
	\end{subfigure}
	\begin{subfigure}{0.31\textwidth}
		\centerline{ \includegraphics[width=\textwidth]{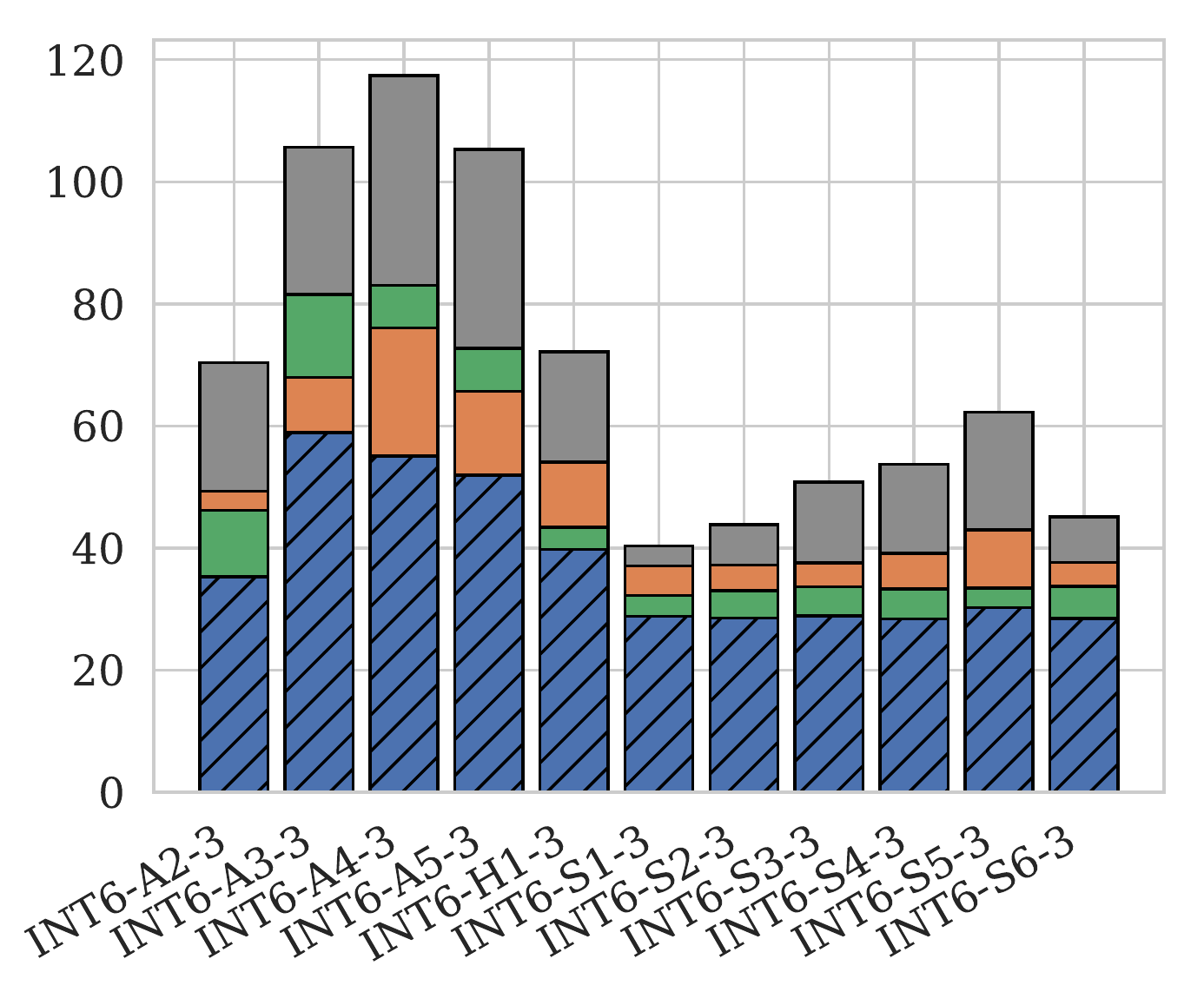}}
		\vspace{-5pt}
		\caption{INT6, 3-phase} \label{fig:result-int6} 
	\end{subfigure}

	\begin{subfigure}{0.31\textwidth}
		\centerline{ \includegraphics[width=\textwidth]{INT7-4p.pdf}}
		\vspace{-5pt}
		\caption{INT7, 4-phase} \label{fig:result-int7-4p} 
	\end{subfigure}
	\begin{subfigure}{0.31\textwidth}
		\centerline{ \includegraphics[width=\textwidth]{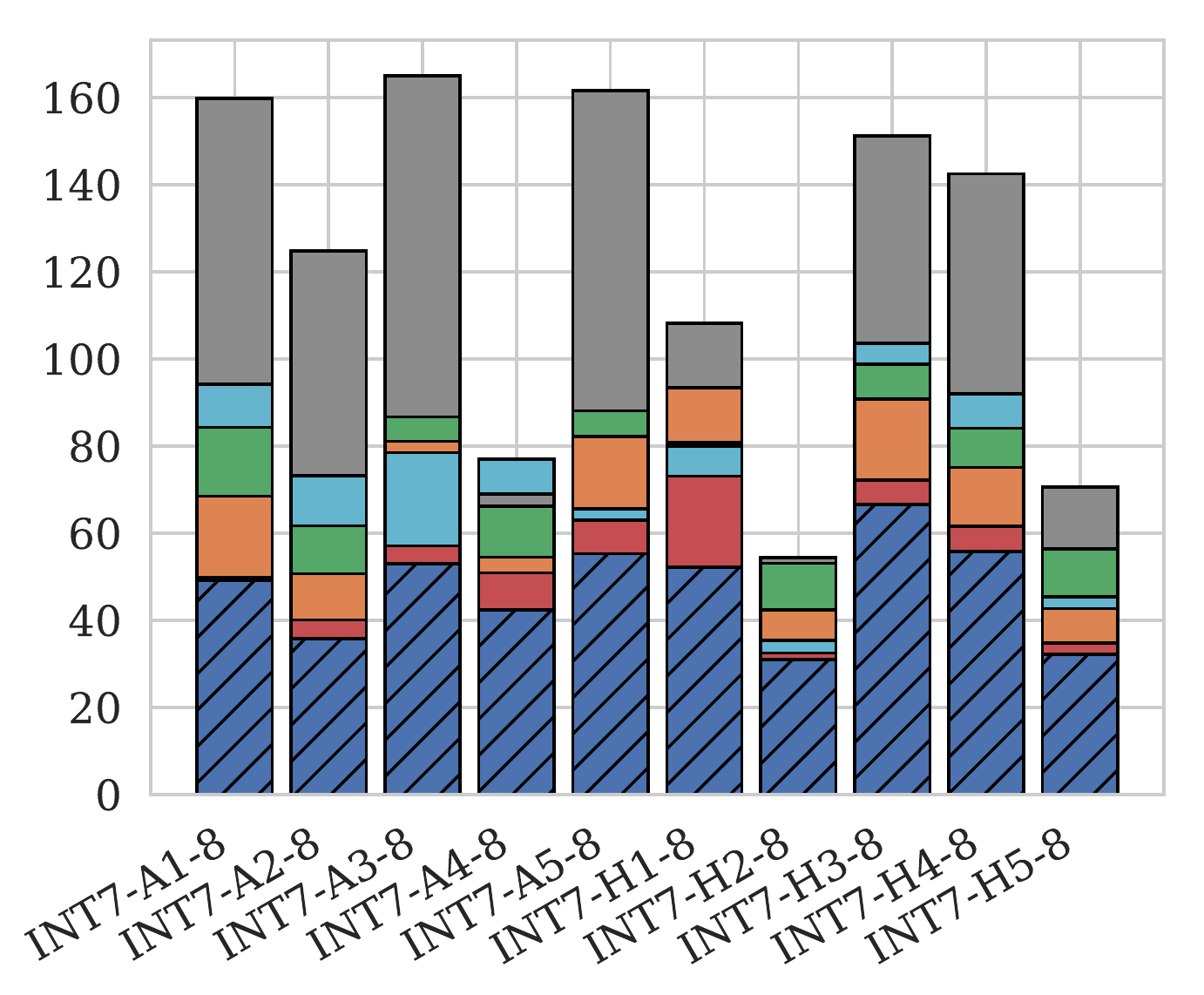}}
		\vspace{-5pt}
		\caption{INT7, 8-phase} \label{fig:result-int-4-8p} 
	\end{subfigure}
	\begin{subfigure}{0.31\textwidth}
		\centerline{ \includegraphics[width=\textwidth]{INT8.pdf}}
		\vspace{-5pt}
		\caption{INT8, 3-phase} \label{fig:result-int8} 
	\end{subfigure}

	\begin{subfigure}{0.31\textwidth}
		\centerline{ \includegraphics[width=\textwidth]{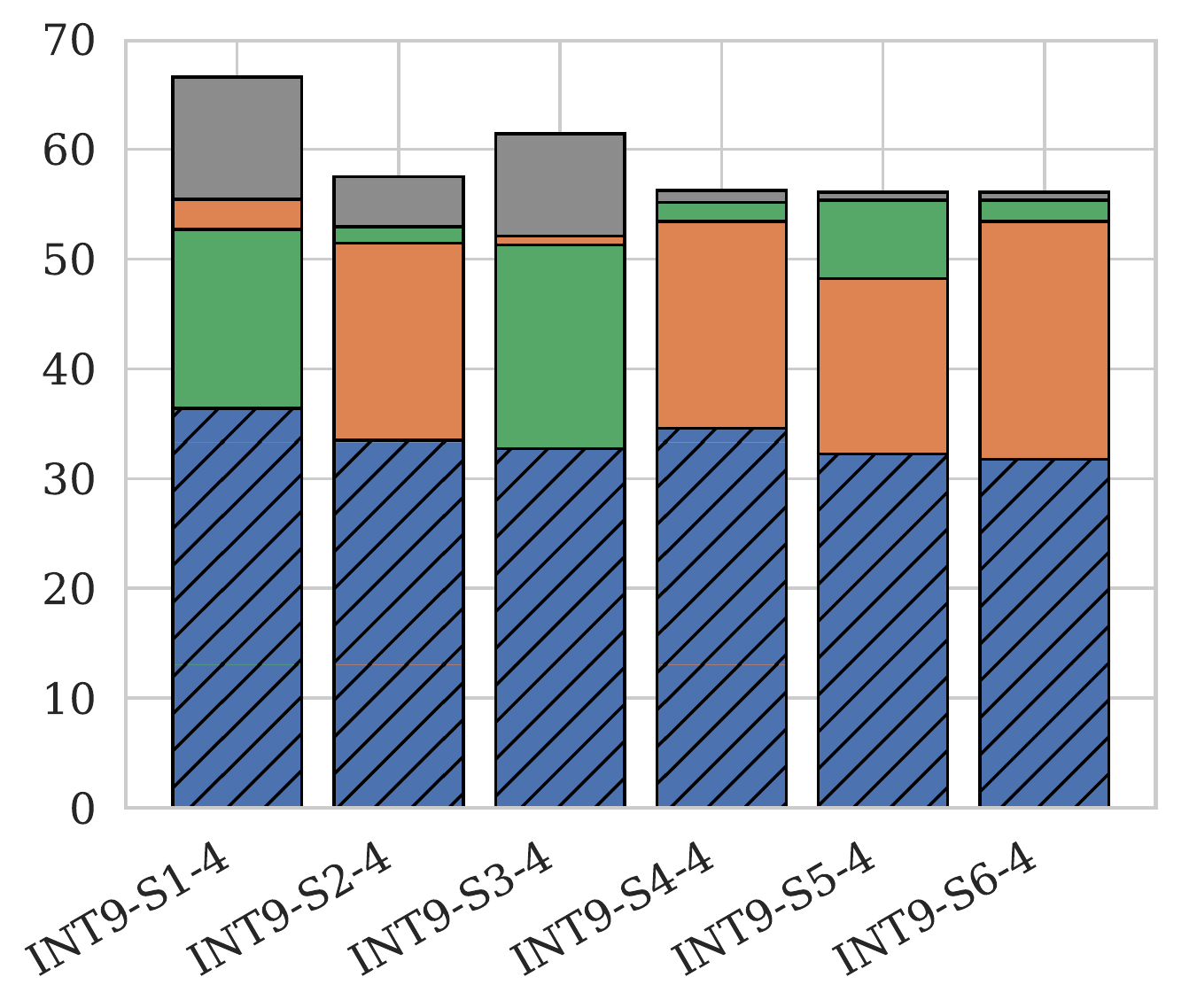}}
		\vspace{-5pt}
		\caption{INT9, 4-phase} \label{fig:result-int9-4p} 
	\end{subfigure}
	\begin{subfigure}{0.31\textwidth}
		\centerline{ \includegraphics[width=\textwidth]{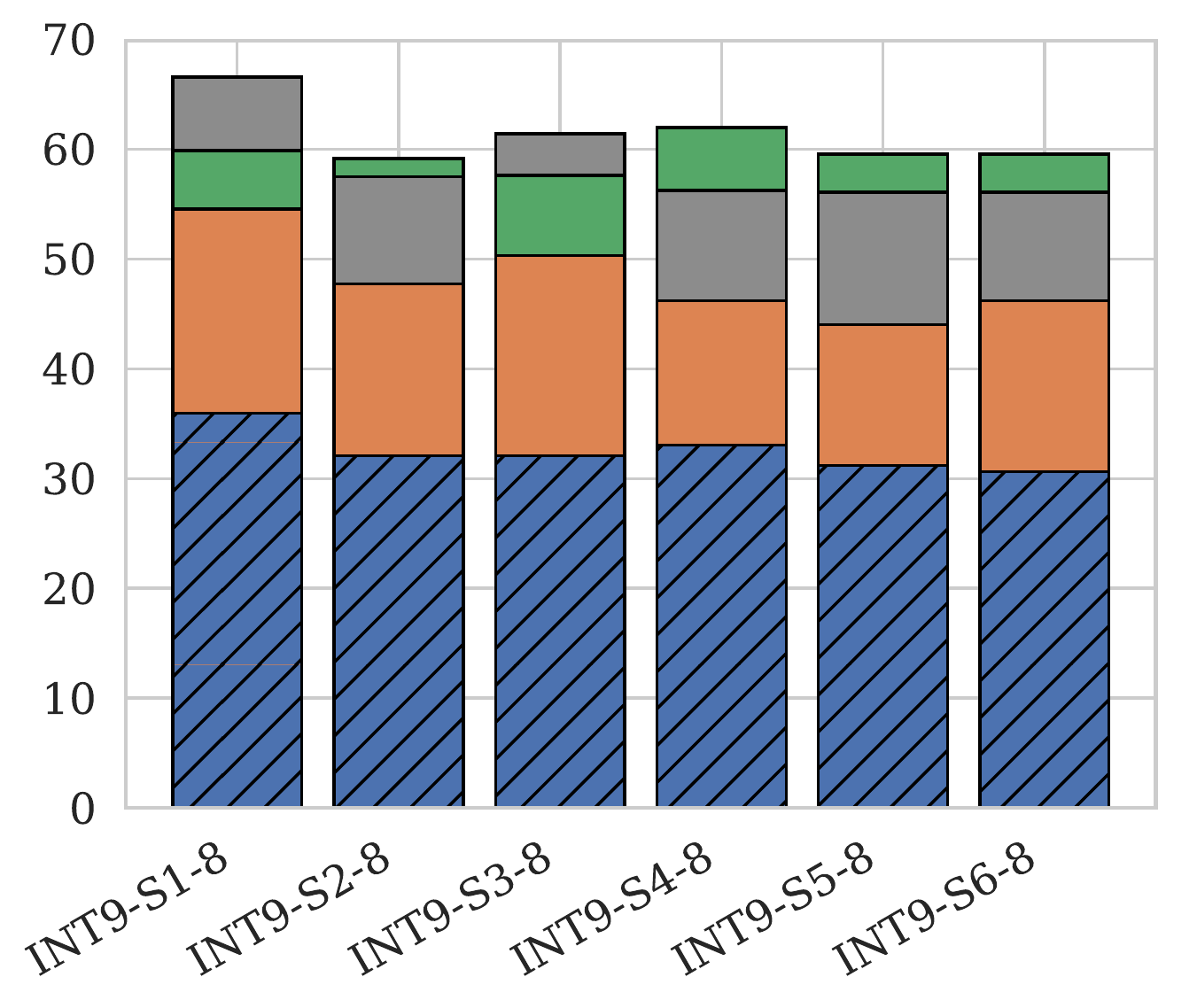}}
		\vspace{-5pt}
		\caption{INT9, 8-phase} \label{fig:result-int9-8p} 
	\end{subfigure}
	\begin{subfigure}{0.31\textwidth}
		\centerline{ \includegraphics[width=\textwidth]{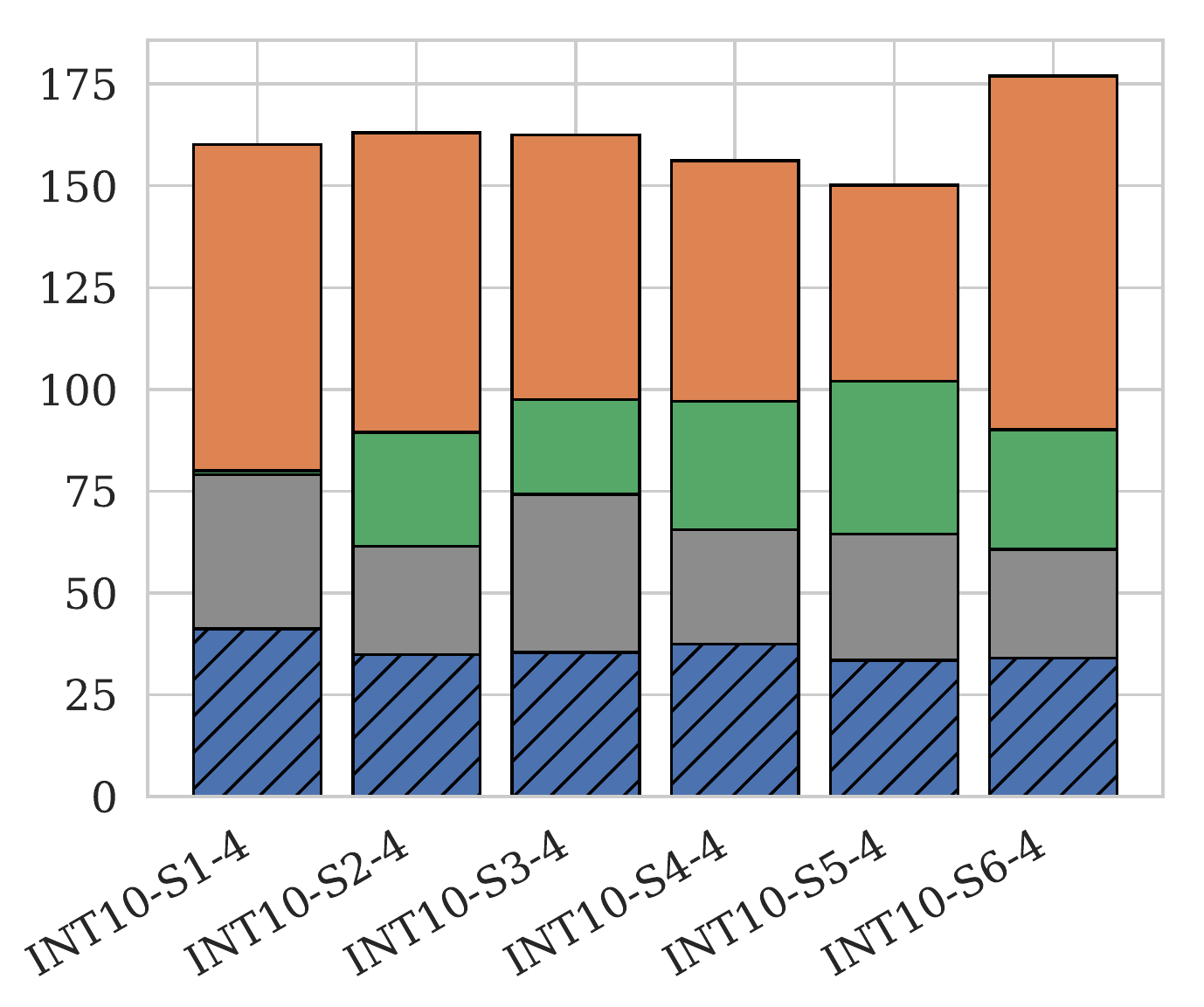}}
		\vspace{-5pt}
		\caption{INT10, 4-phase} \label{fig:result-int10-4p} 
	\end{subfigure}

	\begin{subfigure}{0.31\textwidth}
		\centerline{ \includegraphics[width=\textwidth]{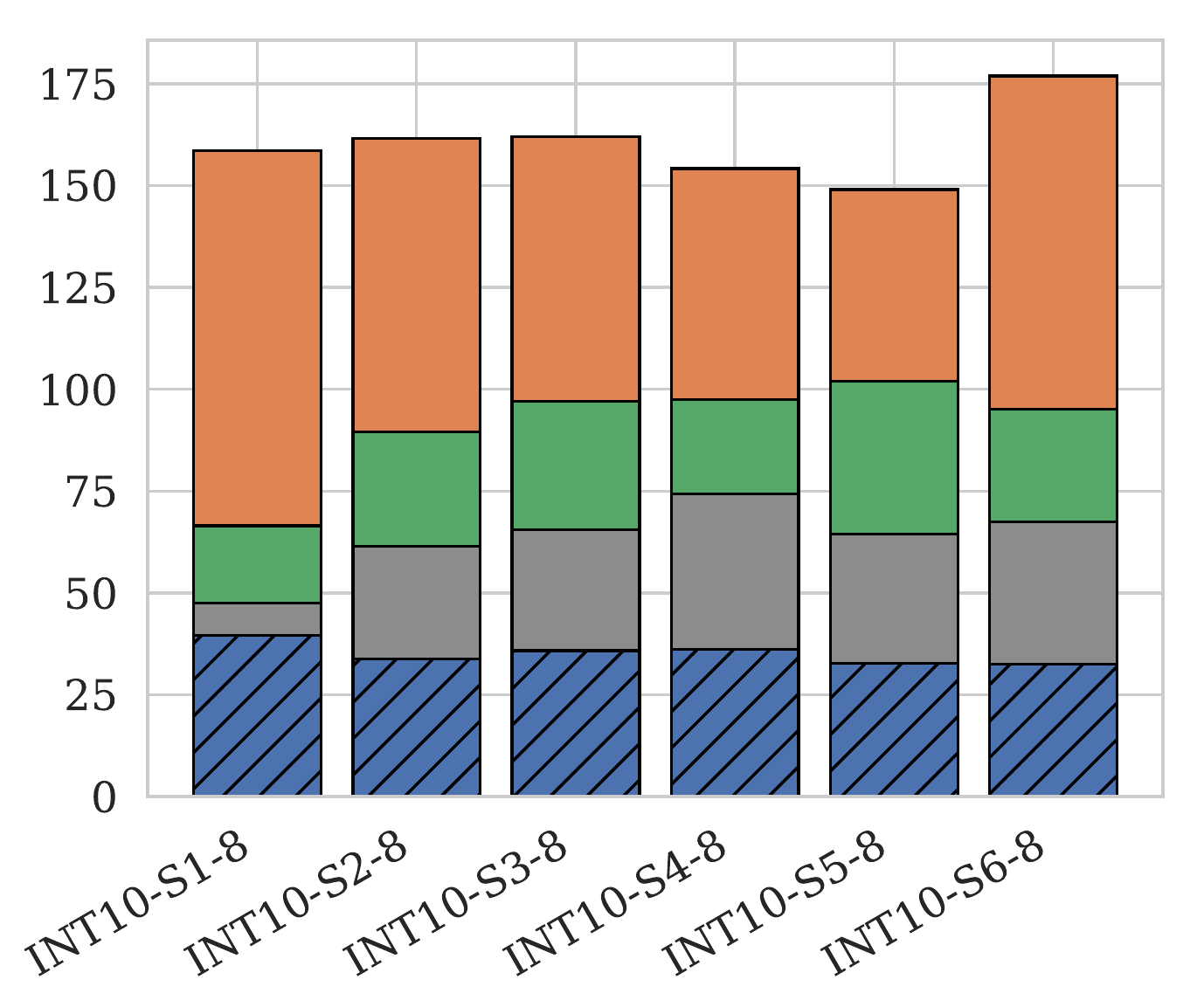}}
		\vspace{-5pt}
		\caption{INT10, 8-phase} \label{fig:result-int7-8p} 
	\end{subfigure}
	\begin{subfigure}{0.31\textwidth}
		\centerline{ \includegraphics[width=\textwidth]{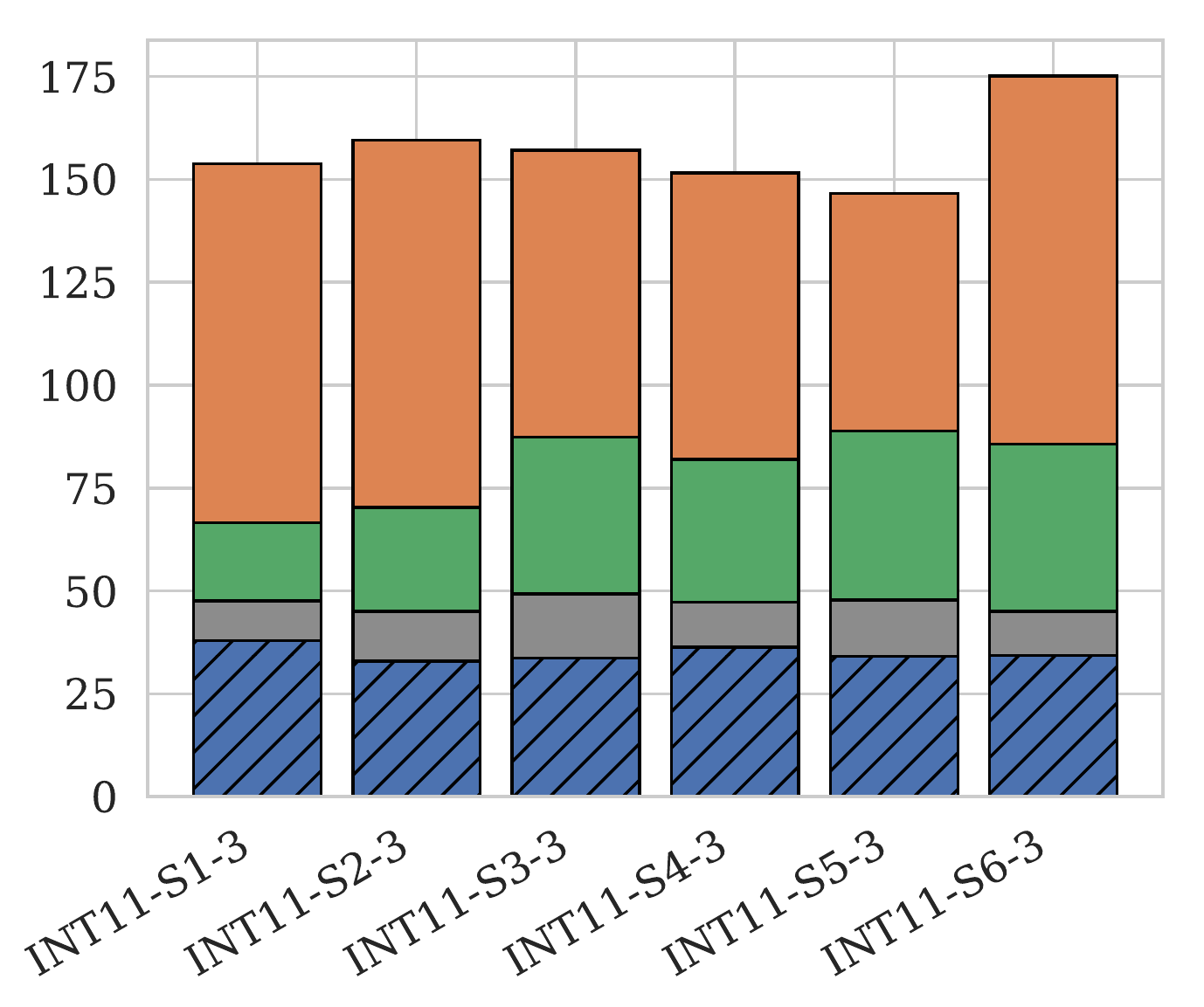}}
		\vspace{-5pt}
		\caption{INT11, 3-phase} \label{fig:result-int11-3p} 
	\end{subfigure}
	\begin{subfigure}{0.31\textwidth}
		\centerline{ \includegraphics[width=\textwidth]{INT11-5p.pdf}}
		\vspace{-5pt}
		\caption{INT11, 5-phase} \label{fig:result-int11-5p} 
	\end{subfigure}

	\caption{The ATT comparison result of all baseline algorithms with AttendLight in single-env regime. Note that FRAP is not applicable to some of the intersections. }
	\label{fig:apdx_result-single_env}
\end{figure}

Tables~\ref{tb:all_results_int1_int5}, \ref{tb:all_results_int6_int9}, and \ref{tb:all_results_int10_int11} show the average travel time (ATT) for all environment instances for FixedTime, MaxPressure, SOTL, FRAP, and AttendLight in single-env and multi-env regimes. When FRAP is not applicable, we report ``NA'' in these tables. As it is shown, AttendLight in single-env regime outperforms other algorithms in 107 cases out of 112 cases and it obtains 46\%, 39\%, 34\%, 16\%, and 9\% average ATT improvement over FixedTime, MaxPressure, SOTL, DQTSC-M, and FRAP, respectively. In addition, Figure~\ref{fig:apdx_result-single_env} illustrates the ATT comparisons results for all intersections. Furthermore, the result of the multi-env regime is also reported under which AttendLight outperforms other benchmarks in 57 cases and it achieves 39\%, 32\%, 26\%, and 5\% improvement over FixedTime, MaxPressure, SOTL, and DQTSC-M, respectively. In comparison to FRAP, AttendLight in multi-env obtains 3\% larger ATT in average.

\begin{table}[htbp]
\caption{Results of all algorithms for {\tt INT1-INT5}}
\label{tb:all_results_int1_int5}
\centering
\begin{adjustbox}{width=.85\textwidth}
\begin{tabular}{lccccccc} 
\toprule
case       & FixedTime & MaxPressure & SOTL   & FRAP  & DQTSC-M & \begin{tabular}[c]{@{}l@{}}AttendLight\\  single-env\end{tabular}  & \begin{tabular}[c]{@{}l@{}}AttendLight\\ multi-env \end{tabular} \\ \midrule 
\multicolumn{8}{l}{\bf \tt INT1, 3 phases} \\ \midrule
{\tt INT1-A1-3}  & 141.44 & 157.59 & 125.93  & 126.73 & 124.34 & \textbf{108.47} & 122.61 \\
{\tt INT1-A2-3}  & 140.96 & 84.86  & 82.65   & 56.12  & 76.74 & \textbf{53.27}  & 62.25  \\
{\tt INT1-A3-3}  & 146.65 & 173.89 & 127.2   & 129.66 & 129.00 & \textbf{112.23} & 124.87 \\
{\tt INT1-A4-3}  & 110.4  & 76.51  & 77.18   & 62.28  & 75.52 & \textbf{61.53}  & 67.69  \\
{\tt INT1-A5-3}  & 157.02 & 183.15 & 132.98  & 130.41 & 139.23 & \textbf{121.94} & 129.79 \\
{\tt INT1-H1-3}  & 141.44 & 76.51  & 132.88  & 77.04  & 77.04 & \textbf{68.92}  & 122.61 \\  \midrule 
\multicolumn{8}{l}{\bf \tt INT2, 3 phases} \\ \midrule
{\tt INT2-A1-3}  & 99.64  & 68.3   & 76.66  & \textbf{42.48} & 54.40  & 43.35  & 68.32  \\
{\tt INT2-A3-3}  & 100.44 & 71.16  & 70.27  & 50.27 & 57.78 & \textbf{46.67}  & 66.46  \\
{\tt INT2-A4-3}  & 76.86  & 47.36  & \textbf{31.44}  & 39.49 & 45.08  & 35.87  & 30.32  \\
{\tt INT2-A5-3}  & 95.14  & 86.26  & \textbf{42.29}  & 48.32 & 60.58 & 52.29  & 43.19  \\
{\tt INT2-H1-3}  & 76.86  & 63.67  & 46.63  & 42.24  & 45.23 & \textbf{39.04}  & 44.07  \\ \midrule 
\multicolumn{8}{l}{\bf \tt INT3, 3 phases} \\ \midrule
{\tt INT3-A1-2}  & 102.39 & 99.34  & 80.49   & NA & 89.49 & \textbf{63.75}    & 81.39  \\
{\tt INT3-A2-2}  & 71.76  & 65.47  & 51.46   & NA & 60.20 & \textbf{45.37}    & 51.21  \\
{\tt INT3-A3-2}  & 109.97 & 115.05 & 82.97   & NA & 86.78 & \textbf{71.99}    & 81.81  \\
{\tt INT3-A4-2}  & 68.78  & 58.44  & 43.09   & NA & 48.06 & \textbf{39.45}    & 43.42  \\
{\tt INT3-A5-2}  & 109.14 & 107.65 & 80.06   & NA & 85.10 & \textbf{66.83}    & 83.42  \\
{\tt INT3-H1-2}  & 102.41 & 58.44  & 80.41   & NA & 62.35 & \textbf{53.81}    & 81.24  \\ \midrule 
\multicolumn{8}{l}{\bf \tt INT4, 3 phases} \\ \midrule
{\tt INT4-A1-3}  & 111.27 & 72.33  & 73.04   & NA & 49.75 & \textbf{41.81}  & 59.36  \\
{\tt INT4-A3-3}  & 99.64  & 77.47  & 76.35   & NA  & 57.56 & \textbf{46.46}  & 53.64  \\
{\tt INT4-A4-3}  & 95.14  & 54.76  & \textbf{30.19}   & NA & 45.14 & 33.47  & 30.58  \\
{\tt INT4-A5-3}  & 70.38  & 73.99  & 46.47   & NA  & 59.17 & 43.75  & \textbf{41.34}  \\
{\tt INT4-H1-3}  & 72.24  & 56.4   & 47.23   & NA  & 46.12 & \textbf{36.06}  & 47.17  \\
{\tt INT4-S1-3}  & 56.96  & 44.88  & 61.32   & NA & 35.67 & \textbf{32.42}  & 36.23  \\
{\tt INT4-S2-3}  & 82.61  & 45.69  & 59.91   & NA & 32.74 & \textbf{30.9}   & 33.93  \\
{\tt INT4-S3-3}  & 52.13  & 52.06  & 46.37   & NA & 34.53 & \textbf{32.68}  & 35.78  \\
{\tt INT4-S4-3}  & 85.47  & 55.4   & 49.2    & NA & 34.48 & \textbf{31.8}   & 37.04  \\
{\tt INT4-S5-3}  & 70.09  & 64.61  & 52.64   & NA & 38.18 & \textbf{34.96}  & 40.1   \\
{\tt INT4-S6-3}  & 87.47  & 47.19  & 72.3    & NA & 33.06 & \textbf{31.07}  & 33.68  \\ \midrule 
\multicolumn{8}{l}{\bf \tt INT5, 3 phases} \\ \midrule
{\tt INT5-A2-3}  & 70.38  & 46.88  & 45.88  & NA & 46.25 & 35.76   & \textbf{29.35}  \\
{\tt INT5-A3-3}  & 99.64  & 64.57  & 68.75  & NA & 56.21 & \textbf{48.11}   & 63.64  \\
{\tt INT5-A4-3}  & 111.27 & 71.13  & 71.87  & NA & 62.26 & 51.46   & \textbf{39.67}  \\
{\tt INT5-A5-3}  & 100.44 & 63.28  & 67.94  & NA & 49.26 & \textbf{47.38}   & 70.52  \\
{\tt INT5-H1-3}  & 72.24  & 48.04  & 43.63  & NA & 46.61 & \textbf{38.2}    & 40.75  \\
{\tt INT5-S1-3}  & 40.42  & 37.84  & 32.31  & NA & 29.11 & \textbf{28.74}   & 29.62  \\
{\tt INT5-S2-3}  & 43.83  & 37.23  & 33.07  & NA & 29.46 & \textbf{28.41}   & 29.96  \\
{\tt INT5-S3-3}  & 50.79  & 37.57  & 33.68  & NA & 30.60 & \textbf{28.82}   & 29.75  \\
{\tt INT5-S4-3}  & 53.71  & 39.12  & 33.34  & NA & 30.52 & \textbf{28.21}   & 29.46  \\
{\tt INT5-S5-3}  & 62.29  & 43.00     & 34.43  & NA  & 31.43 & \textbf{30.27}   & 31.52  \\
{\tt INT5-S6-3}  & 45.18  & 37.69  & 33.75  & NA  & 29.14 & \textbf{28.37}   & 29.3 \\ \bottomrule
\end{tabular}
\end{adjustbox}
\end{table}

\begin{table}[htbp]
\caption{Results of all algorithms for {\tt INT6-INT9}}
\label{tb:all_results_int6_int9}
\centering
\begin{adjustbox}{width=.85\textwidth}
\begin{tabular}{lccccccc} 
\toprule
case       & FixedTime & MaxPressure & SOTL & FRAP & DQTSC-M & \begin{tabular}[c]{@{}l@{}}AttendLight\\  single-env\end{tabular}  & \begin{tabular}[c]{@{}l@{}}AttendLight\\ multi-env \end{tabular} \\ \midrule 
\multicolumn{8}{l}{\bf \tt INT6, 3 phases} \\ \midrule
{\tt INT6-A2-3}  & 70.38  & 49.33  & 46.2    & NA & 42.97 & 35.29    & \textbf{29.01}  \\
{\tt INT6-A3-3}  & 105.64 & 67.97  & 81.53   & NA & 60.91 & \textbf{58.94}    & 70.13  \\
{\tt INT6-A4-3}  & 117.38 & 76.04  & 83.04   & NA & 68.77 & 55.07    & \textbf{38.69}  \\
{\tt INT6-A5-3}  & 105.27 & 65.66  & 72.73   & NA & 59.08 & \textbf{51.92}    & 76.29  \\
{\tt INT6-H1-3}  & 72.18  & 54.11  & 43.43   & NA & 41.70 & 39.77    & \textbf{39.57}  \\
{\tt INT6-S1-3}  & 40.36  & 37.06  & 32.27   & NA & 29.21 & \textbf{28.85}    & 29.66  \\
{\tt INT6-S2-3}  & 43.82  & 37.22  & 33.02   & NA & 29.60 & \textbf{28.51}    & 29.58  \\
{\tt INT6-S3-3}  & 50.78  & 37.54  & 33.64   & NA & 31.25 & \textbf{28.88}    & 29.44  \\
{\tt INT6-S4-3}  & 53.71  & 39.11  & 33.32   & NA & 30.42 & \textbf{28.39}    & 29.27  \\
{\tt INT6-S5-3}  & 62.27  & 42.95  & 33.46   & NA & 32.78 & \textbf{30.26}    & 31.5   \\
{\tt INT6-S6-3}  & 45.14  & 37.64  & 33.71   & NA & 29.18 & \textbf{28.47}    & 29.07  \\ \midrule 
\multicolumn{8}{l}{\bf \tt INT7, 4 phases} \\ \midrule
{\tt INT7-A1-4}  & 159.82 & 58.76  & 76.05  & 50.6  & 65.62 & \textbf{48.56}  & 51.98  \\
{\tt INT7-A2-4}  & 124.78 & 84.57  & 96.34  & 74.24 & 98.75 & \textbf{65.01}  & 69.12  \\
{\tt INT7-A3-4}  & 164.97 & 66.26  & 79.49  & 57.49 & 80.13 & \textbf{53.91}  & 59.11  \\
{\tt INT7-H4-4}  & 142.46 & 77.39  & 80.00  & 67.28 & 85.26 & \textbf{60.47}  & 70.55  \\
{\tt INT7-A4-4}  & 68.97  & 62.86  & 65.41  & 51.61 & 72.32 & \textbf{45.18}  & 50.07  \\
{\tt INT7-A5-4}  & 161.59 & 73.72  & 83.49  & 63.72 & 83.87 & \textbf{57.53}  & 62.32  \\
{\tt INT7-H5-4}  & 70.59  & 52.78  & 53.95  & 38.81 & 48.40 & \textbf{34.38}  & 37.43  \\
{\tt INT7-H1-4}  & 108.17 & 81.77  & 72.9   & 66.32 & 77.21 & \textbf{57.57}  & 61.82  \\
{\tt INT7-H3-4}  & 151.16 & 88.88  & 82.58  & 76.9  & 88.10 & \textbf{69.01}  & 77.5   \\
{\tt INT7-H2-4}  & 54.43  & 45.01  & 50.13  & 33.81 & 36.99 & \textbf{32.26}  & 35.04  \\ \midrule \multicolumn{8}{l}{\bf \tt INT7, 8 phases} \\ \midrule
{\tt INT7-A1-8}  & 159.82 & 68.47  & 84.35  & 49.82 & 94.24 & \textbf{49.11}  & 59.55  \\
{\tt INT7-A2-8}  & 124.78 & 50.72  & 61.74  & 40.14 & 73.17 & \textbf{35.81}  & 40.57  \\
{\tt INT7-A3-8}  & 164.97 & 81.06  & 86.75  & 57.11 & 78.46 & \textbf{53.03}  & 58.12  \\
{\tt INT7-A4-8}  & 68.97  & 54.46  & 66.25  & 50.94 & 76.97 & \textbf{42.4}   & 50.22  \\
{\tt INT7-A5-8}  & 161.59 & 82.23  & 88.12  & 62.96 & 65.63 & \textbf{55.31}  & 62.23  \\
{\tt INT7-H1-8}  & 108.17 & 93.37  & 80.83  & 73.14 & 80.04 & \textbf{52.25}  & 62.41  \\
{\tt INT7-H2-8}  & 54.43  & 42.37  & 53.09  & 32.56 & 35.42 & \textbf{31.03}  & 34.03  \\
{\tt INT7-H3-8}  & 151.16 & 90.83  & 98.84  & 72.18 & 103.61 & \textbf{66.59}  & 77.7   \\
{\tt INT7-H4-8}  & 142.46 & 75.08  & 84.15  & 61.59 & 91.98 & \textbf{55.8}   & 62.2   \\
{\tt INT7-H5-8}  & 70.59  & 42.66  & 56.4   & 34.77 & 45.43 & \textbf{32.19}  & 35.7   \\ \midrule 
\multicolumn{8}{l}{\bf \tt INT8, 3 phases} \\ \midrule
{\tt INT8-A1-3}  & 101.44 & 73.55  & 70.00  & 53.34 & 66.48 & \textbf{51.09}   & 53.09  \\
{\tt INT8-A2-3}  & 66.1   & 45.87  & 46.53  & 42.00 & 45.53 & \textbf{37.19}   & 40.84  \\
{\tt INT8-A3-3}  & 106.93 & 75.53  & 75.26  & 51.00 & 68.91 & \textbf{46.99}   & 49.37  \\
{\tt INT8-A4-3}  & 66.48  & 48.77  & 49.7   & 45.34 & 50.20 & \textbf{39.2}    & 41.57  \\
{\tt INT8-A5-3}  & 101.92 & 68.15  & 74.68  & 58.13 & 57.01 & \textbf{48.97}   & 52.7   \\
{\tt INT8-H1-3}  & 101.92 & 48.77  & 74.68  & 76.33 & 79.21 & \textbf{57.54}   & 53.09  \\ \midrule 
\multicolumn{8}{l}{\bf \tt INT9, 4 phases} \\ \midrule
{\tt INT9-S1-4}  & 66.56  & 55.42  & 52.69   & NA & 41.69 & \textbf{36.4}    & 40.72  \\
{\tt INT9-S2-4}  & 57.51  & 51.45  & 52.94   & NA & 36.73 & \textbf{33.49}   & 35.34  \\
{\tt INT9-S3-4}  & 61.41  & 52.08  & 51.28   & NA & 39.84 & \textbf{32.72}   & 35.38  \\
{\tt INT9-S4-4}  & 56.24  & 53.4   & 55.16   & NA & 39.26 & \textbf{34.56}   & 36.42  \\
{\tt INT9-S5-4}  & 56.09  & 48.21  & 55.37   & NA & 35.66 & \textbf{32.24}   & 34.16  \\
{\tt INT9-S6-4}  & 56.09  & 53.4   & 55.37   & NA & 35.17 & \textbf{31.77}   & 40.72  \\ \midrule 
\multicolumn{8}{l}{\bf \tt INT9, 8 phases} \\ \midrule
{\tt INT9-S1-8}  & 66.56  & 54.54  & 59.87   & NA & 52.70 & \textbf{35.95}  & 39.87  \\
{\tt INT9-S2-8}  & 57.51  & 47.71  & 59.15   & NA & 34.26 & \textbf{32.08}  & 33.55  \\
{\tt INT9-S3-8}  & 61.41  & 50.32  & 57.63   & NA & 37.42 & \textbf{32.07}  & 34.43  \\
{\tt INT9-S4-8}  & 56.24  & 46.21  & 61.97   & NA & 38.44 & \textbf{33.05}  & 35.3   \\
{\tt INT9-S5-8}  & 56.09  & 44.01  & 59.55   & NA & 34.60 & \textbf{31.21}  & 32.75  \\
{\tt INT9-S6-8}  & 56.09  & 46.21  & 59.55   & NA & 33.69 & \textbf{30.63}  & 39.87  \\ \bottomrule
\end{tabular}
\end{adjustbox}
\end{table}

\begin{table}[htbp]
\caption{Results of all algorithms for {\tt INT10-INT11}}
\label{tb:all_results_int10_int11}
\centering
\begin{adjustbox}{width=.9\textwidth}
\begin{tabular}{lccccccc} 
\toprule
case       & FixedTime & MaxPressure & SOTL  & FRAP  & DQTSC-M & \begin{tabular}[c]{@{}l@{}}AttendLight\\  single-env\end{tabular}  & \begin{tabular}[c]{@{}l@{}}AttendLight\\ multi-env \end{tabular} \\ \midrule 
\multicolumn{8}{l}{\bf \tt INT10, 4 phases} \\ \midrule
{\tt INT10-S1-4} & 79.09  & 160.14 & 80.04  & NA & 44.30 & 41.21  & \textbf{37.46}  \\
{\tt INT10-S2-4} & 61.47  & 162.98 & 89.5   & NA & 38.49 & \textbf{34.91}  & 44.76  \\
{\tt INT10-S3-4} & 74.28  & 162.46 & 97.49  & NA & 40.64 & \textbf{35.37}  & 38.04  \\
{\tt INT10-S4-4} & 65.56  & 156.16 & 97.07  & NA & 42.81 & \textbf{37.46}  & 38.22  \\
{\tt INT10-S5-4} & 64.5   & 150.15 & 102    & NA & 36.99 & \textbf{33.44}  & 38.48  \\
{\tt INT10-S6-4} & 60.75  & 176.95 & 90.06  & NA & 37.37 & \textbf{34.03}  & 43.85  \\ \midrule 
\multicolumn{8}{l}{\bf \tt INT10, 8 phases} \\ \midrule
{\tt INT10-S1-8} & 47.54  & 158.52 & 66.52  & NA & 53.07 & 39.59    & \textbf{36.74}  \\
{\tt INT10-S2-8} & 61.47  & 161.62 & 89.5   & NA & 36.78 & \textbf{33.79}    & 35.89  \\
{\tt INT10-S3-8} & 65.56  & 162    & 97.07  & NA & 38.91 & \textbf{35.78}    & 38.49  \\
{\tt INT10-S4-8} & 74.28  & 154.2  & 97.49  & NA & 41.14 & \textbf{36.14}    & 37.1   \\
{\tt INT10-S5-8} & 64.5   & 149.01 & 102    & NA & 38.63 & \textbf{32.73}    & 37.03  \\
{\tt INT10-S6-8} & 67.43  & 176.89 & 95.14  & NA & 36.56 & \textbf{32.52}    & 35.41  \\ \midrule 
\multicolumn{8}{l}{\bf \tt INT11, 3 phases} \\ \midrule
{\tt INT11-S1-3} & 47.54  & 153.77 & 66.52  & NA & 40.18 & \textbf{37.86}    & 41.82  \\
{\tt INT11-S2-3} & 45.04  & 159.59 & 70.25  & NA & 34.98 & \textbf{32.97}    & 58.37  \\
{\tt INT11-S3-3} & 49.24  & 157.1  & 87.33  & NA & 35.51 & \textbf{33.69}    & 46.7   \\
{\tt INT11-S4-3} & 47.2   & 151.54 & 81.91  & NA & 37.67 & \textbf{36.31}    & 39.18  \\
{\tt INT11-S5-3} & 47.76  & 146.58 & 88.9   & NA & 36.01 & \textbf{34.04}    & 43.07  \\
{\tt INT11-S6-3} & 45.03  & 175.13 & 85.71  & NA & 35.48 & \textbf{34.3}     & 51.09  \\ \midrule 
\multicolumn{8}{l}{\bf \tt INT11, 5 phases} \\ \midrule
{\tt INT11-S1-5} & 56.44  & 153.77 & 63.93   & NA & 39.47 & \textbf{36.98}   & 44.02  \\
{\tt INT11-S2-5} & 49.27  & 159.59 & 66.66   & NA & 32.79 & \textbf{32.72}   & 42.14  \\
{\tt INT11-S3-5} & 51.81  & 157.1  & 67.46   & NA & 35.51 & \textbf{33.43}   & 43.7   \\
{\tt INT11-S4-5} & 50.84  & 151.54 & 69.12   & NA & 38.51 & \textbf{35.88}   & 44.25  \\
{\tt INT11-S5-5} & 54.01  & 146.58 & 72.69   & NA & 35.60 & \textbf{32.89}   & 40.29  \\
{\tt INT11-S6-5} & 48.35  & 175.13 & 78.44   & NA & 35.02 & \textbf{33.18}   & 39.24 \\ 
\bottomrule
\end{tabular}
\end{adjustbox}
\end{table}

\subsection{Mean State Results}\label{apdx:mean_state_results}

The proposed AttendLight framework uses the state-attention to produce a unified state representation from a varying number of lane-traffic information. The state attention learns the importance of each lane-traffic $g(s^t_l)$ and the weighted sum of them creates the phase-state $z_l^t$. 
Instead, one could simply use the average operator instead of the state-attention. In other words, just use equal weight $\frac{1}{|\mathcal{L}_p|}$ for all participating lane-traffic $g(s^t_l)$ in phase $p$. 

\begin{figure}
    \centering
    \includegraphics[scale=0.5]{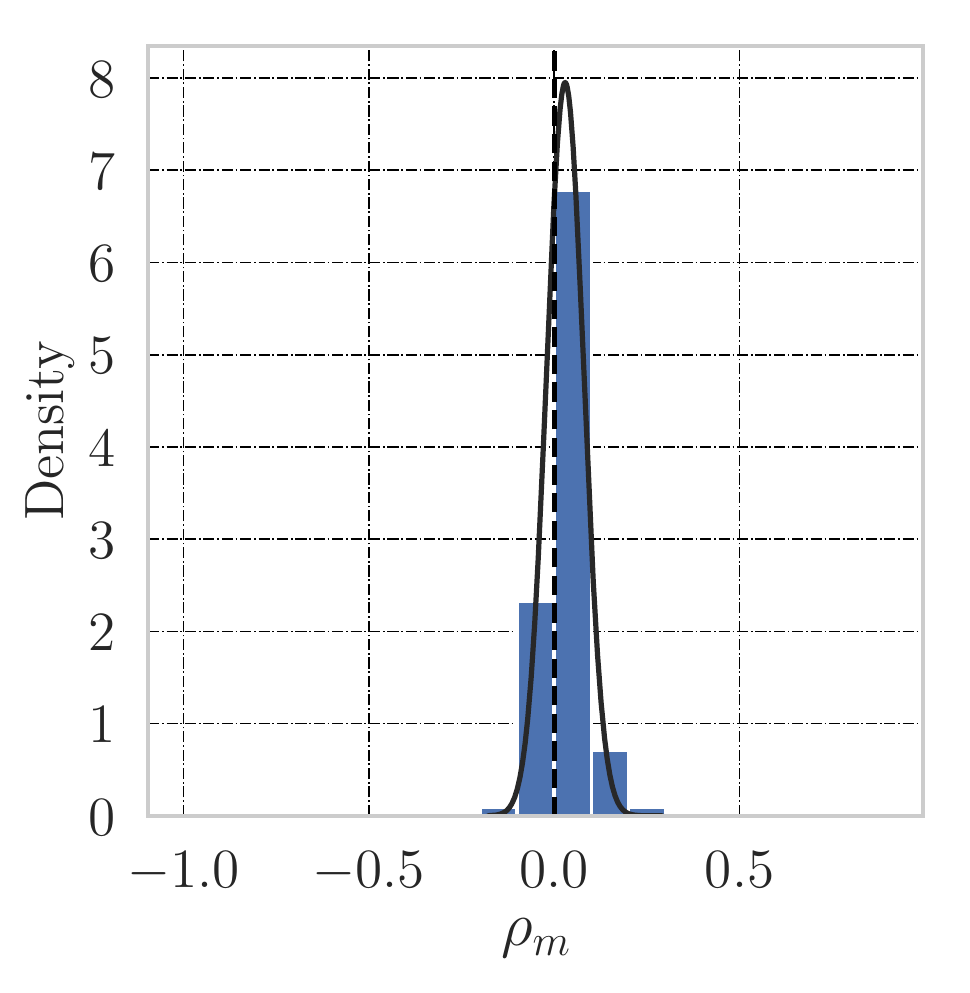}
    \caption{The comparison of state-attention with mean-state. The $\rho$ distribution shows the ATT of mean-state over state-attention.}
    \label{fig:mean-state-vs-state-attention}
\end{figure}

To analyze the performance of this approach, we modified the AttendLight framework accordingly, i.e., used the average operator instead of the state-attention and re-trained the single-env model for all 112 environment instances. Figure \ref{fig:mean-state-vs-state-attention} shows the distribution of $\rho$ for the ATT ratio of mean-state over the attention-state. We can observe that the average is on the positive side, meaning that the ATT of mean-state are bigger than those in with the state-attention. Therefore, utilizing the state-attention model helps to achieve smaller ATT. 

\subsection{Few-Shot Training for Multi-Env Regime}\label{apdx:few_shot_multi_env_regime}

In the multi-env regime, we can run a few-shot training to obtain a specialized policy for a certain intersection. To this end, we start with the trained multi-env policy and calibrate the policy for a specific intersection through a few training steps. 

\begin{figure}
    \centering
\begin{subfigure}[b]{0.32\textwidth}
    \centering
    \includegraphics[width=\textwidth]{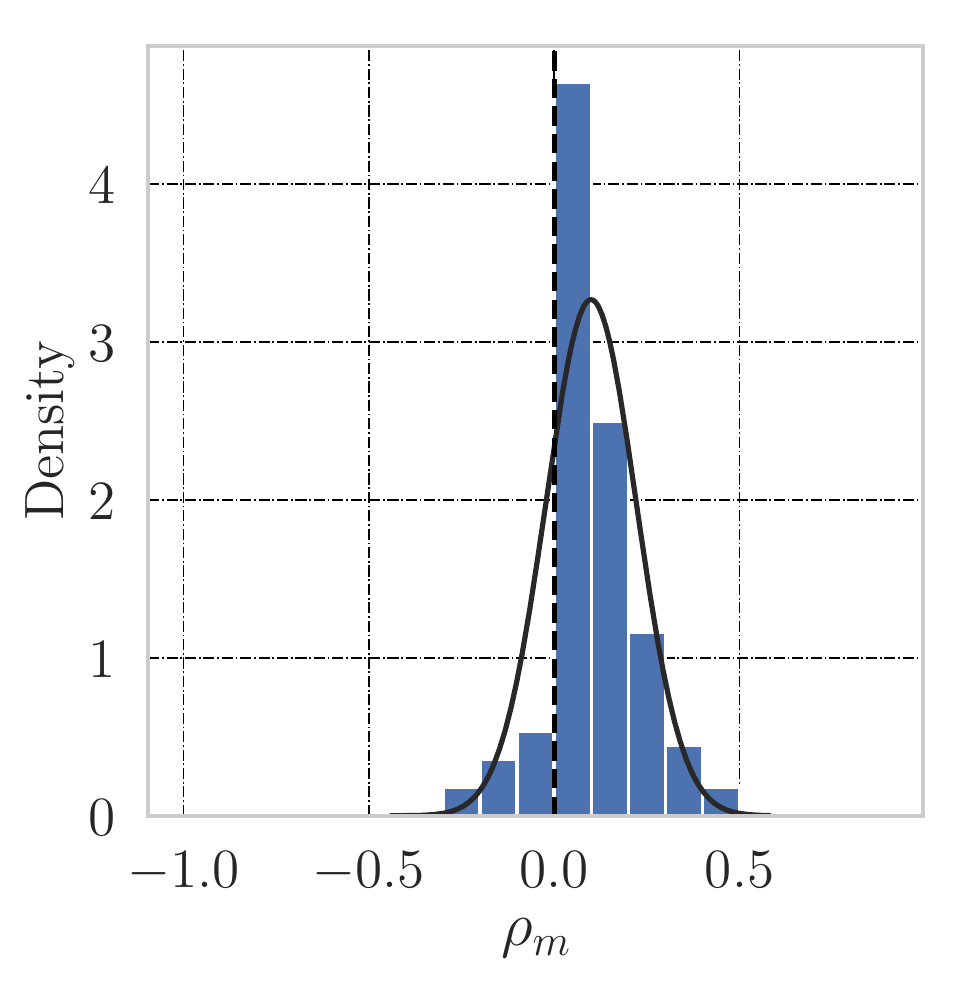}
    \caption{Before fine-tuning}
    \label{fig:result-all-env-over-single-env}
\end{subfigure} 
\begin{subfigure}[b]{0.32\textwidth}
    \centering
    \includegraphics[width=\textwidth]{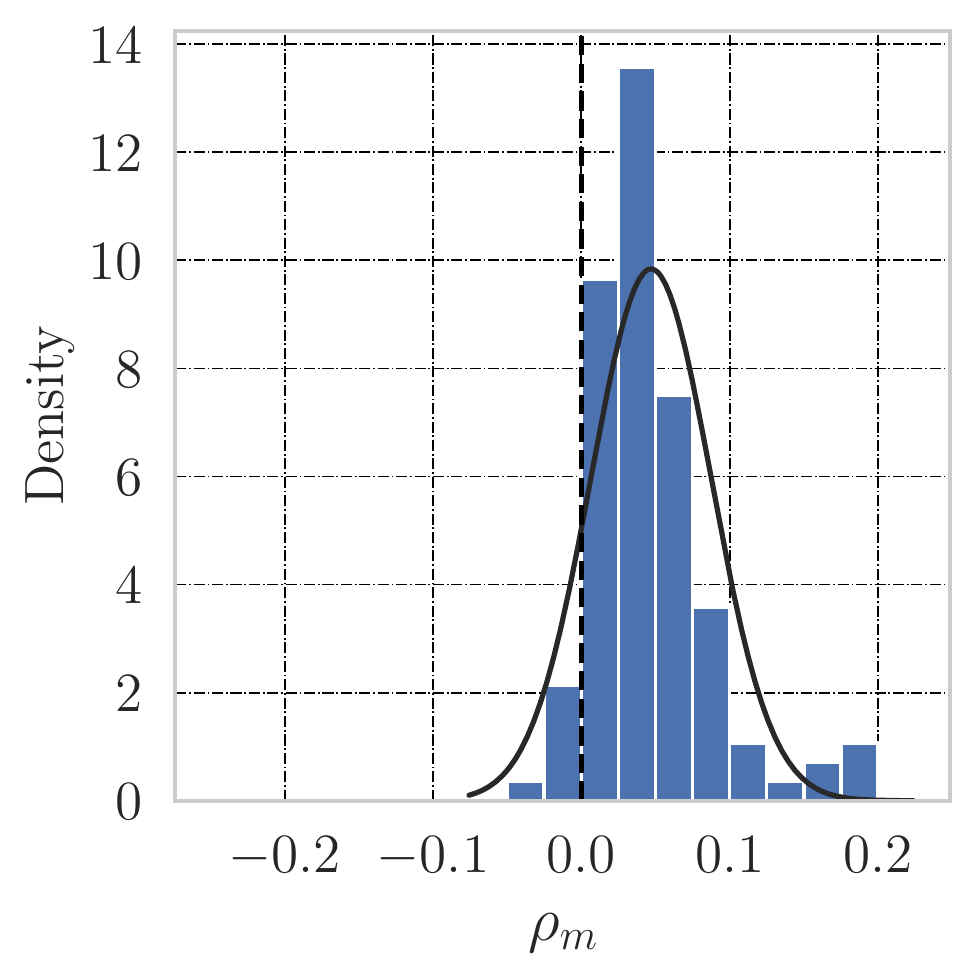}
    \caption{200-episode fine-tuning}
    \label{fig:200-fine-tuned-over-single-env}
\end{subfigure} 
\begin{subfigure}[b]{0.32\textwidth}
    \centering
    \includegraphics[width=\textwidth]{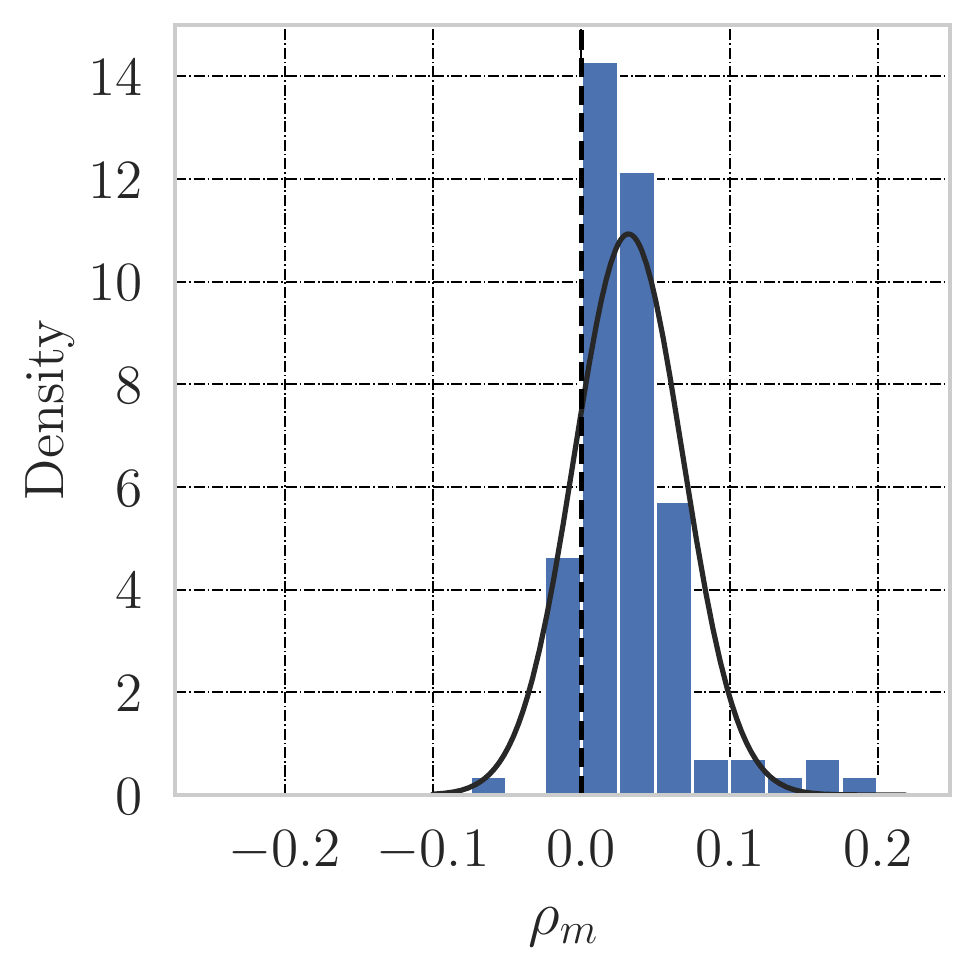}
    \caption{1000-episode fine-tuning}
    \label{fig:1000-fine-tuned-over-single-env}
\end{subfigure} 
    \caption{These plots show the effect of fine-tuning for 200 and 1000 episodes, vs the multi-env without any fine-tuning.}
    \label{fig:fine-tuning-results}
\end{figure}

We implemented this approach and fine-tuned a policy for each of the 112 environment instances. Compared to the result of the single-env regime, the maximum and the average of multi-env gap decrease significantly after 200 training episodes (instead of 100,000 episodes when trained from scratch in single-env regime) such that the ATT-gap decreased to 5\% gap in average (without fine-tuning it was 13\%). After 1000 training steps, this gap decreased to 3\%. Figure \ref{fig:fine-tuning-results} shows the distribution of $\rho_m$ which defines $\rho_m$ by the single-env after fine-tuned and the single-env before the find tuning.
As it is shown, after 200-episode of fine-tuning the distribution of $\rho$ is concentrated close to zero with a small standard deviation, leaning a bit toward the positive values. With 1000-episode, the distribution is leaned more toward zero with a smaller standard deviation. Need to mention that the fine-tuning is quite fast such that it takes 10 and 43 minutes to fine-tune the policy with 200 and 1000 episodes, respectively. 

\subsection{Separated Train/Test ATT ratio distribution}

\begin{figure}
    \centering    
    \begin{subfigure}[b]{0.19\textwidth}
    \centering
    \includegraphics[width=\textwidth]{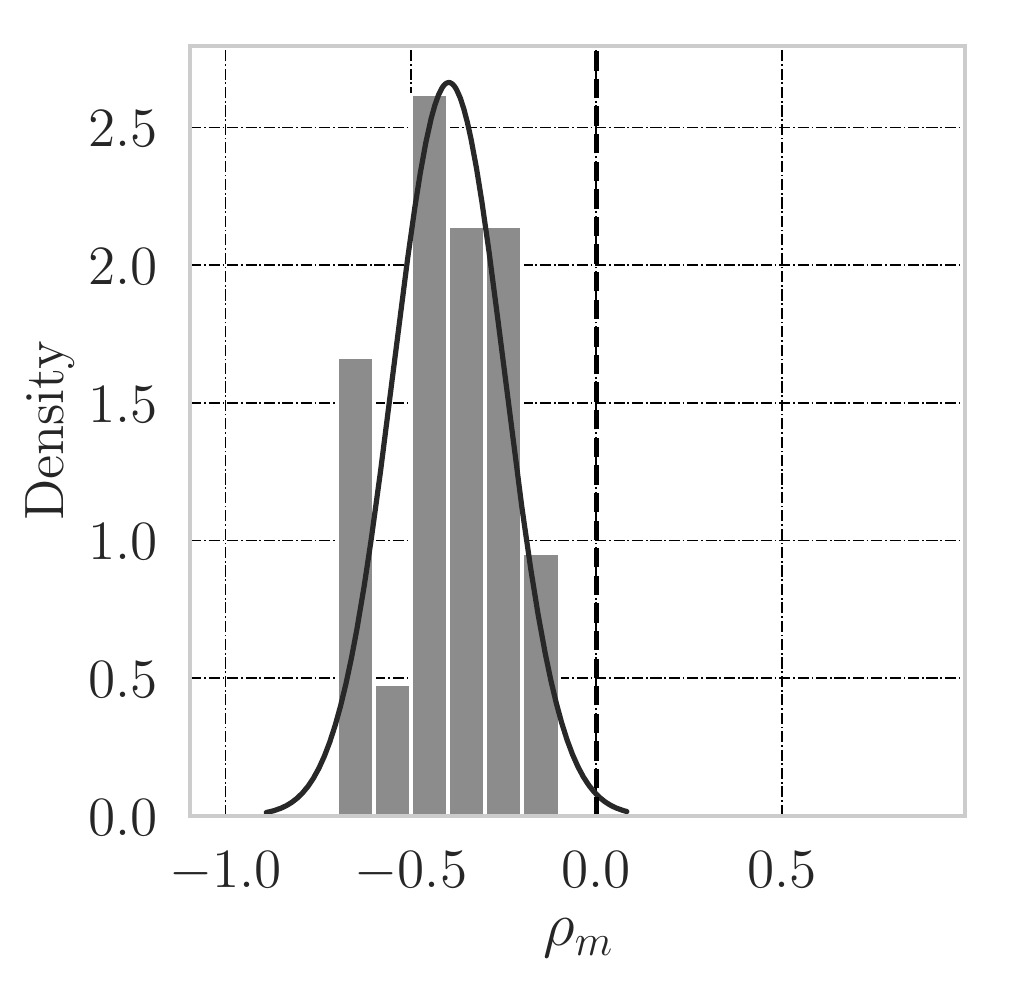}
    \caption{Fixed-time}
    \label{fig:train-env-normal-dist-of-results-fixedtime}
    \end{subfigure} 
    \begin{subfigure}[b]{0.19\textwidth}
    \centering
    \includegraphics[width=\textwidth]{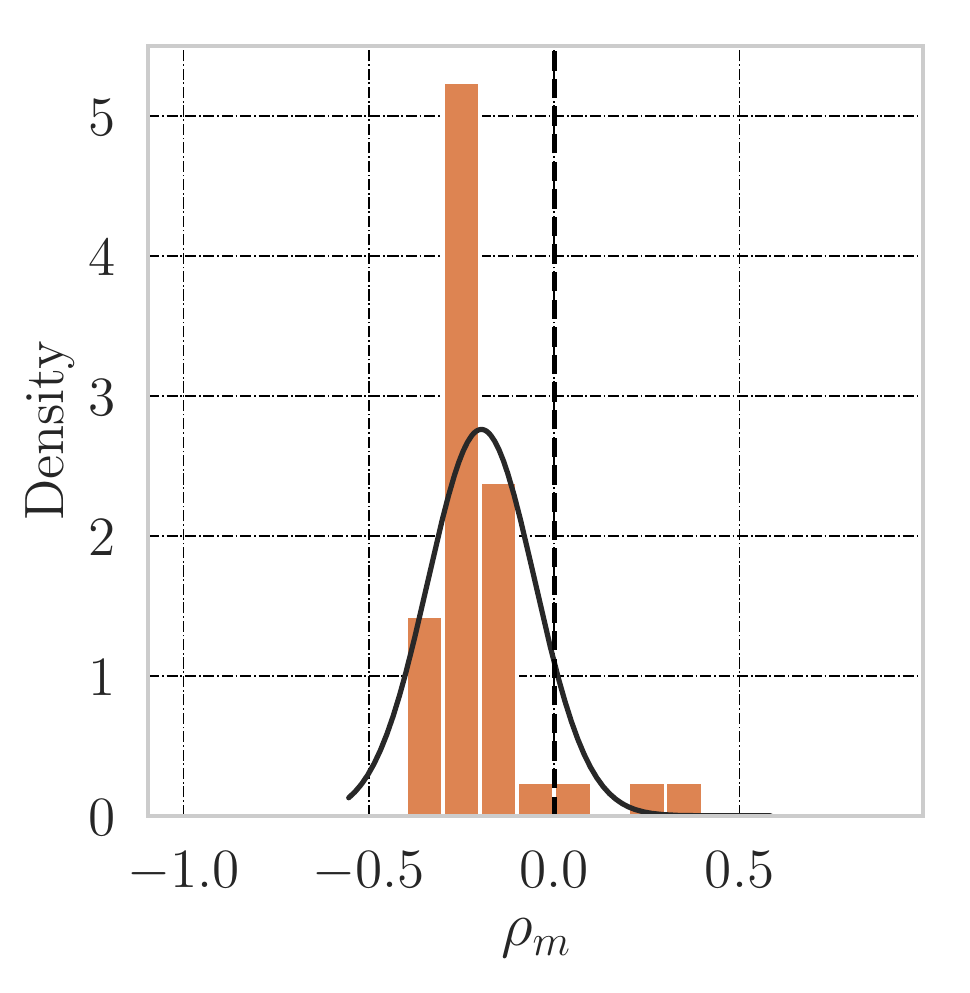}
    \caption{MaxPressure}
    \label{fig:train-env-normal-dist-of-results-maxpressure}
    \end{subfigure} 
    \begin{subfigure}[b]{0.19\textwidth}
    \centering
    \includegraphics[width=\textwidth]{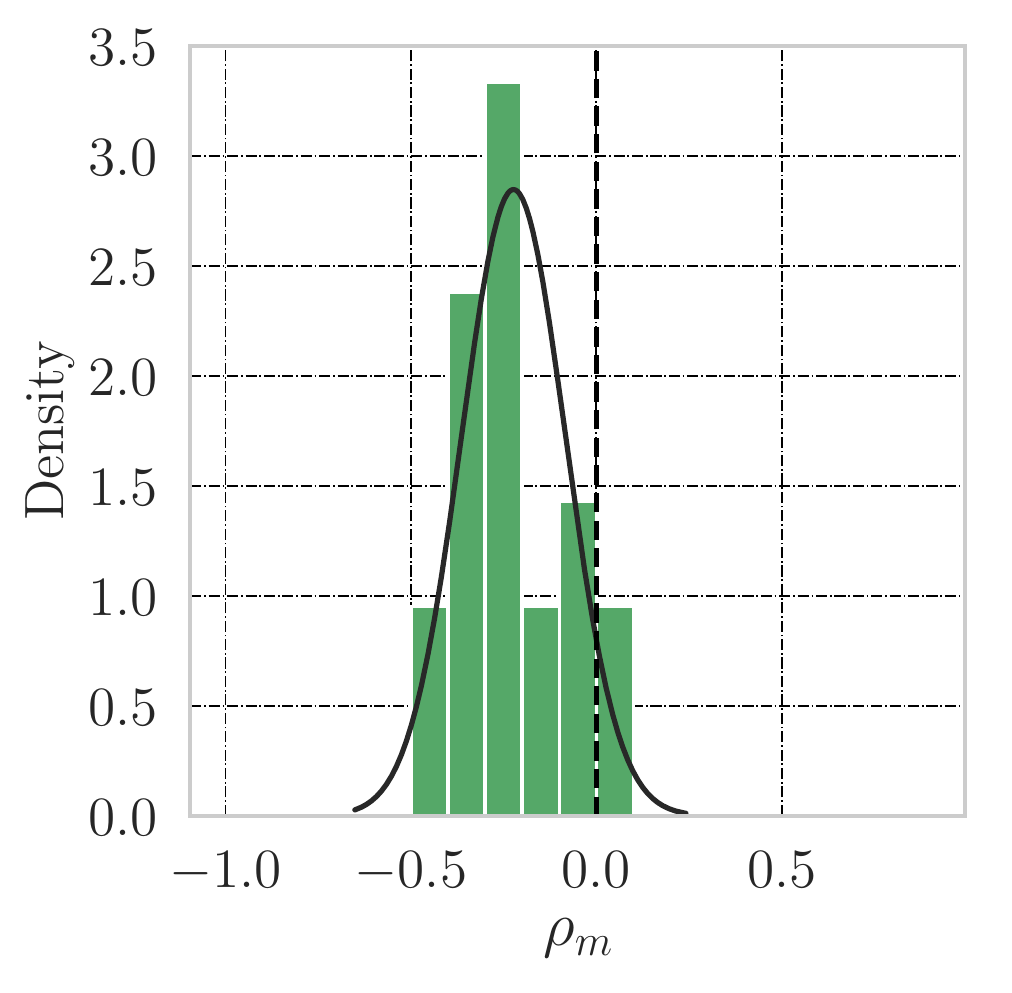}
    \caption{SOTL}
    \label{fig:train-env-normal-dist-of-results-sotl}
    \end{subfigure} 
    \begin{subfigure}[b]{0.19\textwidth}
    \centering
    \includegraphics[width=\textwidth]{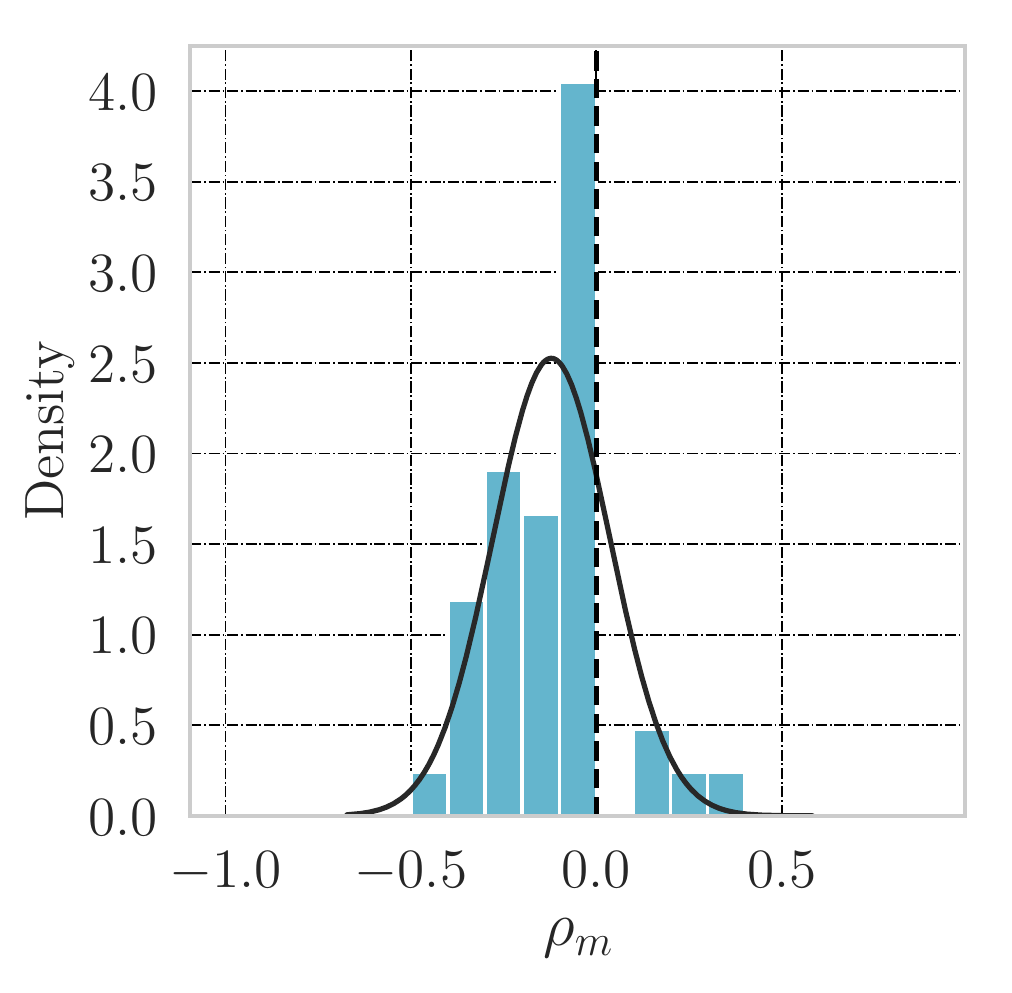}
    \caption{DQTSC-M}
    \label{fig:train-env-normal-dist-of-results-dqtscm}
    \end{subfigure} 
    \begin{subfigure}[b]{0.19\textwidth}
    \centering
    \includegraphics[width=\textwidth]{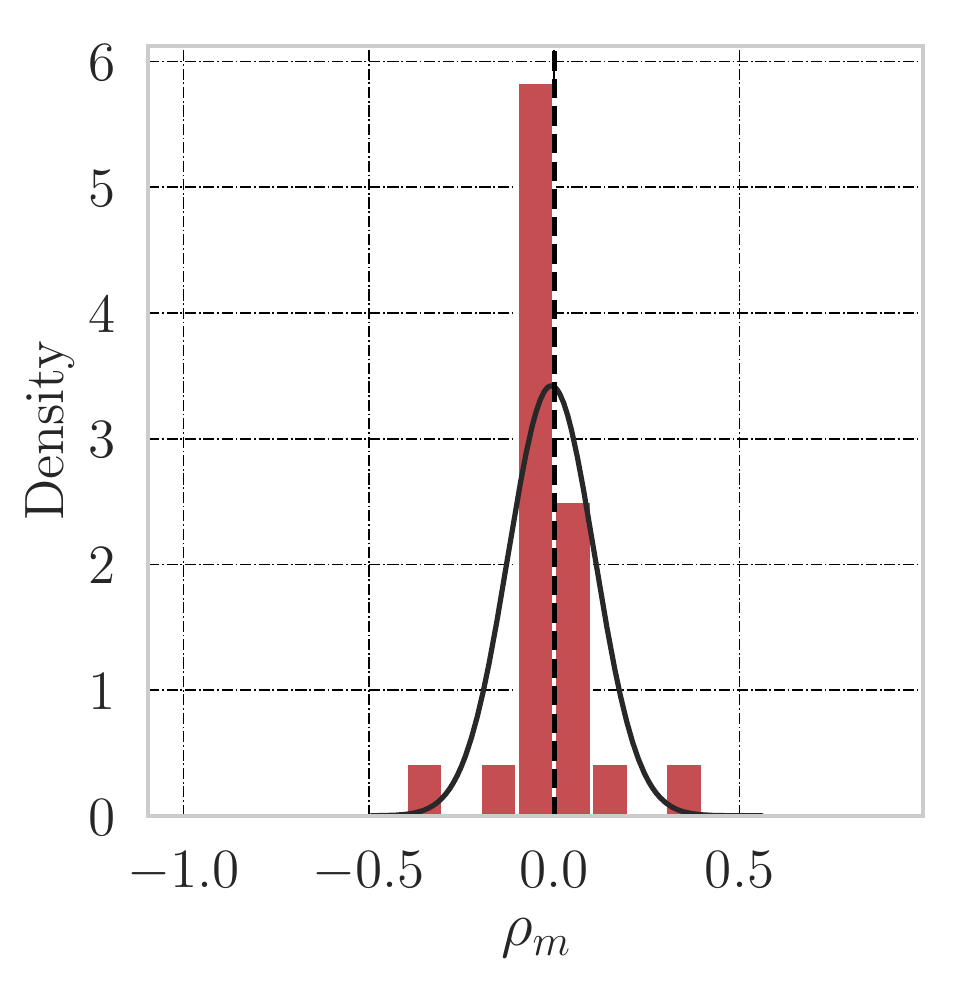}
    \caption{FRAP}
    \label{fig:train-env-normal-dist-of-results-frap}
    \end{subfigure} 

    \begin{subfigure}[b]{0.19\textwidth}
    \centering
    \includegraphics[width=\textwidth]{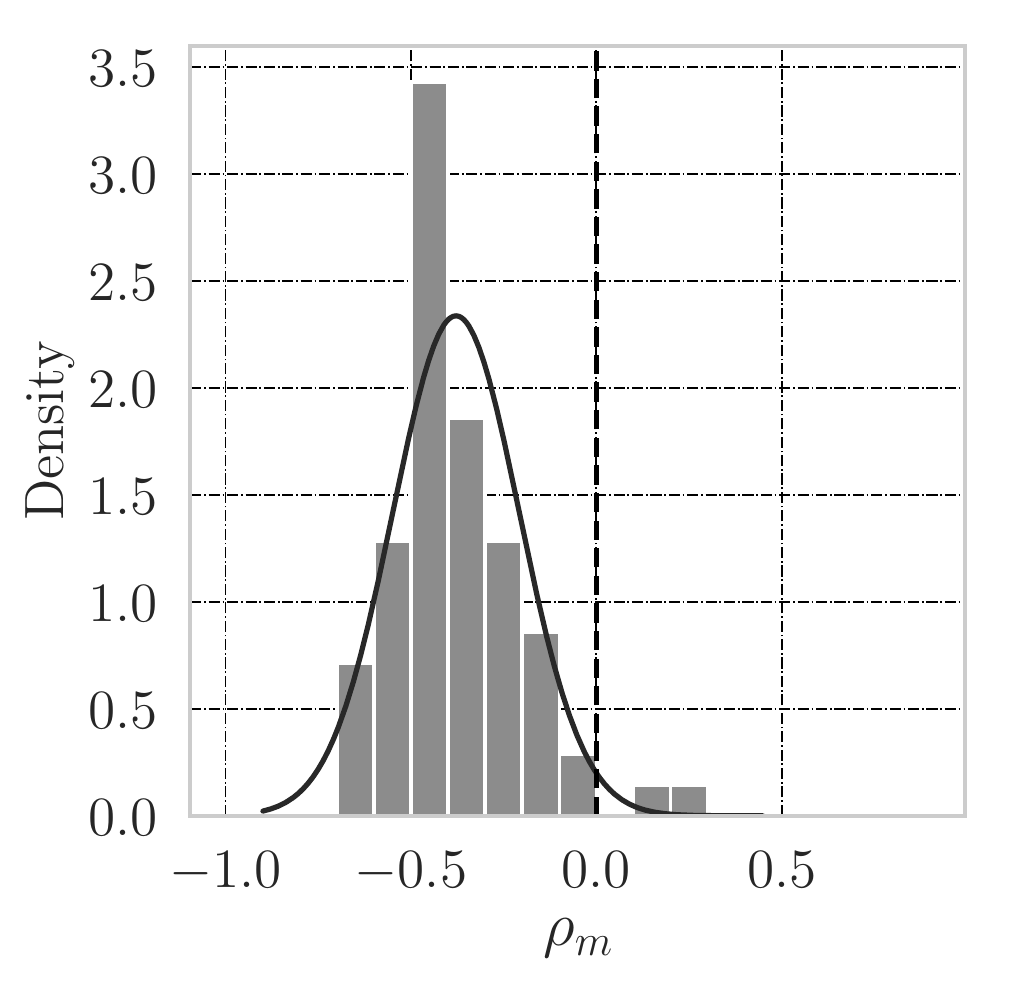}
    \caption{Fixed-time}
    \label{fig:test-env-normal-dist-of-results-fixedtime}
    \end{subfigure} 
    \begin{subfigure}[b]{0.19\textwidth}
    \centering
    \includegraphics[width=\textwidth]{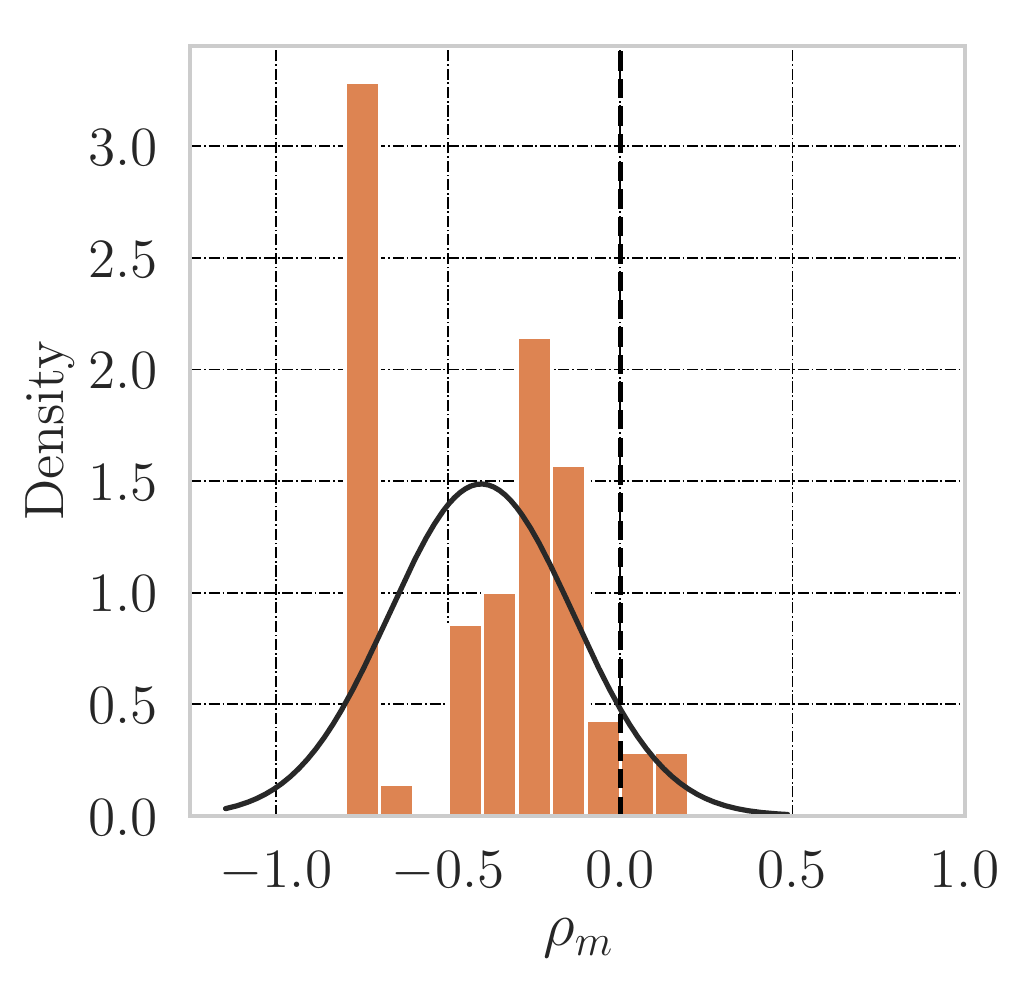}
    \caption{MaxPressure}
    \label{fig:test-env-normal-dist-of-results-maxpressure}
    \end{subfigure} 
    \begin{subfigure}[b]{0.19\textwidth}
    \centering
    \includegraphics[width=\textwidth]{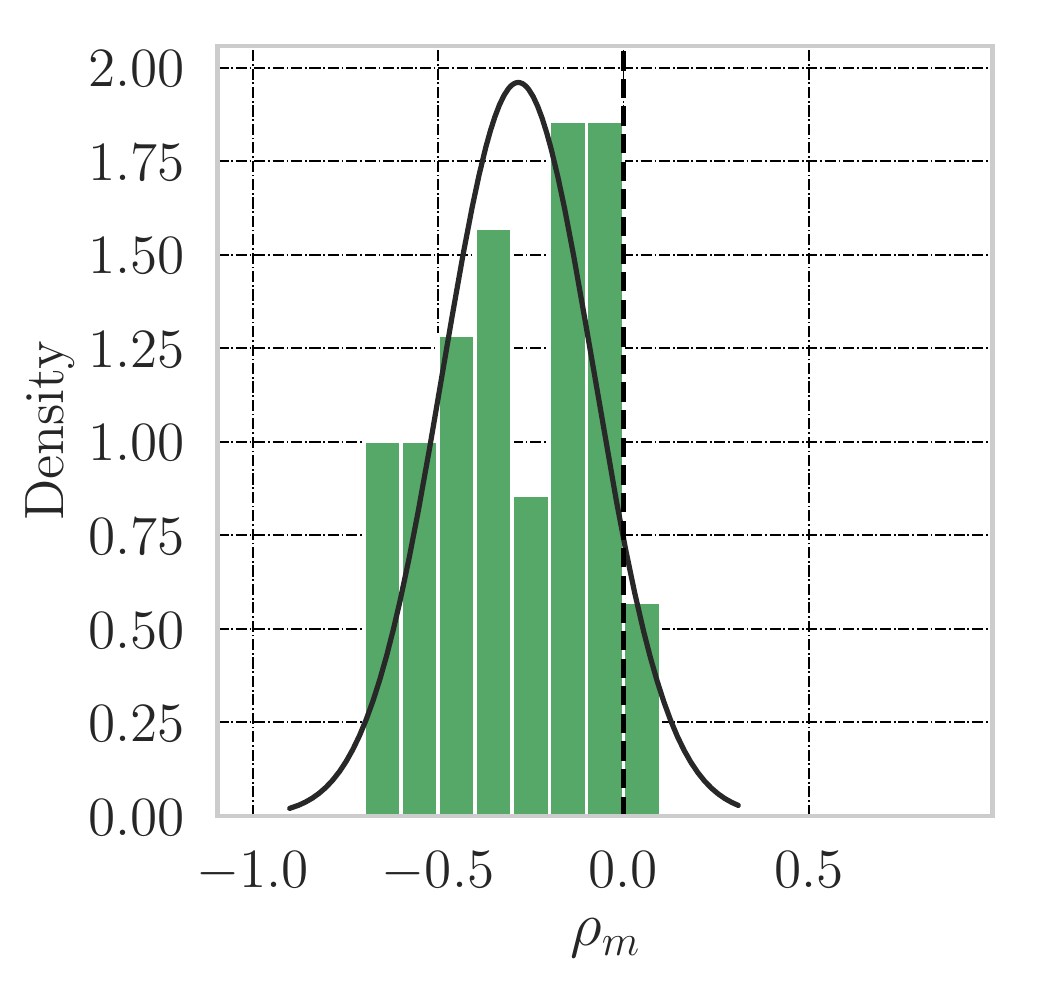}
    \caption{SOTL}
    \label{fig:test-env-normal-dist-of-results-sotl}
    \end{subfigure} 
    \begin{subfigure}[b]{0.19\textwidth}
    \centering
    \includegraphics[width=\textwidth]{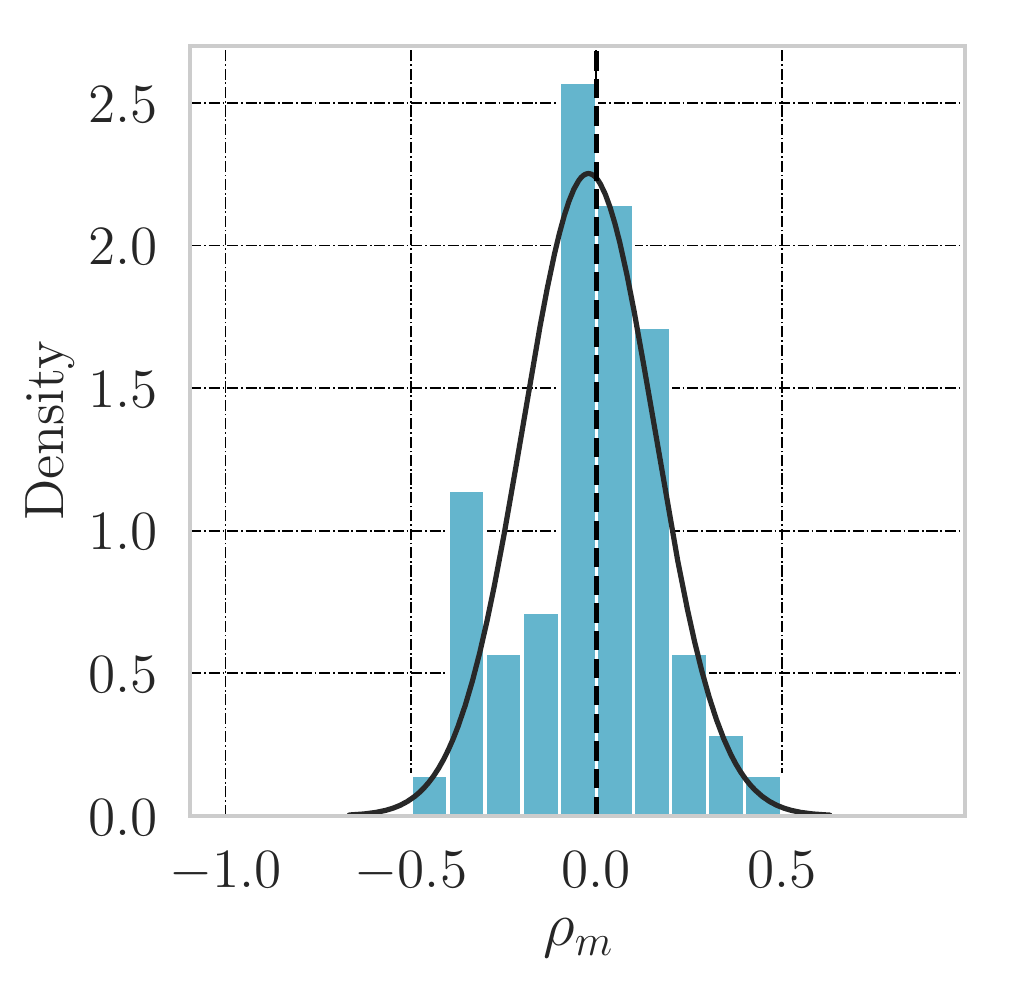}
    \caption{DQTSC-M}
    \label{fig:test-env-normal-dist-of-results-dqtscm}
    \end{subfigure} 
    \begin{subfigure}[b]{0.19\textwidth}
    \centering
    \includegraphics[width=\textwidth]{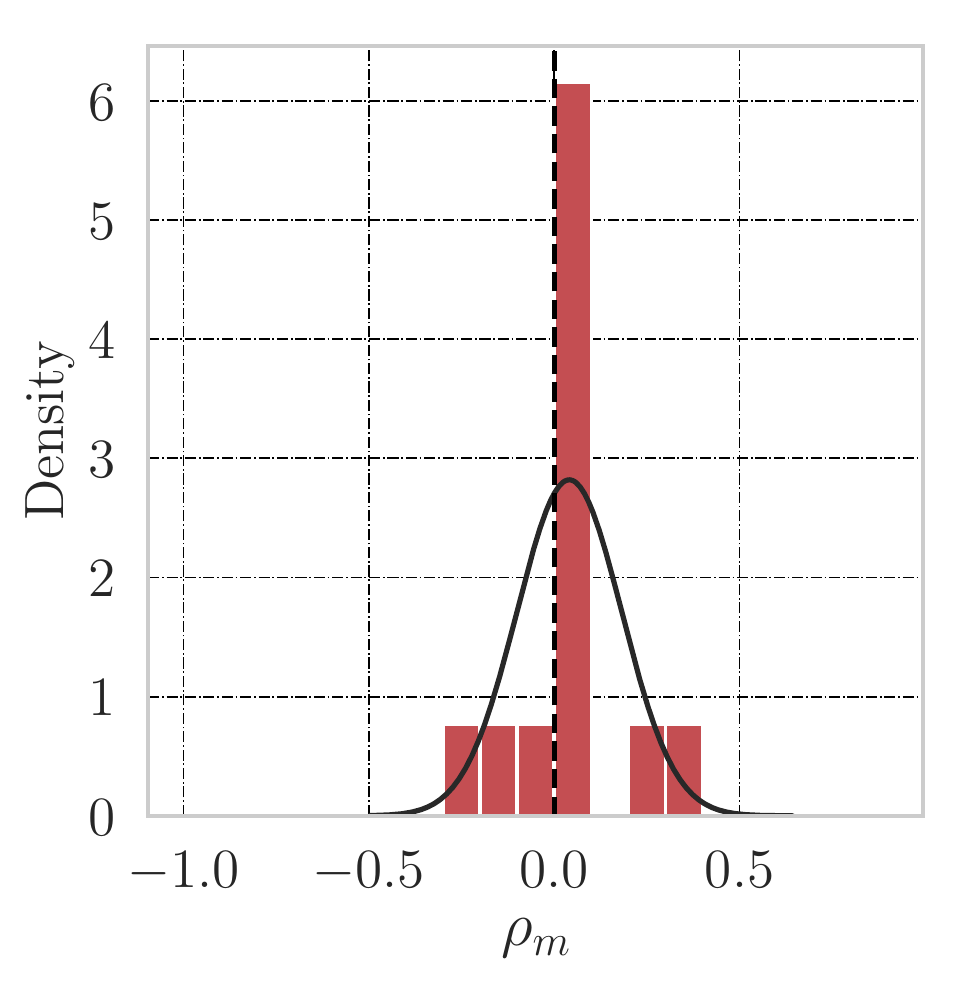}
    \caption{FRAP}
    \label{fig:test-env-normal-dist-of-results-frap}
    \end{subfigure} 
    \caption{These plots illustrate the density of $\rho_m$ over train and test instances. The first row shows the distribution of the training intersection instances and the second row shows the testing intersection instances.}
    \label{fig:train-test-env-normal-dist-of-results}
\end{figure}

\begin{figure}
    \centering    
    \begin{subfigure}[b]{0.19\textwidth}
    \centering
    \includegraphics[width=\textwidth]{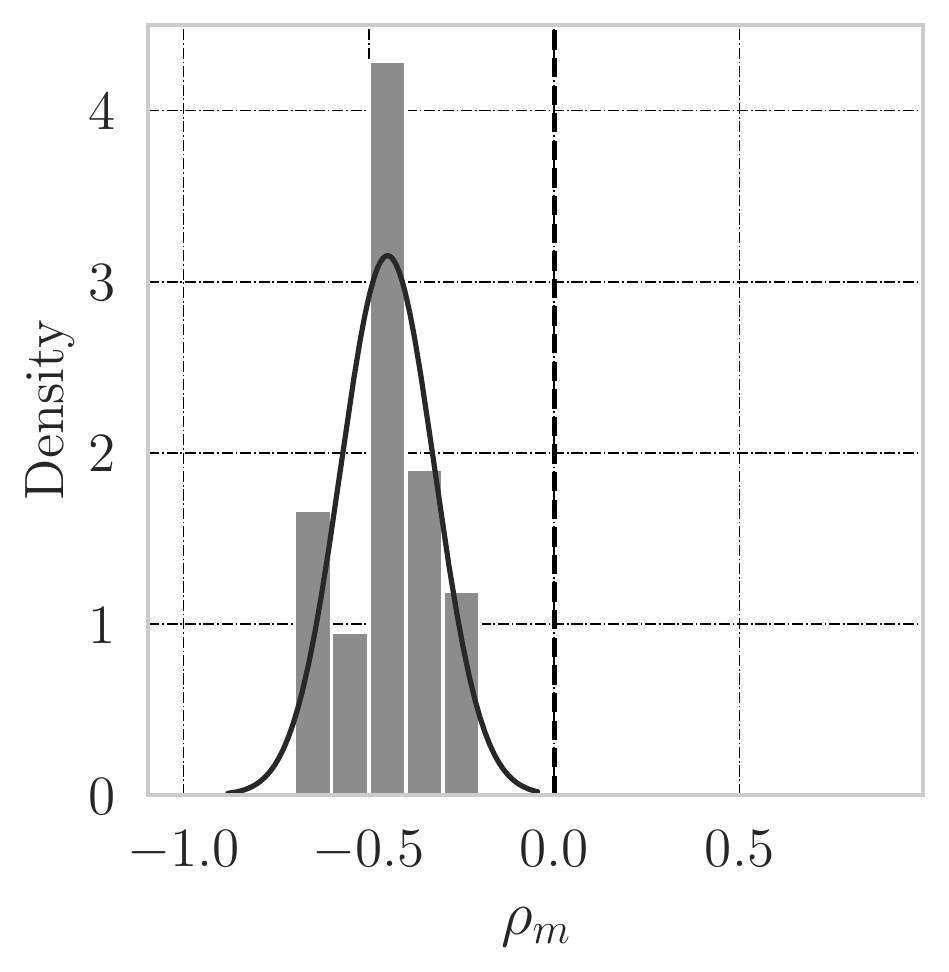}
    \caption{Fixed-time}
    \label{fig:fine-tuned-1000-train-env-normal-dist-of-results-fixedtime}
    \end{subfigure} 
    \begin{subfigure}[b]{0.19\textwidth}
    \centering
    \includegraphics[width=\textwidth]{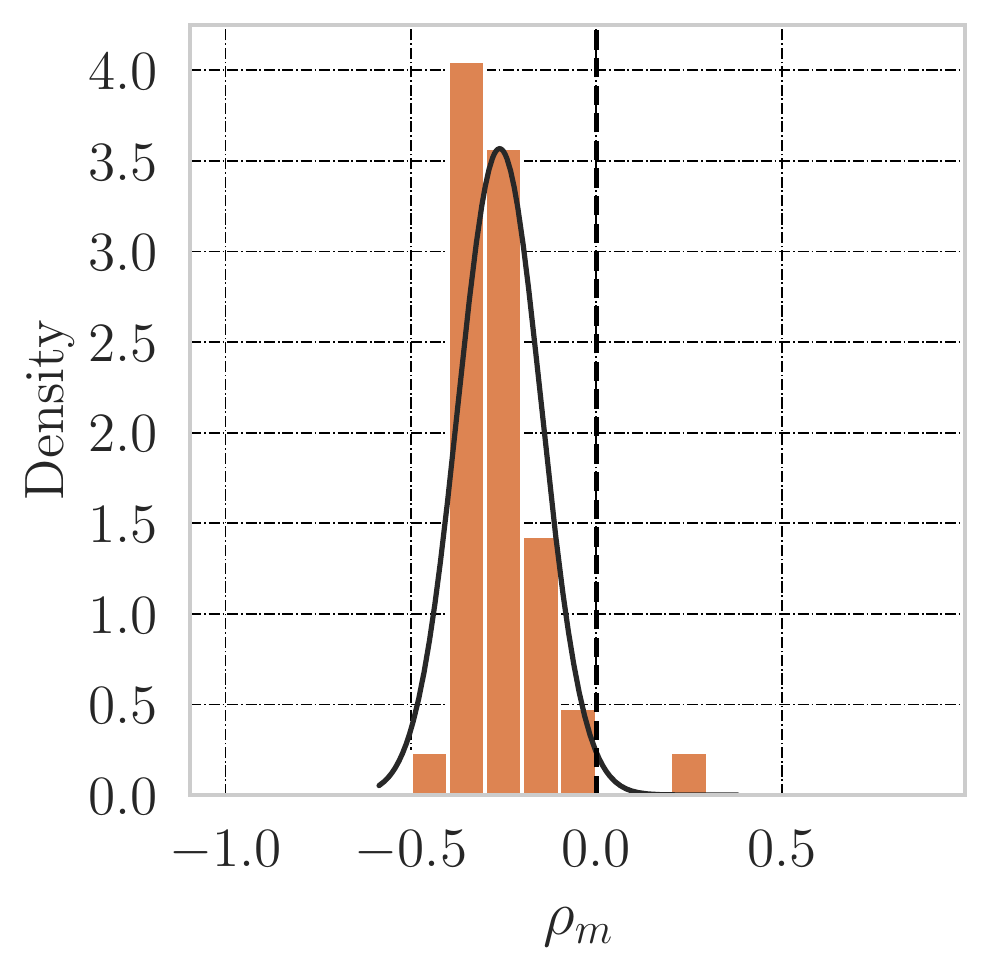}
    \caption{MaxPressure}
    \label{fig:fine-tuned-1000-train-env-normal-dist-of-results-maxpressure}
    \end{subfigure} 
    \begin{subfigure}[b]{0.19\textwidth}
    \centering
    \includegraphics[width=\textwidth]{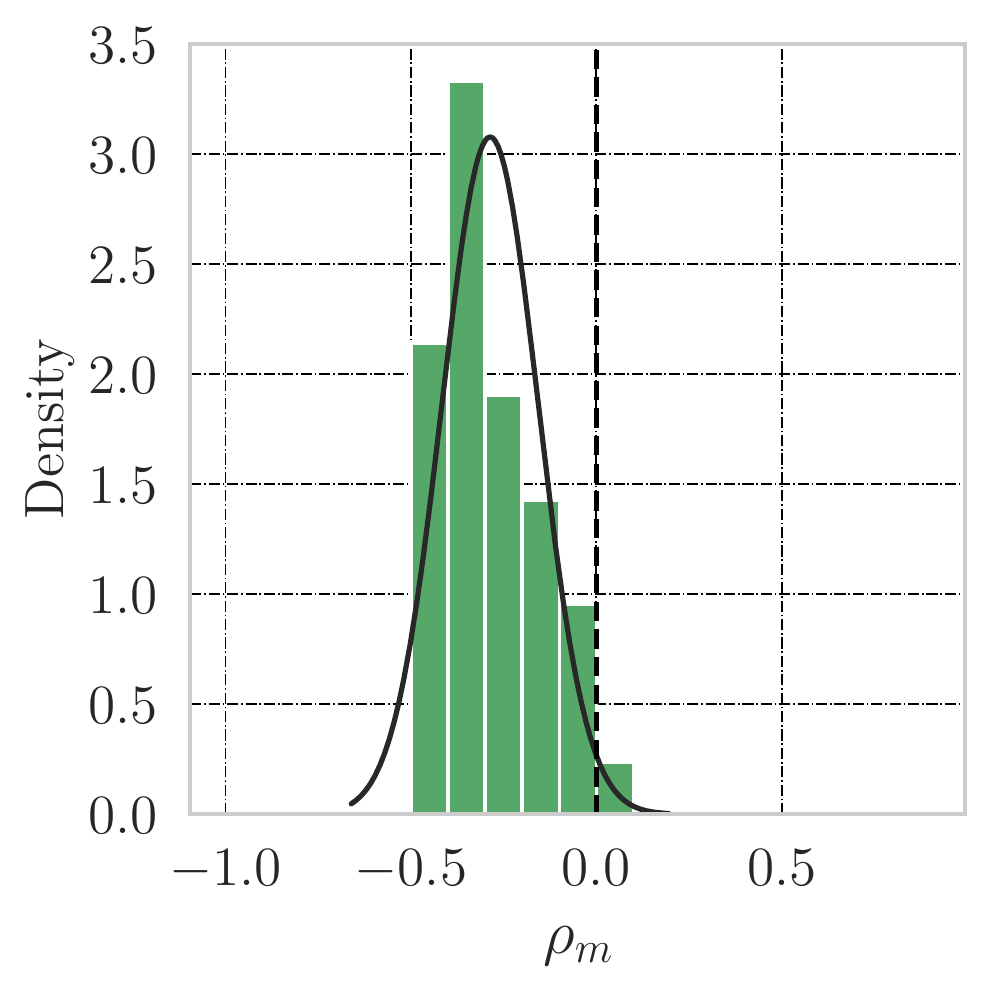}
    \caption{SOTL}
    \label{fig:fine-tuned-1000-train-env-normal-dist-of-results-sotl}
    \end{subfigure} 
    \begin{subfigure}[b]{0.19\textwidth}
    \centering
    \includegraphics[width=\textwidth]{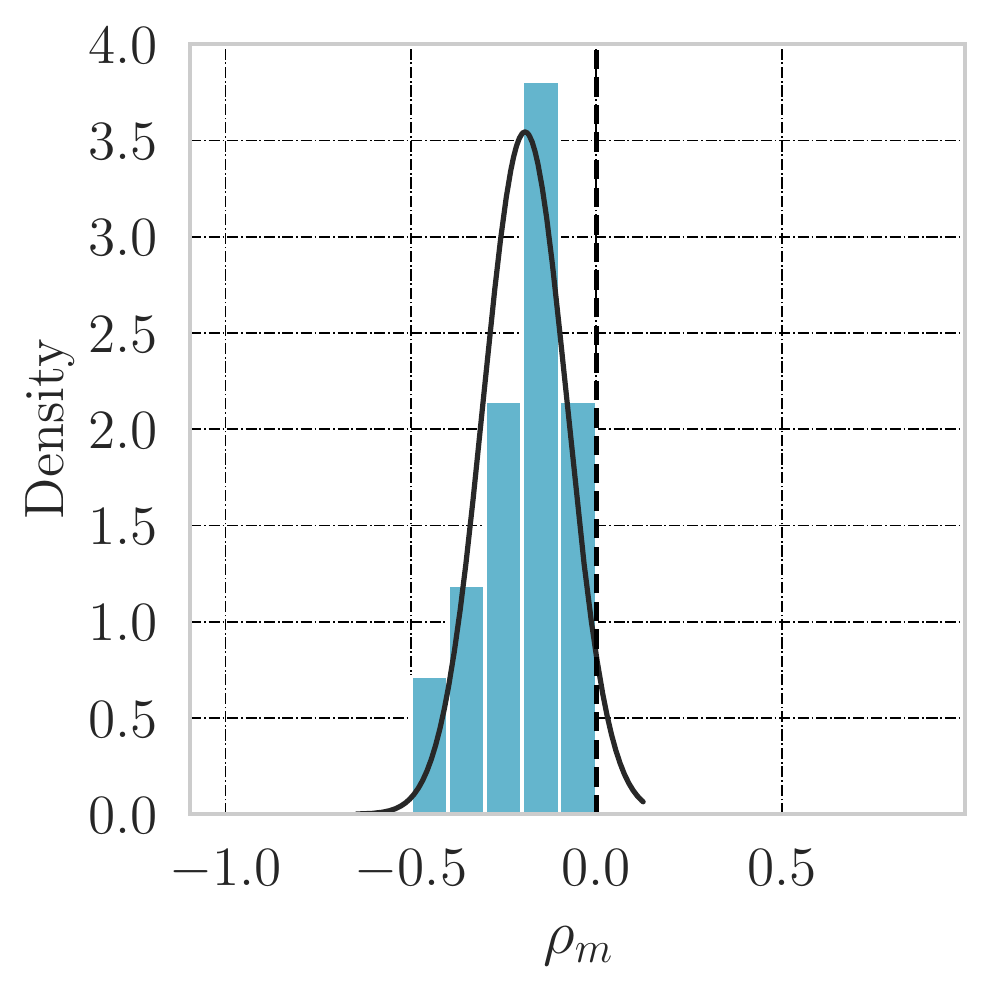}
    \caption{DQTSC-M}
    \label{fig:fine-tuned-1000-train-env-normal-dist-of-results-dqtscm}
    \end{subfigure} 
    \begin{subfigure}[b]{0.19\textwidth}
    \centering
    \includegraphics[width=\textwidth]{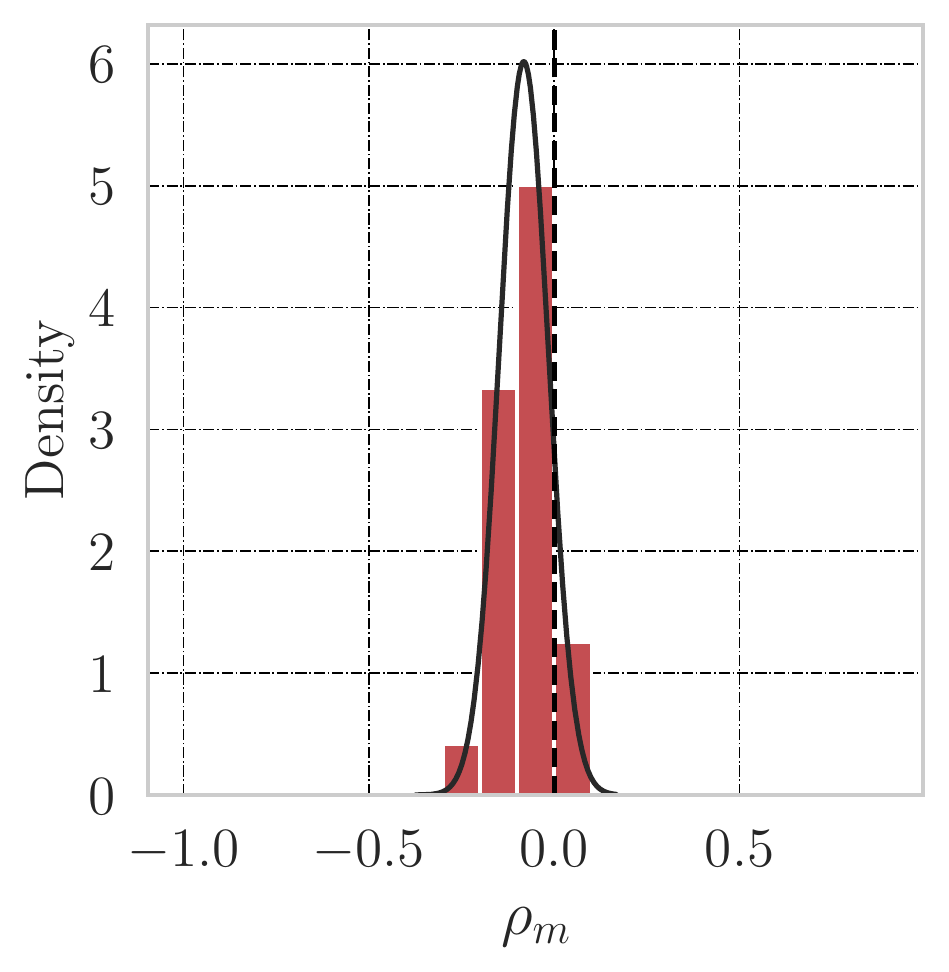}
    \caption{FRAP}
    \label{fig:fine-tuned-1000-train-env-normal-dist-of-results-frap}
    \end{subfigure} 

    \begin{subfigure}[b]{0.19\textwidth}
    \centering
    \includegraphics[width=\textwidth]{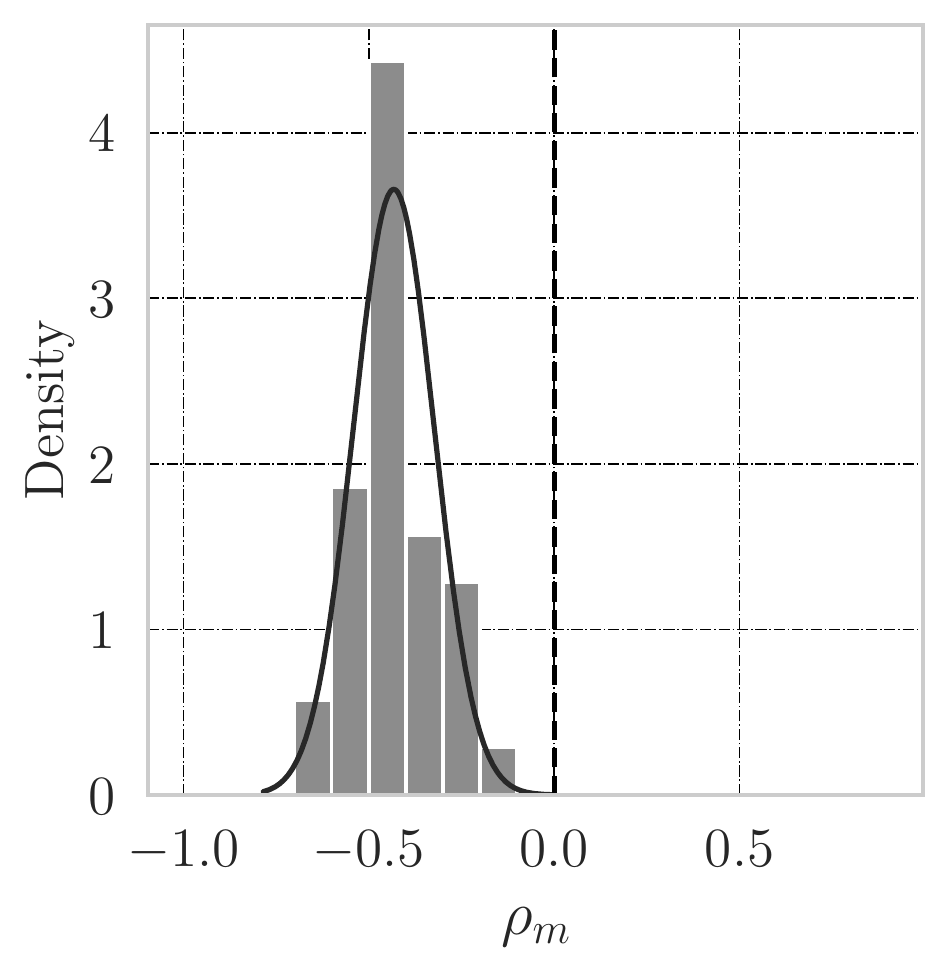}
    \caption{Fixed-time}
    \label{fig:fine-tuned-1000-test-env-normal-dist-of-results-fixedtime}
    \end{subfigure} 
    \begin{subfigure}[b]{0.19\textwidth}
    \centering
    \includegraphics[width=\textwidth]{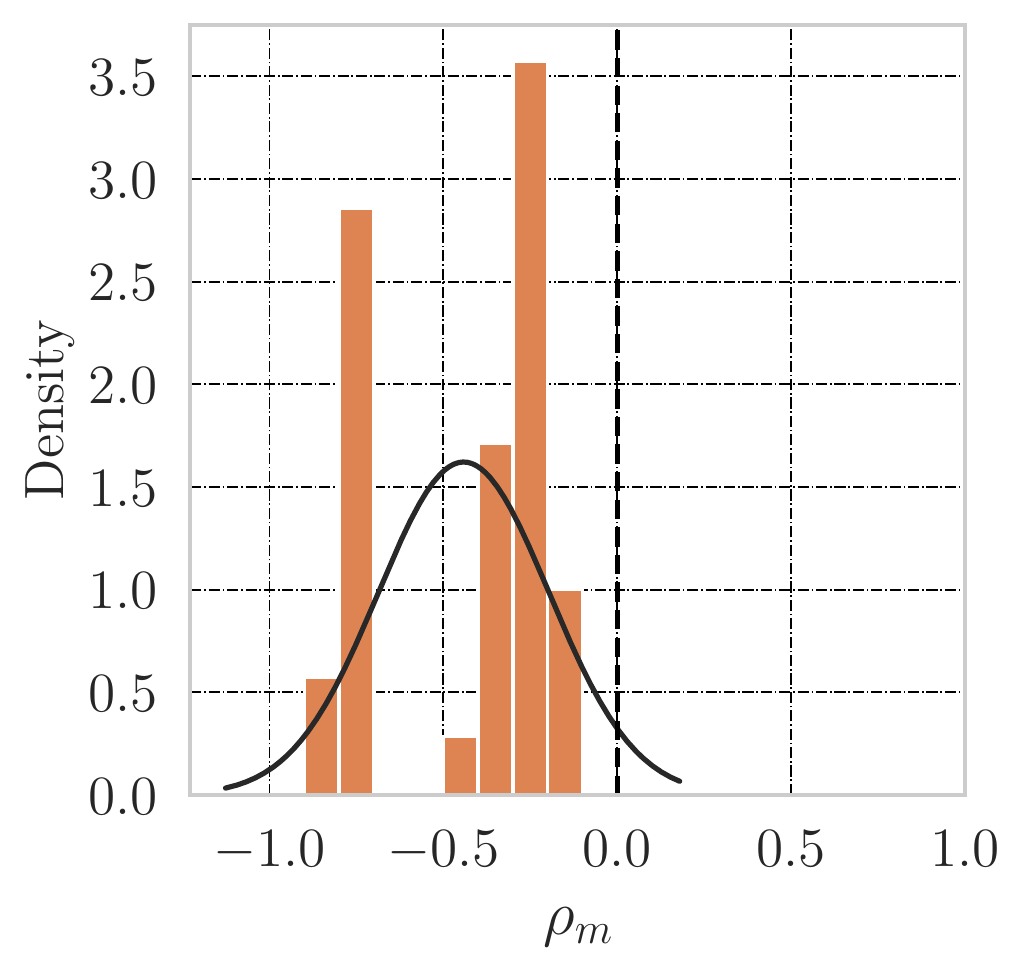}
    \caption{MaxPressure}
    \label{fig:fine-tuned-1000-test-env-normal-dist-of-results-maxpressure}
    \end{subfigure} 
    \begin{subfigure}[b]{0.19\textwidth}
    \centering
    \includegraphics[width=\textwidth]{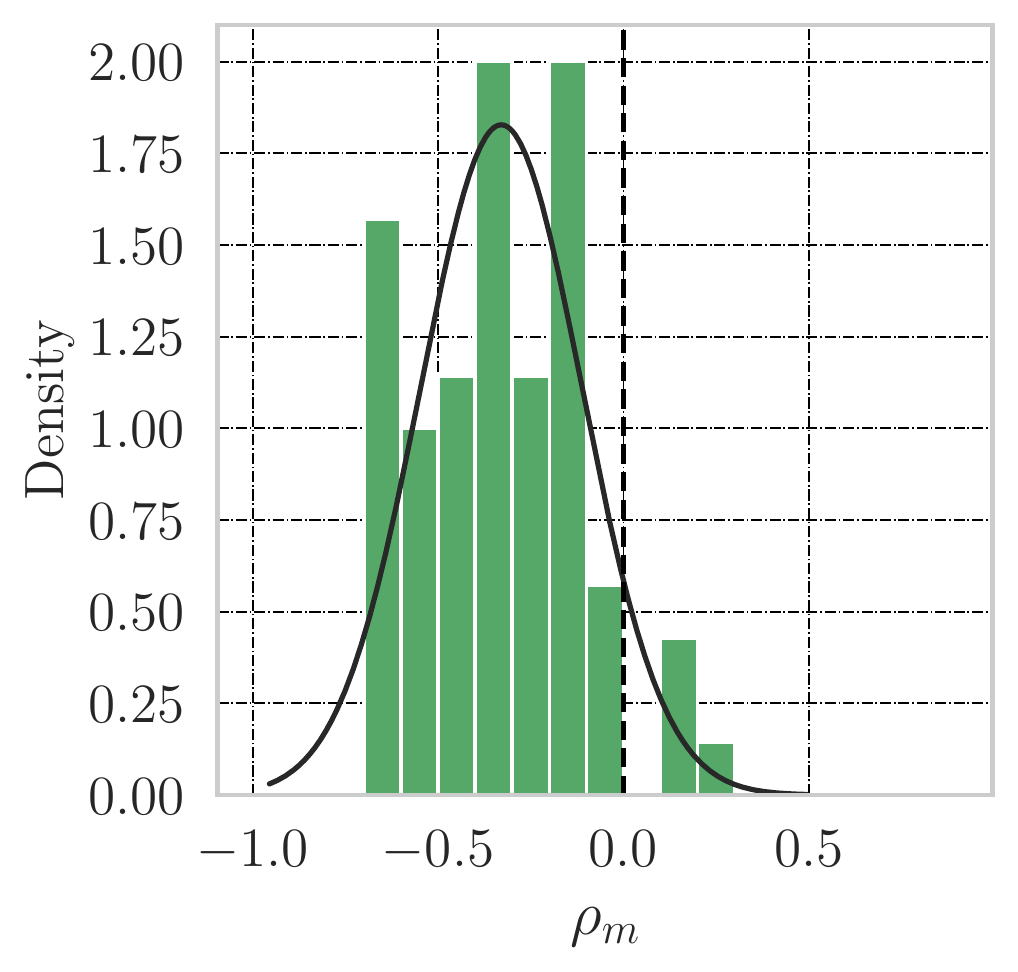}
    \caption{SOTL}
    \label{fig:fine-tuned-1000-test-env-normal-dist-of-results-sotl}
    \end{subfigure} 
    \begin{subfigure}[b]{0.19\textwidth}
    \centering
    \includegraphics[width=\textwidth]{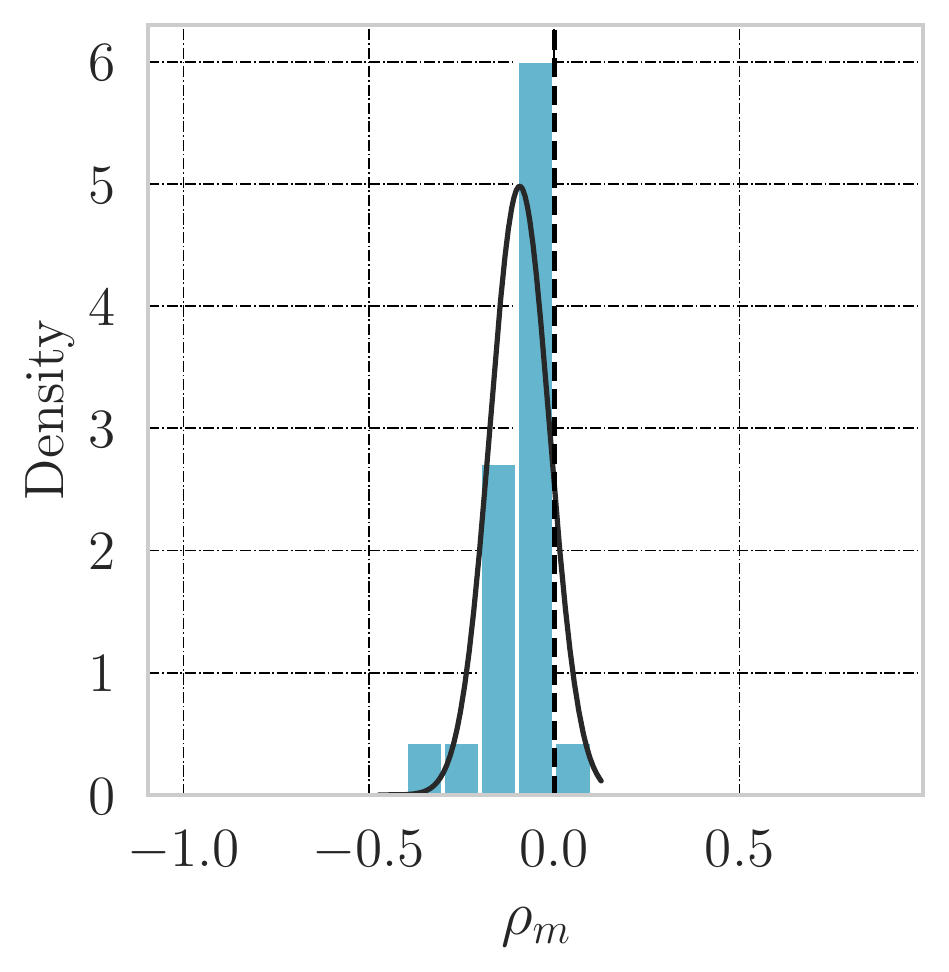}
    \caption{DQTSC-M}
    \label{fig:fine-tuned-1000-test-env-normal-dist-of-results-dqtscm}
    \end{subfigure} 
    \begin{subfigure}[b]{0.19\textwidth}
    \centering
    \includegraphics[width=\textwidth]{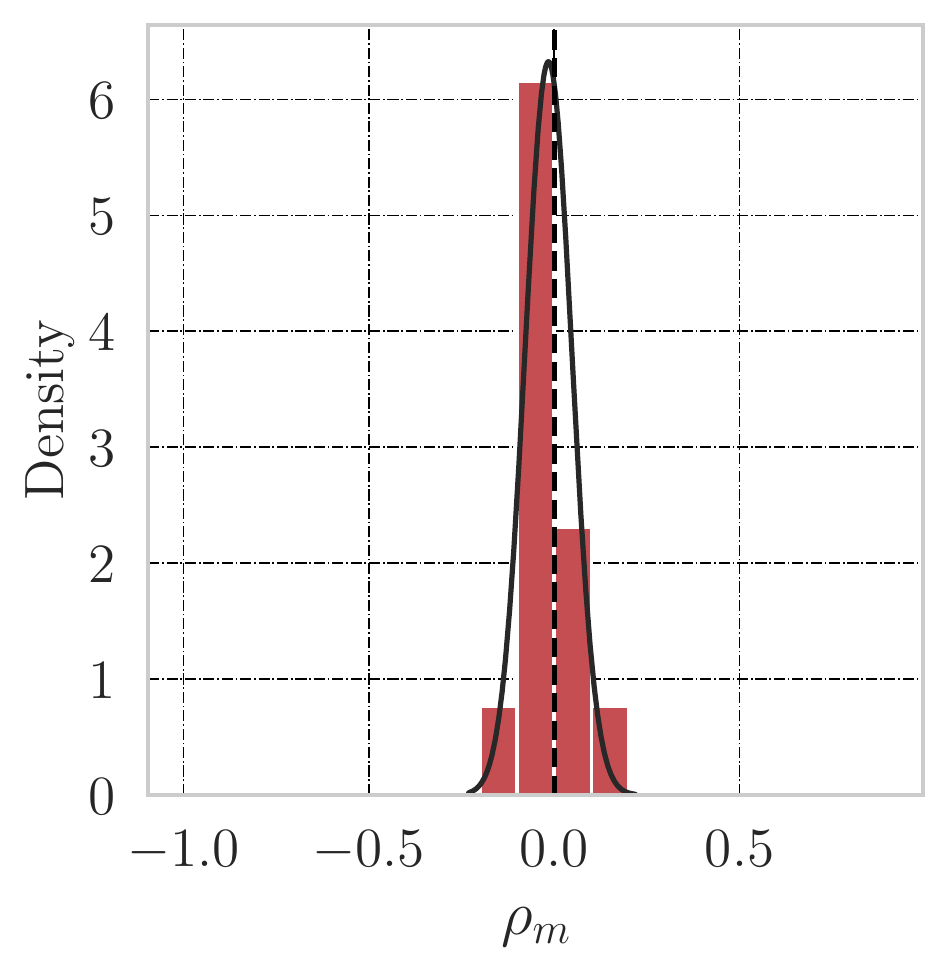}
    \caption{FRAP}
    \label{fig:fine-tuned-1000-test-env-normal-dist-of-results-frap}
    \end{subfigure} 
    \caption{These plots illustrate the density of $\rho_m$ over train and test instances. The first row shows the distribution of the train-env instances and the second row shows the test-env instances.}
    \label{fig:fine-tuned-1000-train-test-env-normal-dist-of-results}
\end{figure}

In page~\pageref{fig:all-env-normal-dist-of-results}, Figure \ref{fig:all-env-normal-dist-of-results} shows the distribution of $\rho_m$ for multi-env regime and all the baselines, for all 112 cases. To analyze the performance on test instances, we also depicted the plots for separated test and train environment instances. As it is shown in Figure~\ref{fig:train-test-env-normal-dist-of-results}, compared to the distribution of the train-instances, the distribution of the test-instances is leaned a bit toward the right side, which complies the same observation in Figure~\ref{fig:result-all-env-over-best-attendlight}. Also, Figure~\ref{fig:fine-tuned-1000-train-test-env-normal-dist-of-results} shows the distribution of $\rho_m$ when all baselines are compared with the multi-env policy which is improved by the few-shot training. As it is shown, in all the baselines the results improved such that the distribution is moved to the left and the variance of the distributions are smaller that those in Figure~\ref{fig:train-test-env-normal-dist-of-results}. 

\section{Stochastic Training Regime}\label{appsec:stoch-regime}

In this section, we propose an alternative multi-env regime, which we refer to it as \textit{stochastic multi-env} regime, that accelerates the training when we have a large number of intersections. Since running Monte-Carlo simulation for the TSCP is quite costly, it might not feasible to incorporate all training intersection instances at each training step.  Unlike the multi-env regime that we considered in this paper, the stochastic multi-env regime chooses a small mini-batch of the environment instances $n\ll|\mathcal{M}|$ to reduce the per iteration cost. The rest of the AttendLight algorithm is the same as the multi-env regime.

To test this approach, we randomly choose $n=5$ environment instances from Table~\ref{tb:details_of_training_cases}, and trained the AttendLight with a single layer LSTM of hidden dimension 256, using the learning rate 0.005 in Adam optimizer. Not surprisingly, this approach runs faster than multi-env regime such that it runs 385 episodes at each hour, 4.5 times faster than multi-env regime. 

\begin{figure}
    \centering

\begin{subfigure}[b]{0.37\textwidth}
    \centering
    \includegraphics[width=\textwidth]{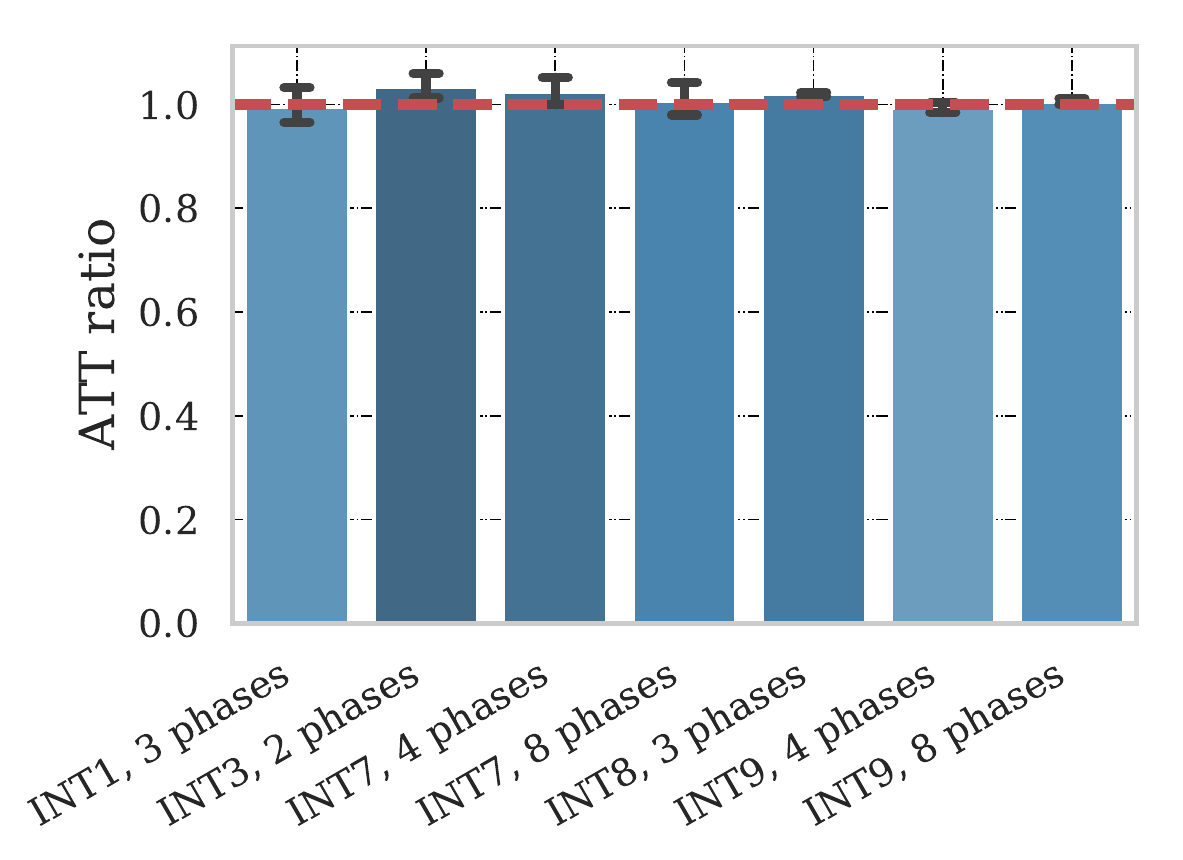}
    \caption{Stochastic multi-env vs multi-env regime on training intersections instances}
    \label{fig:result-5rand-env-over-all-env-training}
\end{subfigure} 
\hspace{5pt}
\begin{subfigure}[b]{0.485\textwidth}
    \centering
    \includegraphics[width=\textwidth]{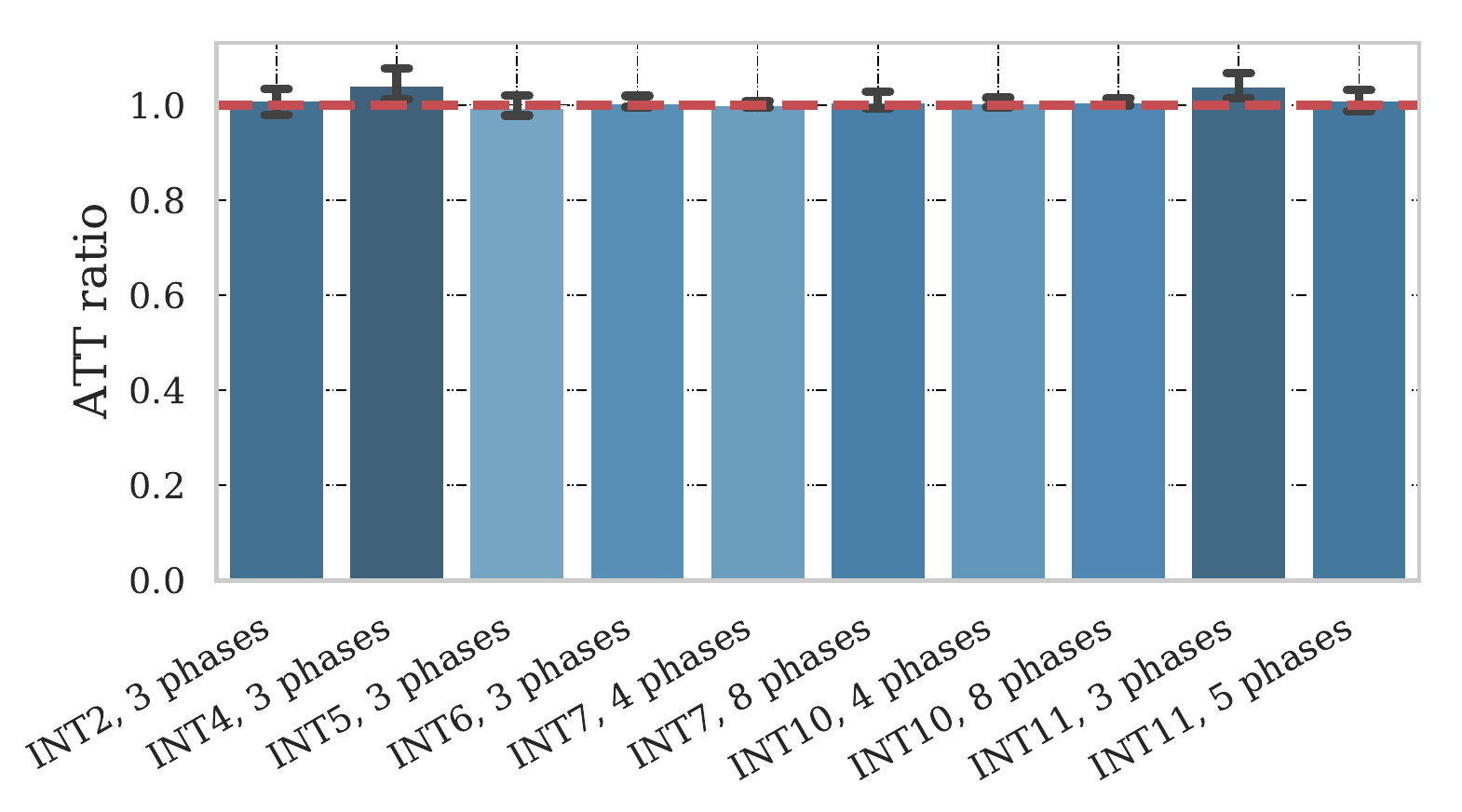}
    \caption{Stochastic multi-env vs multi-env regime on testing intersections instances}
    \label{fig:result-5rand-env-over-all-env-testing}
\end{subfigure}    

\begin{subfigure}[b]{0.37\textwidth}
    \centering
    \includegraphics[width=\textwidth]{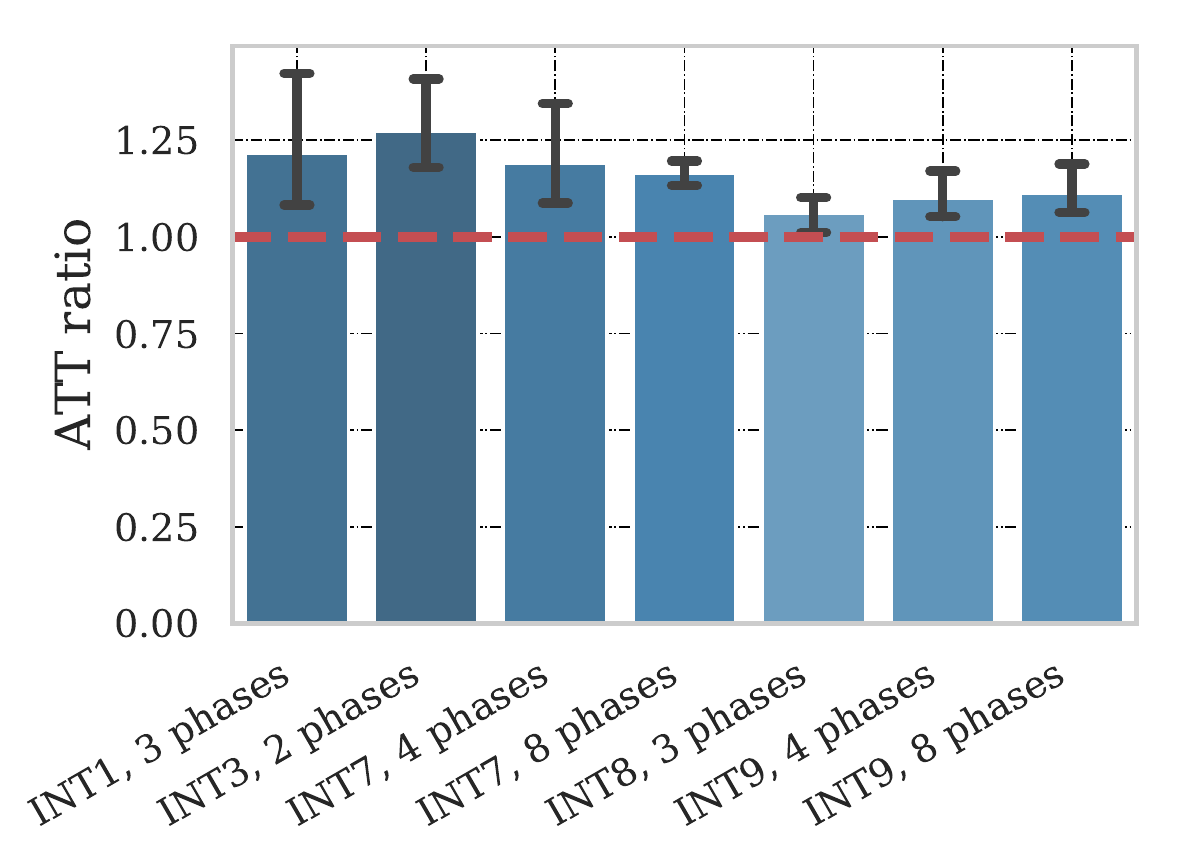}
    \caption{Stochastic multi-env vs single-env regime on training intersections instances}
    \label{fig:result-5rand-env-over-best-attendlight-training}
\end{subfigure}
\hspace{5pt}
\begin{subfigure}[b]{0.485\textwidth}
    \centering
    \includegraphics[width=\textwidth]{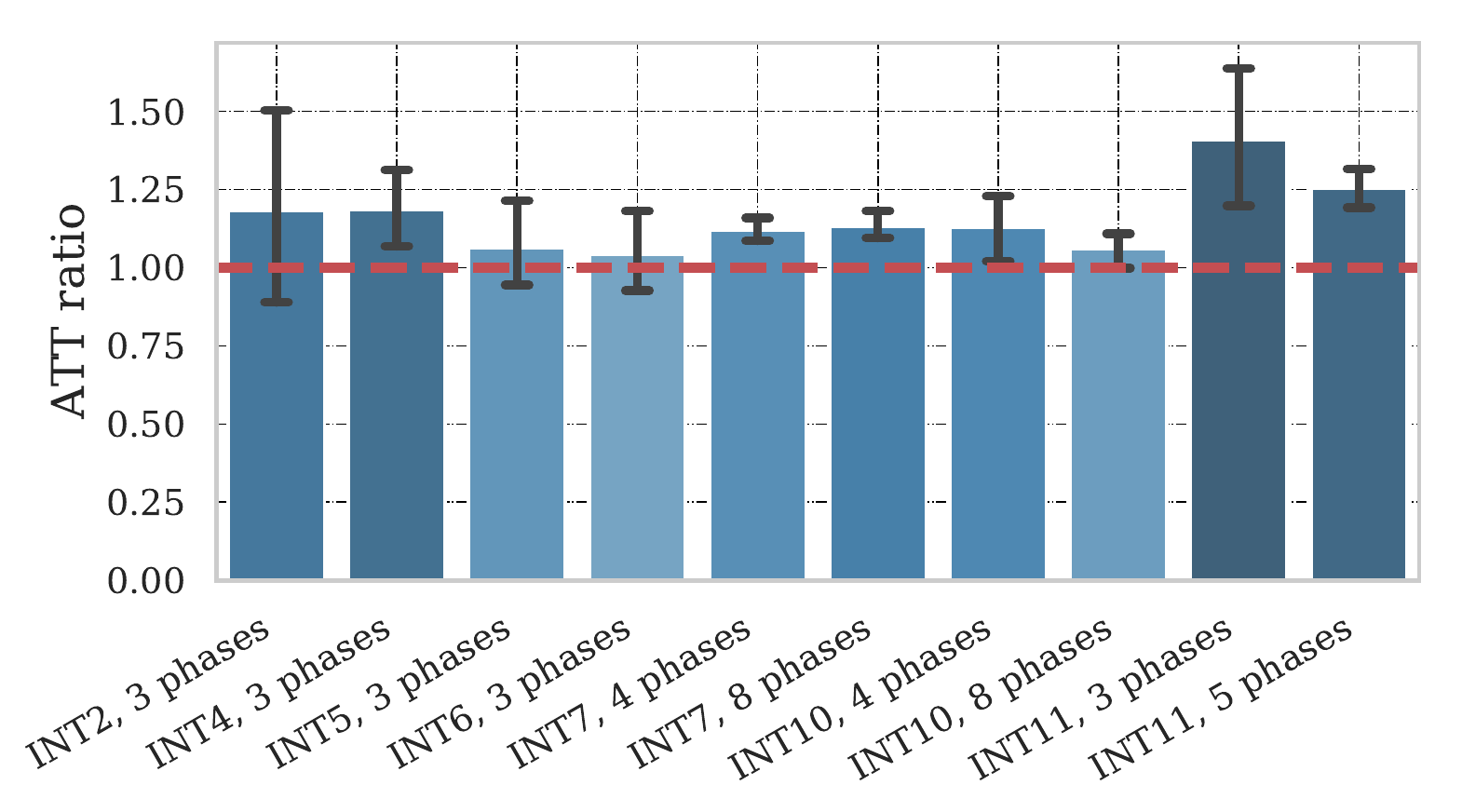}
    \caption{Stochastic multi-env vs single-env regime on testing intersections instances}
    \label{fig:result-5rand-env-over-best-attendlight-testing}
\end{subfigure}
    \centering
    \caption{The ATT of stochastic multi-env policy divided by ATT of multi-env in sub-Figures~a and b, and divided by ATT of single-env policy in sub-Figures~c and d. The error bars represent 95\% confidence interval for ATT ratio. The closer ATT ratio is to one (red dashed line), the less degradation is caused as a result of using a stochastic multi-env model.}
    \label{fig:result-5rand-env-over-best-attendlight}
\end{figure}

Figures~\ref{fig:result-5rand-env-over-all-env-training} and \ref{fig:result-5rand-env-over-all-env-testing} show the ATT on stochastic multi-env divided by the ATT on multi-env regime, and the 95\% confidence interval among all traffic data of an intersection. We observe that most of the results (except two cases) are statistically equal. To do statistical analysis, the paired T-test on the hypothesis of equality of two regimes indicates that the average of means are statistically different at confidence level of 95\% for INT3 and INT8. In addition, at 90\% confidence level, the mean of two algorithms are statistically different for INT4 and INT11-3-phase. 

In addition, Figures~\ref{fig:result-5rand-env-over-best-attendlight-training} and \ref{fig:result-5rand-env-over-best-attendlight-testing} show the ATT ratio for stochastic multi-env regime over the ATT of single-env regime.
Considering the environment instances in the training set (Figure \ref{fig:result-5rand-env-over-best-attendlight-training}), there is an average of 16\% ATT degradation with the standard deviation of 0.15 (it was average of 15\% and standard deviation of 15\% in the multi-env regime). Similarly, on the test set (Figure~\ref{fig:result-5rand-env-over-best-attendlight-testing}) stochastic multi-env regime has 15\% ATT gap with the standard deviation of 0.22 (on multi-env regime it was 13\% and standard deviation of 19\%).

\begin{figure}
\centering
\begin{subfigure}{.19\textwidth}
    \centering
    \includegraphics[width=\textwidth]{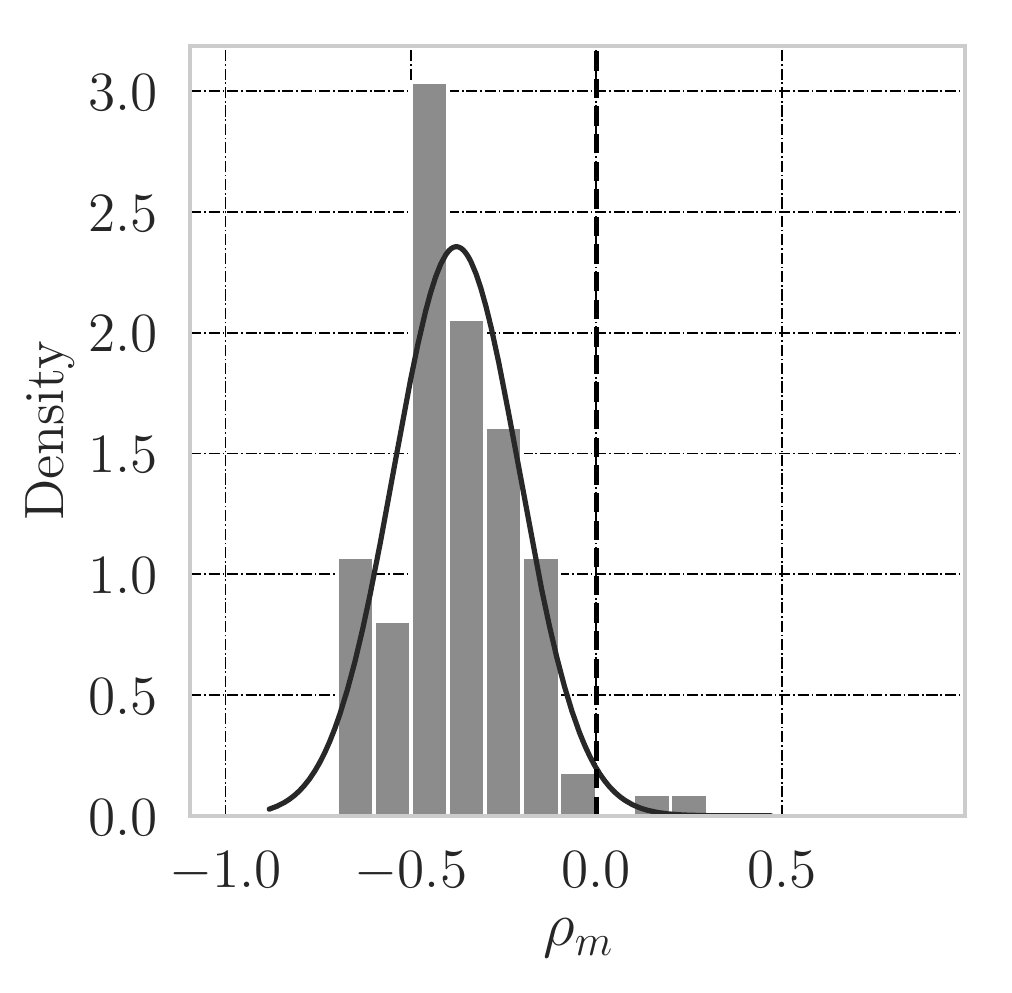}
    \caption{FixedTime}
    \label{fig:5rand-env-normal-dist-of-results-fixedtime}
\end{subfigure} 
\begin{subfigure}{.19\textwidth}
    \centering
    \includegraphics[width=\textwidth]{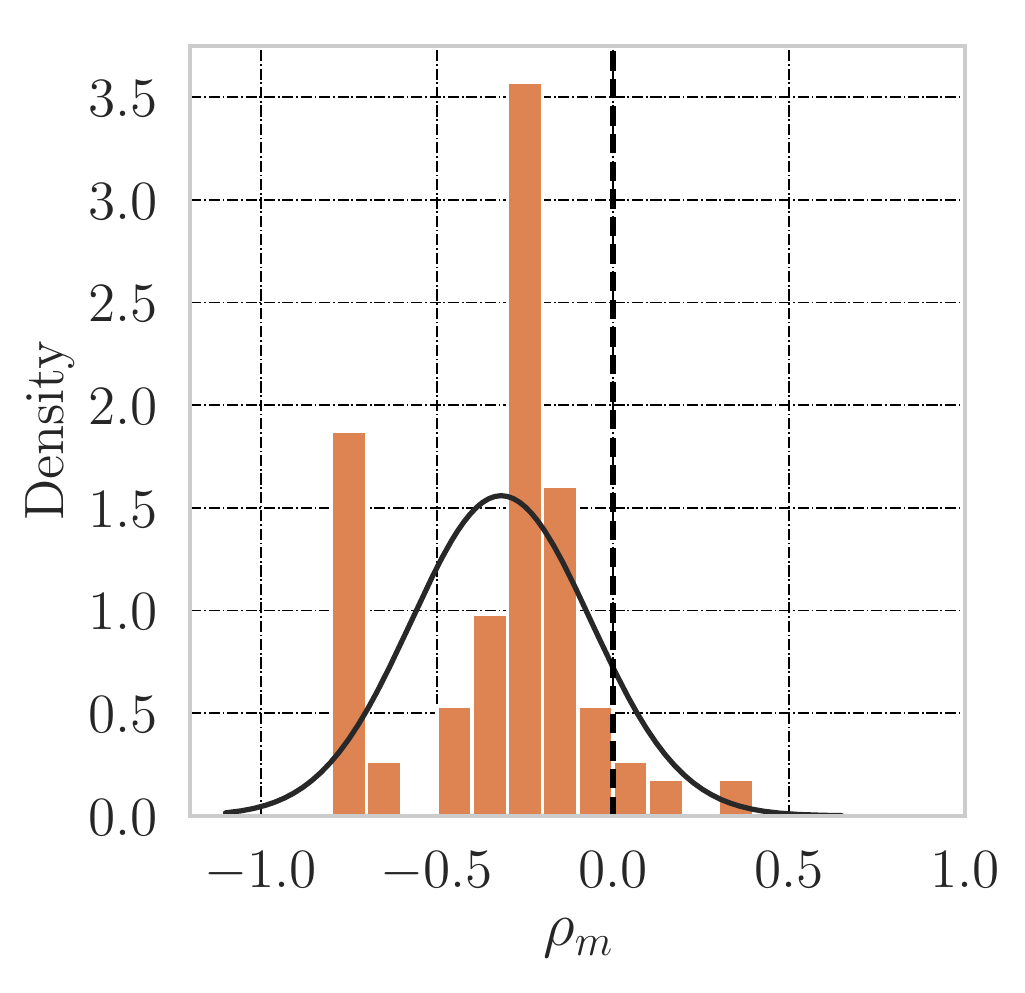}
    \caption{MaxPressure}
    \label{fig:5rand-env-normal-dist-of-results-maxpressure}
\end{subfigure}
\begin{subfigure}{.19\textwidth}
    \centering
    \includegraphics[width=\textwidth]{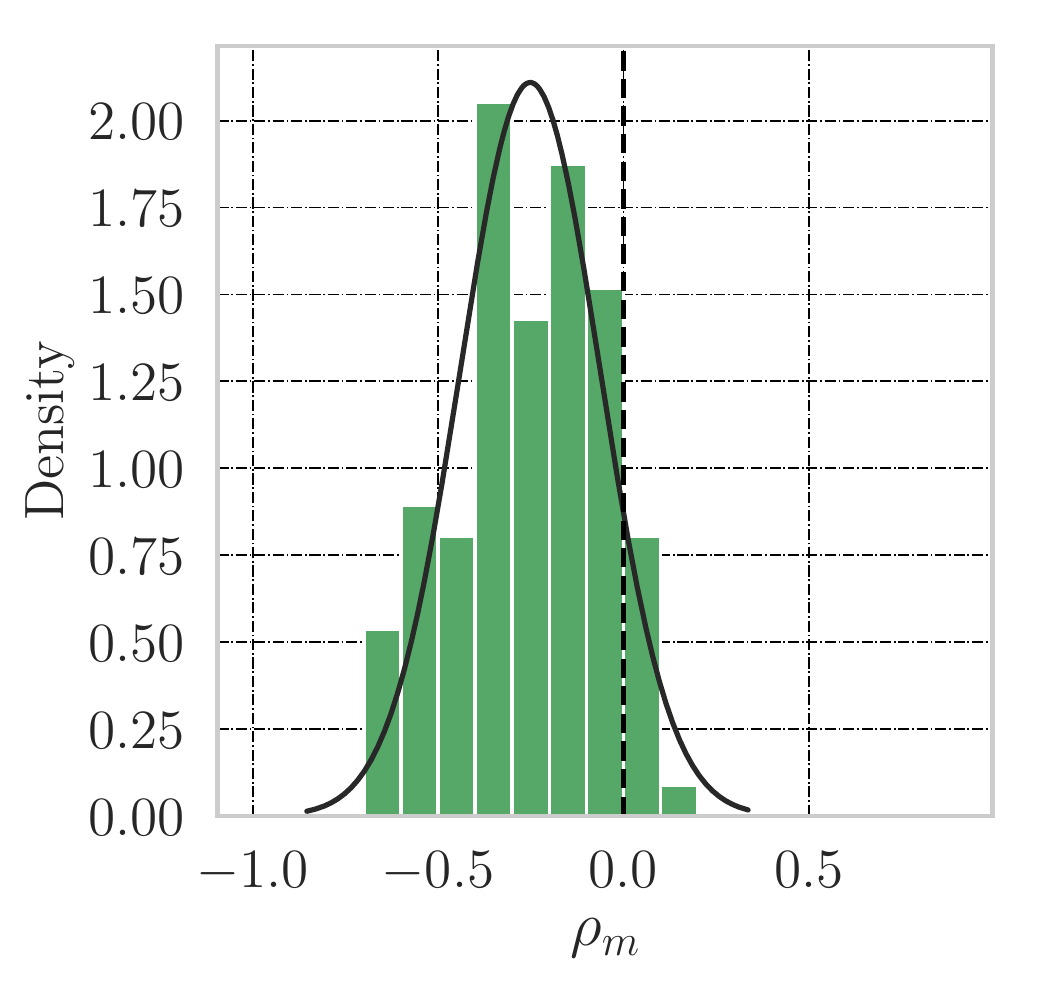}
    \caption{SOTL}
    \label{fig:5rand-env-normal-dist-of-results-sotl}
\end{subfigure} 
\begin{subfigure}{.19\textwidth}
    \centering
    \includegraphics[width=\textwidth]{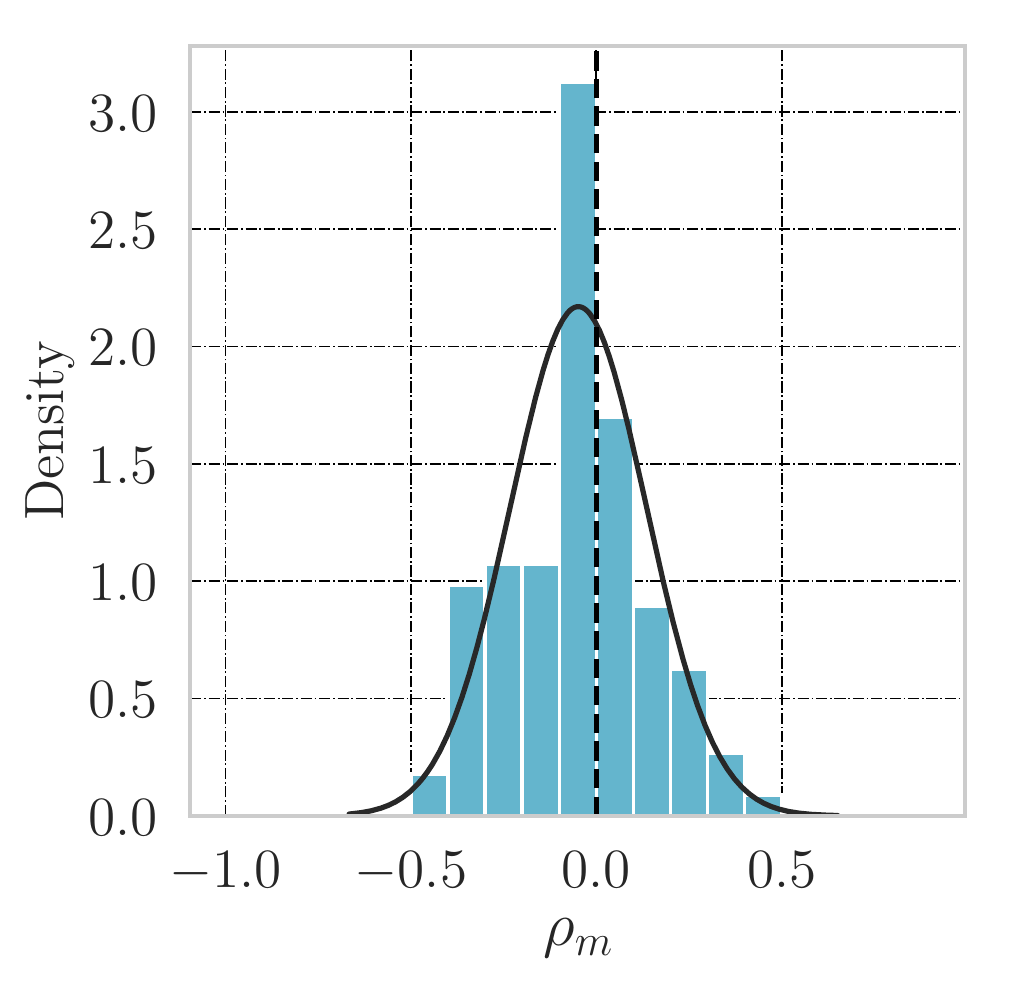}
    \caption{DQTSC-M}
    \label{fig:5rand-env-normal-dist-of-results-dqtscm}
\end{subfigure}
\begin{subfigure}{.19\textwidth}
    \centering
    \includegraphics[width=\textwidth]{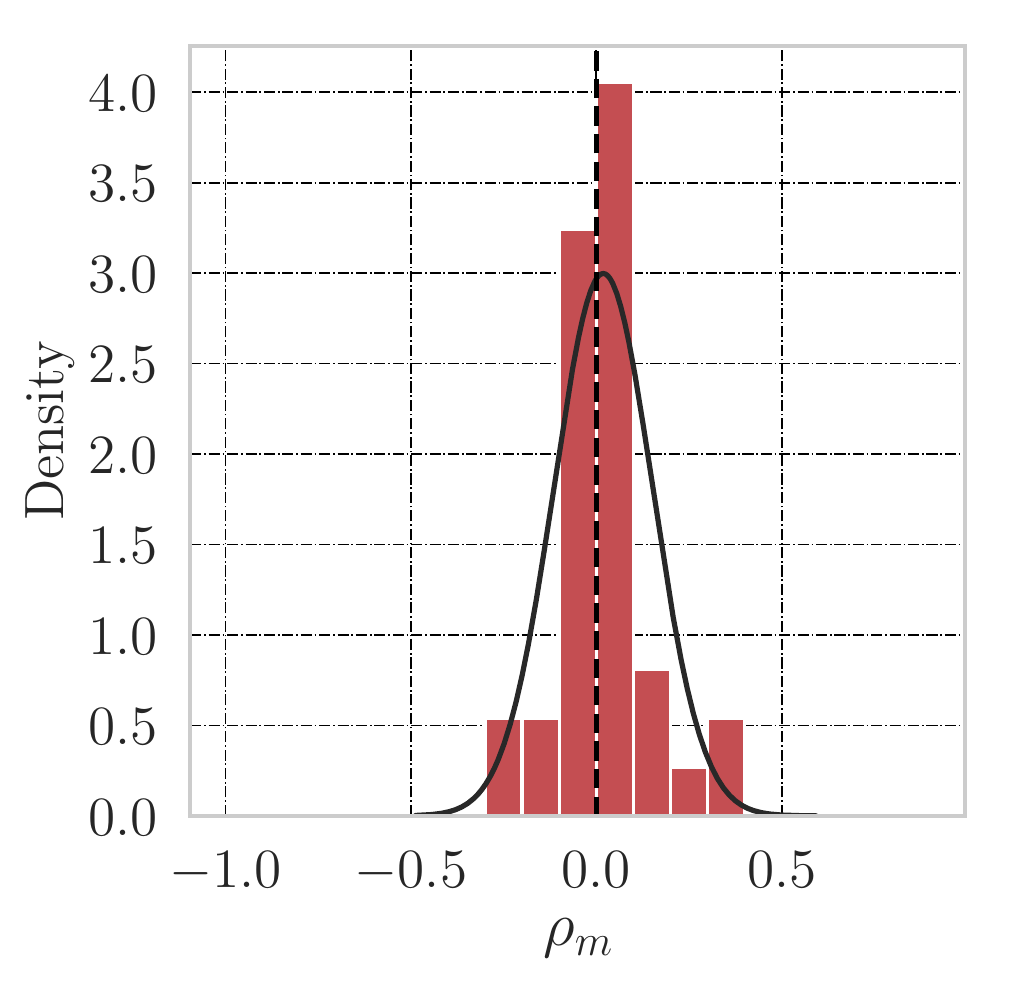}
    \caption{FRAP}
    \label{fig:5rand-env-normal-dist-of-results-frap}
\end{subfigure}
\caption{These plots illustrate the density of $\rho_m$ over all intersections. Here $\rho_m<0$ means that the stochastic multi-env outperforms the baseline algorithm. }
\label{fig:5rand-env-normal-dist-of-results}
\end{figure}

Compared to the benchmark algorithms, in average it achieves 36\%, 32\%, 27\%, and 2\% improvement over FixedTime, MaxPressure, SOTL, and DQTSC-M algorithms, respectively. Similar to the multi-env regime, FRAP gets smaller ATT, which here is 7\%. In addition, Figure~\ref{fig:5rand-env-normal-dist-of-results} shows the distribution of $\rho_m$ for all environment instances, suggesting that in most cases stochastic multi-env regime obtains smaller ATT compared to FixedTime, MaxPressure, and SOTL algorithms.

In summary, stochastic multi-env works well and in most cases obtains statistically equal results with multi-env regime. So, if the number of cases in the training set goes beyond the power available computation resources, the stochastic multi-env regime is a reliable substitute. 

\section{Environment Details}\label{appsec:env-detail}

In these section, we provide the visualization of 11 intersections and details of the 112 environment instances. 

\subsection{Intersection and Phase Visualizations}

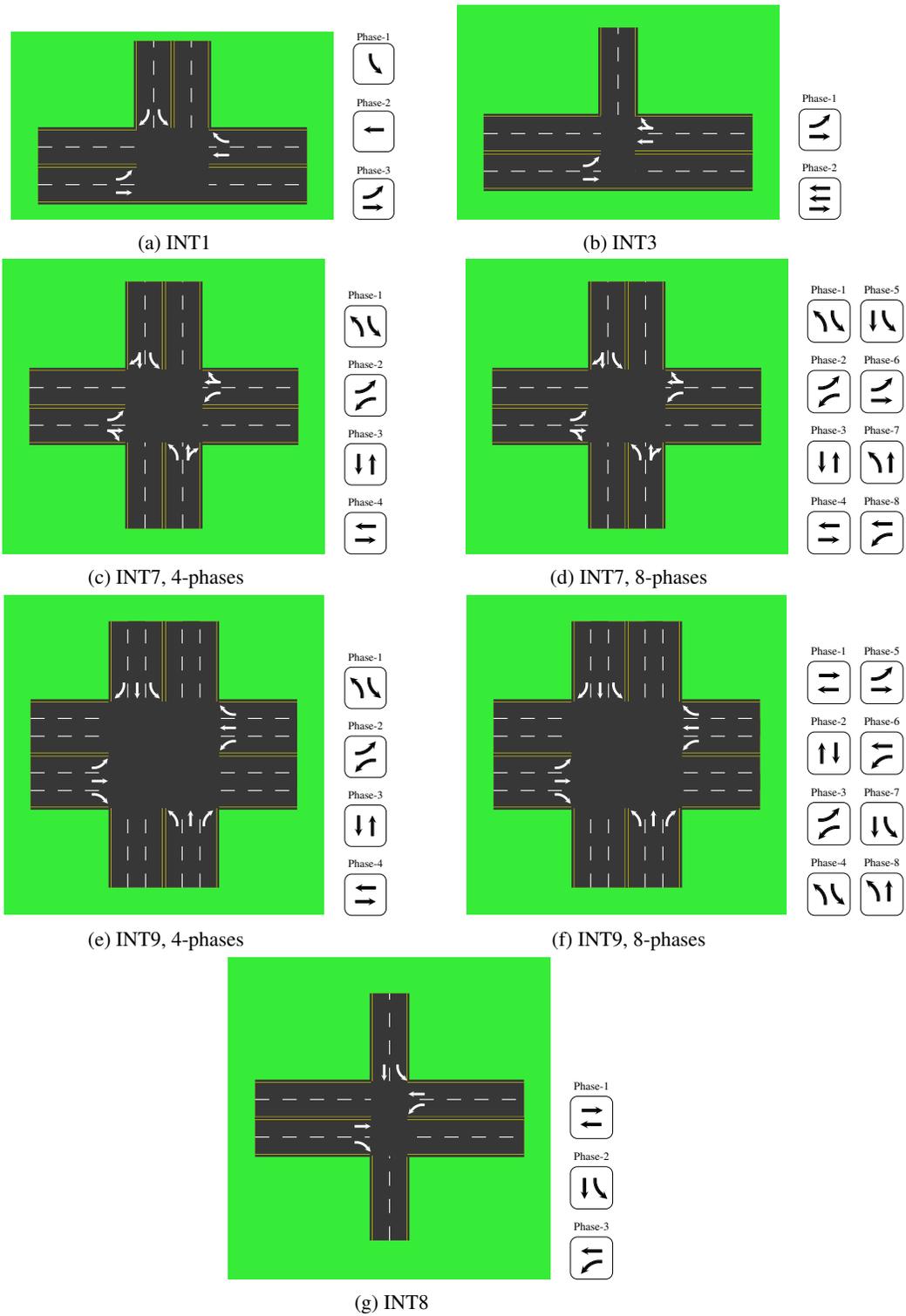
\begin{figure}[htbp]
\captionsetup[subfigure]{width=0.95\textwidth}
\centering
\begin{subfigure}[t]{.38\textwidth}
    \centering
    \resizebox{.94\textwidth}{!}{
        \rule[2cm]{-1.5cm}{.6cm}\input{intersection1_nolabel_app}
    }
    \caption{INT1}
    \label{fig-apdx:INT1}
\end{subfigure}
\begin{subfigure}[t]{.1\textwidth}
    \resizebox{.5\textwidth}{!}{
        \rule[2cm]{-1.2cm}{.6cm}\input{intersection1_traffic_3roads}
    }
\end{subfigure}%
\begin{subfigure}[t]{.38\textwidth}
    \centering
    \resizebox{.94\textwidth}{!}{
        \rule[2cm]{-1.5cm}{.6cm}\input{intersection4}
    }
    \caption{INT3}
    \label{fig-apdx:INT3}
\end{subfigure}
\begin{subfigure}[t]{.1\textwidth}
    \resizebox{.51\textwidth}{!}{
        \rule[2cm]{-1.2cm}{.6cm}\input{intersection4_phases}
    }
\end{subfigure}%

\begin{subfigure}[t]{.38\textwidth}
    \centering
    \resizebox{.94\textwidth}{!}{
        \rule[2cm]{-1.5cm}{.6cm}\input{intersection2}
    }
    \caption{INT7, 4-phases}
    \label{fig-apdx:INT7_4p}
\end{subfigure}
\begin{subfigure}[t]{.1\textwidth}
    \resizebox{.51\textwidth}{!}{
        \rule[2cm]{-1.2cm}{.6cm}\input{intersection2_phases_4}
    }
\end{subfigure}\hspace{5pt}
\begin{subfigure}[t]{.38\textwidth}
    \centering
    \resizebox{.94\textwidth}{!}{
        \rule[2cm]{-1.5cm}{.6cm}\input{intersection2}
    }
    \caption{INT7, 8-phases}
    \label{fig-apdx:INT7_8p}
\end{subfigure}
\begin{subfigure}[t]{.1\textwidth}
    \resizebox{1.1\textwidth}{!}{
        \rule[2cm]{-1.2cm}{.6cm}\input{intersection2_phases_8}
    }
\end{subfigure}%

\begin{subfigure}[t]{.38\textwidth}
    \centering
    \resizebox{.94\textwidth}{!}{
        \rule[2cm]{-1.5cm}{.6cm}\input{intersection3}
    }
    \caption{INT9, 4-phases}
    \label{fig-apdx:INT9_4p}
\end{subfigure}
\begin{subfigure}[t]{.1\textwidth}
    \resizebox{.51\textwidth}{!}{
        \rule[2cm]{-1.2cm}{.6cm}\input{intersection2_phases_4}
    }
\end{subfigure}\hspace{5pt}
\begin{subfigure}[t]{.38\textwidth}
    \centering
    \resizebox{.94\textwidth}{!}{
        \rule[2cm]{-1.5cm}{.6cm}\input{intersection3}
    }
    \caption{INT9, 8-phases}
    \label{fig-apdx:INT9_8p}
\end{subfigure}
\begin{subfigure}[t]{.1\textwidth}
    \resizebox{1.1\textwidth}{!}{
        \rule[2cm]{-1.2cm}{.6cm}\input{intersection3_phases}
    }
\end{subfigure}\hspace{5pt}

\begin{subfigure}[t]{.38\textwidth}
    \centering
    \resizebox{.94\textwidth}{!}{
        \rule[2cm]{-1.5cm}{.6cm}\input{intersection5}
    }
    \caption{INT8}
    \label{fig-apdx:INT8}
\end{subfigure}
\begin{subfigure}[t]{.1\textwidth}
    \resizebox{.52\textwidth}{!}{
        \rule[2cm]{-1.2cm}{.6cm}\input{intersection5_phases}
    }
\end{subfigure}\hspace{5pt}
\caption{All intersection topologies with their available phases which used for the training of the multi-env regime} 
\label{fig:all_training_intersections}
\end{figure}

\begin{figure}[htbp]
\captionsetup[subfigure]{width=0.95\textwidth}
\centering
\begin{subfigure}[t]{.38\textwidth}
    \centering
    \resizebox{.94\textwidth}{!}{
        \rule[2cm]{-1.5cm}{.6cm}\input{intersection1_nolabel_app}
    }
    \caption{INT2}
    \label{fig-apdx:INT2}
\end{subfigure}
\begin{subfigure}[t]{.1\textwidth}
    \resizebox{.5\textwidth}{!}{
        \rule[2cm]{-1.2cm}{.6cm}\input{intersection1_phases}
    }
\end{subfigure}\hspace{5pt}
\begin{subfigure}[t]{.38\textwidth}
    \centering
    \resizebox{.94\textwidth}{!}{
        \rule[2cm]{-1.5cm}{.6cm}\input{intersection9}
    }
    \caption{INT4}
    \label{fig-apdx:INT9}
\end{subfigure}
\begin{subfigure}[t]{.1\textwidth}
    \resizebox{.52\textwidth}{!}{
        \rule[2cm]{-1.2cm}{.6cm}\input{intersection9_phases}
    }
\end{subfigure}\hspace{5pt}
\begin{subfigure}[t]{.38\textwidth}
    \centering
    \resizebox{.94\textwidth}{!}{
        \rule[2cm]{-1.5cm}{.6cm}\input{intersection8}
    }
    \caption{INT5}
    \label{fig-apdx:INT5}
\end{subfigure}
\begin{subfigure}[t]{.1\textwidth}
    \resizebox{.52\textwidth}{!}{
        \rule[2cm]{-1.2cm}{.6cm}\input{intersection8_phases}
    }
\end{subfigure}\hspace{5pt}
\begin{subfigure}[t]{.38\textwidth}
    \centering
    \resizebox{.94\textwidth}{!}{
        \rule[2cm]{-1.5cm}{.6cm}\input{intersection7}
    }
    \caption{INT6}
    \label{fig-apdx:INT6}
\end{subfigure}
\begin{subfigure}[t]{.1\textwidth}
    \resizebox{.52\textwidth}{!}{
        \rule[2cm]{-1.2cm}{.6cm}\input{intersection7_phases}
    }
\end{subfigure}\hspace{5pt}
\begin{subfigure}[t]{.38\textwidth}
    \centering
    \resizebox{.94\textwidth}{!}{
        \rule[2cm]{-1.5cm}{.6cm}\input{intersection6}
    }
    \caption{INT10, 4-phases}
    \label{fig-apdx:INT10_4p}
\end{subfigure}
\begin{subfigure}[t]{.1\textwidth}
    \resizebox{.52\textwidth}{!}{
        \rule[2cm]{-1.2cm}{.6cm}\input{intersection2_phases_4}
    }
\end{subfigure}\hspace{5pt}
\begin{subfigure}[t]{.38\textwidth}
    \centering
    \resizebox{.94\textwidth}{!}{
        \rule[2cm]{-1.5cm}{.6cm}\input{intersection6}
    }
    \caption{INT10, 8-phases}
    \label{fig-apdx:INT10_8p}
\end{subfigure}
\begin{subfigure}[t]{.1\textwidth}
    \resizebox{1.1\textwidth}{!}{
        \rule[2cm]{-1.2cm}{.6cm}\input{intersection2_phases_8}
    }
\end{subfigure}\hspace{5pt}

\begin{subfigure}[t]{.38\textwidth}
    \centering
    \resizebox{.94\textwidth}{!}{
        \rule[2cm]{-1.5cm}{.6cm}\input{intersection11}
    }
    \caption{INT11, 4-phases}
    \label{fig-apdx:INT11_4p}
\end{subfigure}
\begin{subfigure}[t]{.1\textwidth}
    \resizebox{.52\textwidth}{!}{
        \rule[2cm]{-1.2cm}{.6cm}\input{intersection11_phases_3}
    }
\end{subfigure}\hspace{5pt}
\begin{subfigure}[t]{.38\textwidth}
    \centering
    \resizebox{.94\textwidth}{!}{
        \rule[2cm]{-1.5cm}{.6cm}\input{intersection11}
    }
    \caption{INT11, 8-phases}
    \label{fig-apdx:INT11_8p}
\end{subfigure}
\begin{subfigure}[t]{.1\textwidth}
    \resizebox{1.1\textwidth}{!}{
        \rule[2cm]{-1.2cm}{.6cm}\input{intersection11_phases_5}
    }
\end{subfigure}\hspace{5pt}

\caption{All intersection topologies with their available phases which used for testing the multi-env regime} 
\label{fig:all_testing_intersections}
\end{figure}
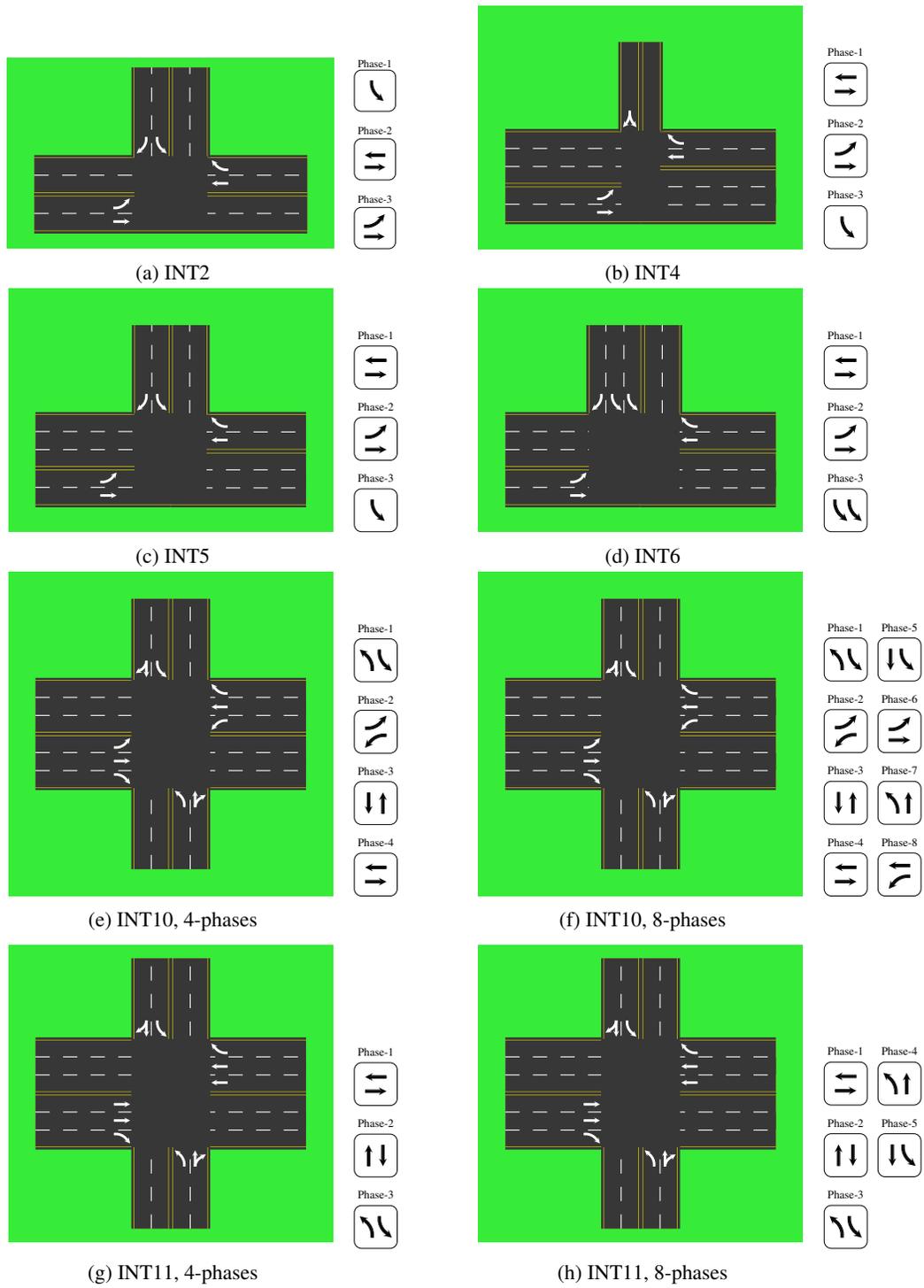

Figure \ref{fig:all_training_intersections} and \ref{fig:all_testing_intersections} show the visualization of all 11 intersections. The intersections in Figure~\ref{fig:all_training_intersections} used to train the multi-env regime and the intersections in Figure~\ref{fig:all_testing_intersections} are used in the testing of the multi-env regime.

In order to analyze the performance of the trained model on the same intersection with different set of phases, we consider intersection of Figure \ref{fig-apdx:INT1} in the set of training environments and the intersection of Figure \ref{fig-apdx:INT2} in the test set.

\subsection{Description of Each Environment Instance}\label{sec:apdx:description_of_all_cases}

Tables \ref{tb:details_of_training_cases} and \ref{tb:details_of_testing_cases} provides the details of all training and testing problems. 

\begin{table}[htbp]
\centering
\caption{Intersections used for the training of the multi-env regime}
    \centering
        \begin{adjustbox}{width=1\textwidth}
\begin{tabular}{lccccccccc}
\toprule
case         & {\tt INT7-H1-4} & {\tt INT7-H1-8} & {\tt INT1-H1-3} & {\tt INT3-H1-2} & {\tt INT7-A1-8} & {\tt INT7-A2-8}     & {\tt INT7-A3-8}     & {\tt INT7-A4-8} & {\tt INT7-A5-8} \\ \midrule
Intersection & {\tt INT7     } & {\tt INT7     } & {\tt INT1     } & {\tt INT3     } & {\tt INT7     } & {\tt INT7     }     & {\tt INT7     }     & {\tt INT7     } & {\tt INT7     } \\
\#phase      & 4         & 8         & 3         & 2         & 8         & 8             & 8             & 8         & 8         \\
traffic      & {\tt H1}        & {\tt H1}        & {\tt H1}        & {\tt H1}        & {\tt A1}        & {\tt A2}            & {\tt A3}            & {\tt A4}        & {\tt A5}        \\
\#roads      & 4         & 4         & 3         & 3         & 4         & 4             & 4             & 4         & 4         \\
\#lane       & 2         & 2         & 2         & 2         & 2         & 2             & 2             & 2         & 2         \\ \midrule
case         & {\tt INT7-A1-4} & {\tt INT7-A2-4} & {\tt INT7-A3-4} & {\tt INT7-A4-4} & {\tt INT7-A5-4} & {\tt INT8-H1-3}     & {\tt INT9-S6-4}     & {\tt INT9-S6-8} & {\tt INT1-A1-3} \\ \midrule
Intersection & {\tt INT7     } & {\tt INT7     } & {\tt INT7     } & {\tt INT7     } & {\tt INT7     } & {\tt INT8     }     & {\tt INT9     }     & {\tt INT9     } & {\tt INT1     } \\
\#phase      & 4         & 4         & 4         & 4         & 4         & 3             & 4             & 8         & 3         \\
traffic      & {\tt A1}        & {\tt A2}        & {\tt A3}        & {\tt A4}        & {\tt A5}        & {\tt H1}            & {\tt S6}            & {\tt S6}        & {\tt A1}        \\
\#roads      & 4         & 4         & 4         & 4         & 4         & 4             & 4             & 4         & 3         \\
\#lane       & 2         & 2         & 2         & 2         & 2         & 2             & 3             & 3         & 2         \\ \midrule
case         & {\tt INT1-A2-3} & {\tt INT1-A3-3} & {\tt INT1-A4-3} & {\tt INT1-A5-3} & {\tt INT3-A1-2} & {\tt INT3-A2-2}     & {\tt INT3-A3-2}     & {\tt INT3-A4-2} & {\tt INT3-A5-2} \\ \midrule
Intersection & {\tt INT1     } & {\tt INT1     } & {\tt INT1     } & {\tt INT1     } & {\tt INT3     } & {\tt INT3     }     & {\tt INT3     }     & {\tt INT3     } & {\tt INT3     } \\
\#phase      & 3         & 3         & 3         & 3         & 2         & 2             & 2             & 2         & 2         \\
traffic      & {\tt A2}        & {\tt A3}        & {\tt A4}        & {\tt A5}        & {\tt A1}        & {\tt A2}            & {\tt A3}            & {\tt A4}        & {\tt A5}        \\
\#roads      & 3         & 3         & 3         & 3         & 3         & 3             & 3             & 3         & 3         \\
\#lane       & 2         & 2         & 2         & 2         & 2         & 2             & 2             & 2         & 2         \\ \midrule
case         & {\tt INT8-A1-3} & {\tt INT8-A2-3} & {\tt INT8-A3-3} & {\tt INT8-A4-3} & {\tt INT8-A5-3} & {\tt INT9-S1-4}     & {\tt INT9-S2-4}     & {\tt INT9-S3-4} & {\tt INT9-S4-4} \\ \midrule
Intersection & {\tt INT8     } & {\tt INT8     } & {\tt INT8     } & {\tt INT8     } & {\tt INT8     } & {\tt INT9     }     & {\tt INT9     }     & {\tt INT9     } & {\tt INT9     } \\
\#phase      & 3         & 3         & 3         & 3         & 3         & 4             & 4             & 4         & 4         \\
traffic      & {\tt A1}        & {\tt A2}        & {\tt A3}        & {\tt A4}        & {\tt A5}        & {\tt S1}            & {\tt S2}            & {\tt S3}        & {\tt S4}        \\
\#roads      & 4         & 4         & 4         & 4         & 4         & 4             & 4             & 4         & 4         \\
\#lane       & 2         & 2         & 2         & 2         & 2         & 3             & 3             & 3         & 3         \\ \midrule
case         & {\tt INT9-S5-4} & {\tt INT9-S4-8} & {\tt INT9-S5-8} & {\tt INT9-S1-8} & {\tt INT9-S2-8} & \multicolumn{2}{l}{INT9-S3-8} &           &           \\ \midrule
Intersection & {\tt INT9     } & {\tt INT9     } & {\tt INT9     } & {\tt INT9     } & {\tt INT9     } & {\tt INT9     }     &               &           &           \\
\#phase      & 4         & 8         & 8         & 8         & 8         & 8             &               &           &           \\
traffic      & {\tt S5}        & {\tt S4}        & {\tt S5}        & {\tt S1}        & {\tt S2}        & {\tt S3}            &               &           &           \\
\#roads      & 4         & 4         & 4         & 4         & 4         & 4             &               &           &           \\
\#lane       & 3         & 3         & 3         & 3         & 3         & 3             &               &           &          \\
\bottomrule
\end{tabular}
\end{adjustbox}
\label{tb:details_of_training_cases}
\end{table}

\begin{table}[htbp]
    \caption{Intersections used for the testing of the multi-env regime}
    \label{tb:details_of_testing_cases}
    \centering
        \begin{adjustbox}{width=1\textwidth}
\begin{tabular}{lccccccccc}
\toprule
case         & {\tt INT7-H2-4}  & {\tt INT7-H3-4}  & {\tt INT7-H4-4}  & {\tt INT7-H5-4}  & {\tt INT7-H2-8}  & {\tt INT7-H3-8}  & {\tt INT7-H4-8}  & {\tt INT7-H5-8}  & {\tt INT2-A4-3}  \\ \midrule
Intersection & {\tt INT7     }  & {\tt INT7     }  & {\tt INT7     }  & {\tt INT7     }  & {\tt INT7     }  & {\tt INT7     }  & {\tt INT7     }  & {\tt INT7     }  & {\tt INT2     }  \\
\#phase      & 4          & 4          & 4          & 4          & 8          & 8          & 8          & 8          & 3          \\
traffic      & {\tt H2}         & {\tt H3}         & {\tt H4}         & {\tt H5}         & {\tt H2}         & {\tt H3}         & {\tt H4}         & {\tt H5}         & {\tt A4}         \\
\#roads      & 4          & 4          & 4          & 4          & 4          & 4          & 4          & 4          & 3          \\
\#lane       & 2          & 2          & 2          & 2          & 2          & 2          & 2          & 2          & 2          \\ \midrule
case         & {\tt INT2-A3-3}  & {\tt INT2-A5-3}  & {\tt INT2-A1-3}  & {\tt INT2-H1-3}  & {\tt INT4-A4-3}  & {\tt INT4-A3-3}  & {\tt INT4-A5-3}  & {\tt INT4-A1-3}  & {\tt INT4-H1-3}  \\ \midrule
Intersection & {\tt INT2     }  & {\tt INT2     }  & {\tt INT2     }  & {\tt INT2     }  & {\tt INT4     }  & {\tt INT4     }  & {\tt INT4     }  & {\tt INT4     }  & {\tt INT4     }  \\
\#phase      & 3          & 3          & 3          & 3          & 3          & 3          & 3          & 3          & 3          \\
traffic      & {\tt A3}         & {\tt A5}         & {\tt A1}         & {\tt H1}         & {\tt A4}         & {\tt A3}         & {\tt A5}         & {\tt A1}         & {\tt H1}         \\
\#roads      & 3          & 3          & 3          & 3          & 3          & 3          & 3          & 3          & 3          \\
\#lane       & 2          & 2          & 2          & 2          & 3          & 3          & 3          & 3          & 3          \\ \midrule
case         & {\tt INT11-S2-3} & {\tt INT11-S4-3} & {\tt INT11-S3-3} & {\tt INT11-S5-3} & {\tt INT11-S1-3} & {\tt INT11-S6-3} & {\tt INT11-S2-5} & {\tt INT11-S4-5} & {\tt INT11-S3-5} \\ \midrule
Intersection & {\tt INT11     } & {\tt INT11     } & {\tt INT11     } & {\tt INT11     } & {\tt INT11     } & {\tt INT11     } & {\tt INT11     } & {\tt INT11     } & {\tt INT11     } \\
\#phase      & 3          & 3          & 3          & 3          & 3          & 3          & 5          & 5          & 5          \\
traffic      & {\tt S2}         & {\tt S4}         & {\tt S3}         & {\tt S5}         & {\tt S1}         & {\tt S6}         & {\tt S2}         & {\tt S4}         & {\tt S3}         \\
\#roads      & 4          & 4          & 4          & 4          & 4          & 4          & 4          & 4          & 4          \\
\#lane       & 3          & 3          & 3          & 3          & 3          & 3          & 3          & 3          & 3          \\ \midrule
case         & {\tt INT11-S5-5} & {\tt INT11-S1-5} & {\tt INT11-S6-5} & {\tt INT10-S2-4} & {\tt INT10-S4-4} & {\tt INT10-S3-4} & {\tt INT10-S5-4} & {\tt INT10-S1-4} & {\tt INT10-S6-4} \\ \midrule
Intersection & {\tt INT11     } & {\tt INT11     } & {\tt INT11     } & {\tt INT10     } & {\tt INT10     } & {\tt INT10     } & {\tt INT10     } & {\tt INT10     } & {\tt INT10     } \\
\#phase      & 5          & 5          & 5          & 4          & 4          & 4          & 4          & 4          & 4          \\
traffic      & {\tt S5}         & {\tt S1}         & {\tt S6}         & {\tt S2}         & {\tt S4}         & {\tt S3}         & {\tt S5}         & {\tt S1}         & {\tt S6}         \\
\#roads      & 4          & 4          & 4          & 4          & 4          & 4          & 4          & 4          & 4          \\
\#lane       & 3          & 3          & 3          & 3          & 3          & 3          & 3          & 3          & 3          \\ \midrule
case         & {\tt INT10-S2-8} & {\tt INT10-S4-8} & {\tt INT10-S3-8} & {\tt INT10-S5-8} & {\tt INT10-S1-8} & {\tt INT10-S6-8} & {\tt INT4-S1-3}  & {\tt INT4-S2-3}  & {\tt INT4-S3-3}  \\ \midrule
Intersection & {\tt INT10     } & {\tt INT10     } & {\tt INT10     } & {\tt INT10     } & {\tt INT10     } & {\tt INT10     } & {\tt INT4     }  & {\tt INT4     }  & {\tt INT4     }  \\
\#phase      & 8          & 8          & 8          & 8          & 8          & 8          & 3          & 3          & 3          \\
traffic      & {\tt S2}         & {\tt S4}         & {\tt S3}         & {\tt S5}         & {\tt S1}         & {\tt S6}         & {\tt S1}         & {\tt S2}         & {\tt S3}         \\
\#roads      & 4          & 4          & 4          & 4          & 4          & 4          & 4          & 4          & 4          \\
\#lane       & 3          & 3          & 3          & 3          & 3          & 3          & 3          & 3          & 3          \\ \midrule
case         & {\tt INT4-S4-3}  & {\tt INT4-S5-3}  & {\tt INT4-S6-3}  & {\tt INT5-A1-3}  & {\tt INT5-A3-3}  & {\tt INT5-A4-3}  & {\tt INT5-A5-3}  & {\tt INT5-H1-3}  & {\tt INT5-S1-3}  \\ \midrule
Intersection & {\tt INT4     }  & {\tt INT4     }  & {\tt INT4     }  & {\tt INT5     }  & {\tt INT5     }  & {\tt INT5     }  & {\tt INT5     }  & {\tt INT5     }  & {\tt INT5     }  \\
\#phase      & 3          & 3          & 3          & 3          & 3          & 3          & 3          & 3          & 3          \\
traffic      & {\tt S4}         & {\tt S5}         & {\tt S6}         & {\tt A1}         & {\tt A3}         & {\tt A4}         & {\tt A5}         & {\tt H1}         & {\tt S1}         \\
\#roads      & 4          & 4          & 4          & 4          & 4          & 4          & 4          & 4          & 4          \\
\#lane       & 3          & 3          & 3          & 3          & 3          & 3          & 3          & 3          & 3          \\ \midrule
case         & {\tt INT5-S2-3}  & {\tt INT5-S3-3}  & {\tt INT5-S4-3}  & {\tt INT5-S5-3}  & {\tt INT5-S6-3}  & {\tt INT6-A1-3}  & {\tt INT6-A3-3}  & {\tt INT6-A4-3}  & {\tt INT6-A5-3}  \\ \midrule
Intersection & {\tt INT5     }  & {\tt INT5     }  & {\tt INT5     }  & {\tt INT5     }  & {\tt INT5     }  & {\tt INT6     }  & {\tt INT6     }  & {\tt INT6     }  & {\tt INT6     }  \\
\#phase      & 3          & 3          & 3          & 3          & 3          & 3          & 3          & 3          & 3          \\
traffic      & {\tt S2}         & {\tt S3}         & {\tt S4}         & {\tt S5}         & {\tt S6}         & {\tt A1}         & {\tt A3}         & {\tt A4}         & {\tt A5}         \\
\#roads      & 4          & 4          & 4          & 4          & 4          & 4          & 4          & 4          & 4          \\
\#lane       & 3          & 3          & 3          & 3          & 3          & 3          & 3          & 3          & 3          \\ \midrule
case         & {\tt INT6-H1-3}  & {\tt INT6-S1-3}  & {\tt INT6-S2-3}  & {\tt INT6-S3-3}  & {\tt INT6-S4-3}  & {\tt INT6-S5-3}  & {\tt INT6-S6-3}  &            &            \\ \midrule
Intersection & {\tt INT6     }  & {\tt INT6     }  & {\tt INT6     }  & {\tt INT6     }  & {\tt INT6     }  & {\tt INT6     }  & {\tt INT6     }  &            &            \\
\#phase      & 3          & 3          & 3          & 3          & 3          & 3          & 3          &            &            \\
traffic      & {\tt H1}         & {\tt S1}         & {\tt S2}         & {\tt S3}         & {\tt S4}         & {\tt S5}         & {\tt S6}         &            &            \\
\#roads      & 4          & 4          & 4          & 4          & 4          & 4          & 4          &            &            \\
\#lane       & 3          & 3          & 3          & 3          & 3          & 3          & 3          &            &            \\
\bottomrule
\end{tabular}
\end{adjustbox}
\end{table}

\section{Other Application Area for Future Research}\label{apdx:other_application_area}

\subsection{Assemble-to-Order Systems}
This problem refers to the situation where parts of different types need to be assembled into a few known finished products, and the goal is to find the sequence of the products to assemble. Since the number and type of input parts may vary among different products, the problem does not have a fixed-sized input. Moreover, the number of orders at each time changes due to the stochasticity of the customer’s demand. So, the number of output products does not have a fixed-size. 

AttendLight can be utilized to solve this problem. The state-attention extracts the state of each product, reflecting the availability of required parts, numbers of possible finished products, etc. The action-attention is responsible for deciding what to produce at each time with the goal of minimizing objectives total make-span, tardiness, and delay time. 


\subsection{Dynamic Matching Problem} 
In this problem, entities of different types arrive at the system and wait in the queue until they are matched together. Once a matching happens, the entities immediately leave the system and some feedback is observed in terms of a reward signal. The objective is to maximize the long-term cumulative reward while keeping the system stable. This problem is a generalization of the assemble-to-order systems, where the entities can be humans, advertisements, commodities, etc. The AttendLight model can be used in this problem to first, extract the state of each matching option from the system state, and then the action attention chooses the matchings with the highest value.

\subsection{Wireless Resource Allocation} 
In this problem, the goal is to efficiently allocate the spectrum and power of the wireless router to its users. Each user may send or receive packets of different sizes. Users and as well as packets may have different priorities. Therefore, the main task is to send the packets considering their priority, size, arrival time, etc. to minimize a relevant cost function such as the queue length or the total response time of each request. AttendLight can solve this problem: first, state-attention learns to extract the state of each packet with considering the arrival time, priority, size of the packet, type of the data, and other information that may not be available for all packets. Then, the action-attention decides to send which packet next. 

\end{document}

%% file: intersection1.tex
\definecolor{LightGreen}{HTML}{CEFBCE}
\definecolor{LightRed}{HTML}{FFDDDD}

\input{tikz_utils}

\begin{tikzpicture}[]
\node [bg, minimum height =7cm, minimum width=12cm] at (.0, 1.5) {} ;
\draw[road2 node] (-5cm,.7cm)  -- (5cm,.7cm);
\draw[road2 node] (-5cm,-.7cm)  -- (5cm,-.7cm);
\def\shiftr{18.5pt}
\draw[road2 node] (-.7cm, .7cm+\shiftr ) -- (-.7,4cm +\shiftr);
\draw[road2 node] (.7cm, .7cm+\shiftr) -- (.7cm,4cm+ +\shiftr);
\node[fill=way, minimum height=2.7cm, minimum width=2.66cm] at (0,0.05) {};

\leftturn(-2.1,-0.5){0};
\straight(-2.1,-1){0};
\straight(2.1,0.4){180};
\rightturn(2.1,.9){180};
\leftturn(-0.5, 2.1){270};
\rightturn(-.9,2.1){270};

\pic[scale=.25,fill=gray,text=black] at (-4,3.5) {compass};

\node[text=LightGreen] at (-4.3,-.4) {\Large $l^{in}_1$};
\node[text=LightGreen] at (-4.3,-1) {\Large $l^{in}_2$};
\node[text=LightGreen] at (4.3,.4) {\Large $l^{in}_3$};
\node[text=LightGreen] at (4.3,1.03) {\Large $l^{in}_4$};
\node[text=LightGreen, rotate=-90] at (-0.38,4) {\Large $l^{in}_5$};
\node[text=LightGreen, rotate=-90] at (-1.03,4) {\Large $l^{in}_6$};

\node[text=LightRed] at (4.3,-.38) {\Large $l^{out}_4$};
\node[text=LightRed] at (4.3,-1) {\Large $l^{out}_3$};
\node[text=LightRed] at (-4.3,.4) {\Large $l^{out}_2$};
\node[text=LightRed] at (-4.3,1) {\Large $l^{out}_1$};
\node[text=LightRed, rotate=-90] at (0.4,3.9) {\Large $l^{out}_6$};
\node[text=LightRed, rotate=-90] at (1.05,3.9) {\Large $l^{out}_5$};
share
\end{tikzpicture}

%% file: intersection1_traffic.tex
\definecolor{LightGreen}{HTML}{04300B}
\definecolor{LightRed}{HTML}{580101}

\input{phase_utils}

\begin{tikzpicture}[]
\node[] (ph1_txt) at (-0,2.0) {\footnotesize Phase-1};
\node [wbg, minimum height =1.2cm, minimum width=1.2cm,rounded corners=0.2cm] (ph1_sign) at (.0, 1.2) {} ;
\leftturn(.10, 1.50cm){270};
\rightturn(-.1, 1.50cm){270};

\node [wbg, minimum height =1.2cm, minimum width=4.5cm, rounded corners=0.2cm] (ph1_tr) at (3.25, 1.2) {} ;
\node[] at (3.25,1.50) { $v_6 = \{{\color{LightGreen}l^{in}_6} \to {\color{LightRed}l^{out}_1}, \, {\color{LightGreen}l^{in}_6} \to {\color{LightRed}l^{out}_2}\}$};

\node[] at (3.25,1) { $v_5 = \{{\color{LightGreen}l^{in}_5} \to {\color{LightRed}l^{out}_3}, \, {\color{LightGreen}l^{in}_5} \to {\color{LightRed}l^{out}_4}\}$};

\draw[dotted] ([xshift=-.1cm,yshift=-.01cm]ph1_sign.north east) -- ([xshift=.1cm,yshift=-.01cm]ph1_tr.north west);
\draw[dotted] ([xshift=-.1cm,yshift=.01cm]ph1_sign.south east) -- ([xshift=.1cm,yshift=.01cm]ph1_tr.south west);

\node[]  (ph2_txt) at (-.0,.05) {\footnotesize Phase-2};
\node [wbg, minimum height =1.2cm, minimum width=1.2cm,rounded corners=0.2cm] (ph2_sign) at (.0, -0.8) {} ;
\rightturn(.3,-0.7){180};
\straight(.3,-0.98){180};
\straight(-.3,-1.23){0};

\node [wbg, minimum height =1.7cm, minimum width=4.5cm, rounded corners=0.2cm] (ph2_tr) at (3.2, -0.8) {} ;
\node[] at (3.25,-0.3) { $v_4 = \{{\color{LightGreen}l^{in}_4} \to {\color{LightRed}l^{out}_5}, \, {\color{LightGreen}l^{in}_4} \to {\color{LightRed}l^{out}_6}\}$};

\node[] at (3.25,-.8) { $v_3 = \{{\color{LightGreen}l^{in}_3} \to {\color{LightRed}l^{out}_1}, \, {\color{LightGreen}l^{in}_3} \to {\color{LightRed}l^{out}_2}\}$};

\node[] at (3.25,-1.3) { $v_2 = \{{\color{LightGreen}l^{in}_2} \to {\color{LightRed}l^{out}_3}, \, {\color{LightGreen}l^{in}_2} \to {\color{LightRed}l^{out}_4}\}$};

\draw[dotted] ([xshift=-.1cm,yshift=-.01cm]ph2_sign.north east) -- ([xshift=.1cm,yshift=-.01cm]ph2_tr.north west);
\draw[dotted] ([xshift=-.1cm,yshift=.01cm]ph2_sign.south east) -- ([xshift=.1cm,yshift=.01cm]ph2_tr.south west);

\node[] (ph3_txt) at (-0,-1.95) {\footnotesize Phase-3};
\node (ph3_sign) [wbg, minimum height =1.2cm, minimum width=1.2cm,rounded corners=0.2cm] at (.0, -2.8) {} ;
\leftturn(-.320, -2.75cm){0};
\straight(-.320, -3.05cm){0};

\node [wbg, minimum height =1.2cm, minimum width=4.5cm, rounded corners=0.2cm] (ph3_tr) at (3.2, -2.8) {} ;
\node[] at (3.25,-2.5) { $v_1 = \{{\color{LightGreen}l^{in}_1} \to {\color{LightRed}l^{out}_5}, \, {\color{LightGreen}l^{in}_1} \to {\color{LightRed}l^{out}_6}\}$};

\node[] at (3.25,-3) { $v_2 = \{{\color{LightGreen}l^{in}_2} \to {\color{LightRed}l^{out}_3}, \, {\color{LightGreen}l^{in}_2} \to {\color{LightRed}l^{out}_4}\}$};

\draw[dotted] ([xshift=-.1cm,yshift=-.01cm]ph3_sign.north east) -- ([xshift=.1cm,yshift=-.01cm]ph3_tr.north west);
\draw[dotted] ([xshift=-.1cm,yshift=.01cm]ph3_sign.south east) -- ([xshift=.1cm,yshift=.01cm]ph3_tr.south west);





share
\end{tikzpicture}

%% file: att_model_2.tex
\usetikzlibrary{arrows.meta}
\usetikzlibrary{intersections}

\definecolor{LightGreen}{HTML}{000000}
\definecolor{LightRed}{HTML}{000000}
\definecolor{embcolor}{HTML}{BEE8FF} 
\definecolor{inpattcolor}{HTML}{FDD3FE} 
\definecolor{rnn1color}{HTML}{91FFF3} 
\definecolor{rnn2color}{HTML}{D6FFFA} 
\definecolor{phattcolor}{HTML}{FDA2FF} 

\input{box_utils}
\input{att_utils}
\input{phase_utils}


\begin{tikzpicture}[>={LaTeX[width=1.5mm,length=1.5mm,angle'=45]}]
    \coordinate (toporig) at ( 0.0, 0.0);
    
    \node [text=LightGreen]
    at([xshift=-0.8cm,yshift=.1cm]toporig.south) {\large$s_1^{t}$};
    \tikzcuboid {%
        shiftx=0cm,%
        shifty=0cm,%
        scale=.30,%
        rotation=0,%
        densityx=1,%
        densityy=1,%
        densityz=1,%
        dimx=4,%
        dimy=1,%
        dimz=1,%
        linefront=green!75!black,%
        linetop=green!50!black,%
        lineright=green!25!black,%
        fillfront=green!25!white,%
        filltop=green!50!white,%
        fillright=green!75!white,%
        emphedge=Y,%
        emphstyle=very thin,
    }
    \coordinate (o1) at ( 0.0, 0.0);
    
    \node [text=LightGreen] at([xshift=-0.8cm,yshift=-.9cm]toporig.south) {\large$s_2^{t}$};
    \tikzcuboid{%
        shiftx=0cm,%
        shifty=-1cm,%
        scale=.30,%
        rotation=0,%
        densityx=1,%
        densityy=1,%
        densityz=1,%
        dimx=4,%
        dimy=1,%
        dimz=1,%
        linefront=green!75!black,%
        linetop=green!50!black,%
        lineright=green!25!black,%
        fillfront=green!25!white,%
        filltop=green!50!white,%
        fillright=green!75!white,%
        emphedge=Y,%
        emphstyle=very thin,
    }
    \coordinate (o2) at ( 0.0, -1.0);
    
    \node [text=LightGreen] at([xshift=-0.8cm,yshift=-1.9cm]toporig.south) {\large$s_3^{t}$};
    \tikzcuboid{%
        shiftx=0cm,%
        shifty=-2cm,%
        scale=.30,%
        rotation=0,%
        densityx=1,%
        densityy=1,%
        densityz=1,%
        dimx=4,%
        dimy=1,%
        dimz=1,%
        linefront=green!75!black,%
        linetop=green!50!black,%
        lineright=green!25!black,%
        fillfront=green!25!white,%
        filltop=green!50!white,%
        fillright=green!75!white,%
        emphedge=Y,%
        emphstyle=very thin,
    }
    \coordinate (o3) at ( 0.0, -2.0);

    \path (.5cm,-2.7cm) -- (.5cm,-3.2cm) node [black, font=\Large, midway, sloped] {$\dots$};

    \node [text=LightRed] at([xshift=-0.8cm,yshift=-3.9cm]toporig.south) {\large$s_{L-2}^{t}$};
    \tikzcuboid{%
        shiftx=0cm,%
        shifty=-4cm,%
        scale=.30,%
        rotation=0,%
        densityx=1,%
        densityy=1,%
        densityz=1,%
        dimx=4,%
        dimy=1,%
        dimz=1,%
        linefront=red!75!black,%
        linetop=red!50!black,%
        lineright=red!25!black,%
        fillfront=red!25!white,%
        filltop=red!50!white,%
        fillright=red!75!white,%
        emphedge=Y,%
        emphstyle=very thin,
    }
    \coordinate (o4) at ( 0.0, -4.0);

    \node [text=LightRed] at([xshift=-0.8cm,yshift=-4.9cm]toporig.south) {\large$s_{L-1}^{t}$};
    \tikzcuboid{%
        shiftx=0cm,%
        shifty=-5cm,%
        scale=.30,%
        rotation=0,%
        densityx=1,%
        densityy=1,%
        densityz=1,%
        dimx=4,%
        dimy=1,%
        dimz=1,%
        linefront=red!75!black,%
        linetop=red!50!black,%
        lineright=red!25!black,%
        fillfront=red!25!white,%
        filltop=red!50!white,%
        fillright=red!75!white,%
        emphedge=Y,%
        emphstyle=very thin,
    }
    \coordinate (o5) at ( 0.0, -5.0);

    \node [text=LightRed] at([xshift=-0.8cm,yshift=-5.9cm]toporig.south) {\large$s_L^{t}$};
    \tikzcuboid{%
        shiftx=0cm,%
        shifty=-6cm,%
        scale=.3,%
        rotation=0,%
        densityx=1,%
        densityy=1,%
        densityz=1,%
        dimx=4,%
        dimy=1,%
        dimz=1,%
        linefront=red!75!black,%
        linetop=red!50!black,%
        lineright=red!25!black,%
        fillfront=red!25!white,%
        filltop=red!50!white,%
        fillright=red!75!white,%
        emphedge=Y,%
        emphstyle=very thin,
    }
    \coordinate (o6) at ( 0.0, -6.0);

    \coordinate [right of=toporig, xshift=0cm,yshift=-2.9cm] (midorig);
    
    \node (semb)[rectangle, draw=black, right of=midorig, xshift=1cm, minimum width=2.6cm, minimum height=7cm,fill=embcolor, rounded corners =.2cm] {};
    \node[] at ([yshift=.3cm]semb.north) {\large Embedding};

    \tikzcuboid {%
        shiftx=2.2cm,%
        shifty=0cm,%
        scale=.22,%
        rotation=0,%
        densityx=1,%
        densityy=1,%
        densityz=1,%
        dimx=8,%
        dimy=1,%
        dimz=1,%
        linefront=green!75!black,%
        linetop=green!50!black,%
        lineright=green!25!black,%
        fillfront=green!25!white,%
        filltop=green!50!white,%
        fillright=green!75!white,%
        emphedge=Y,%
        emphstyle=very thin,
    }
    \draw [->,black, line width=.35mm] (1.3cm,0.05cm) -- (2cm,.05cm);
    
    \tikzcuboid{%
        shiftx=2.2cm,%
        shifty=-1cm,%
        scale=.22,%
        rotation=0,%
        densityx=1,%
        densityy=1,%
        densityz=1,%
        dimx=8,%
        dimy=1,%
        dimz=1,%
        linefront=green!75!black,%
        linetop=green!50!black,%
        lineright=green!25!black,%
        fillfront=green!25!white,%
        filltop=green!50!white,%
        fillright=green!75!white,%
        emphedge=Y,%
        emphstyle=very thin,
    }
    \draw [->,black, line width=.35mm] (1.3cm,-0.95cm) -- (2cm,-.95cm);
    
    \tikzcuboid{%
        shiftx=2.2cm,%
        shifty=-2cm,%
        scale=.22,%
        rotation=0,%
        densityx=1,%
        densityy=1,%
        densityz=1,%
        dimx=8,%
        dimy=1,%
        dimz=1,%
        linefront=green!75!black,%
        linetop=green!50!black,%
        lineright=green!25!black,%
        fillfront=green!25!white,%
        filltop=green!50!white,%
        fillright=green!75!white,%
        emphedge=Y,%
        emphstyle=very thin,
    }
    \draw [->,black, line width=.35mm] (1.3cm,-1.95cm) -- (2cm,-1.95cm);

    \path (3.0cm,-2.7cm) -- (3.0cm,-3.2cm) node [black, font=\Large, midway, sloped] {$\dots$};

    \tikzcuboid{%
        shiftx=2.2cm,%
        shifty=-4cm,%
        scale=.22,%
        rotation=0,%
        densityx=1,%
        densityy=1,%
        densityz=1,%
        dimx=8,%
        dimy=1,%
        dimz=1,%
        linefront=red!75!black,%
        linetop=red!50!black,%
        lineright=red!25!black,%
        fillfront=red!25!white,%
        filltop=red!50!white,%
        fillright=red!75!white,%
        emphedge=Y,%
        emphstyle=very thin,
    }
    \draw [->,black, line width=.35mm] (1.3cm,-3.95cm) -- (2cm,-3.95cm);
    
    \tikzcuboid{%
        shiftx=2.2cm,%
        shifty=-5cm,%
        scale=.22,%
        rotation=0,%
        densityx=1,%
        densityy=1,%
        densityz=1,%
        dimx=8,%
        dimy=1,%
        dimz=1,%
        linefront=red!75!black,%
        linetop=red!50!black,%
        lineright=red!25!black,%
        fillfront=red!25!white,%
        filltop=red!50!white,%
        fillright=red!75!white,%
        emphedge=Y,%
        emphstyle=very thin,
    }
    \draw [->,black, line width=.35mm] (1.3cm,-4.95cm) -- (2cm,-4.95cm);
    
    \tikzcuboid{%
        shiftx=2.2cm,%
        shifty=-6cm,%
        scale=.22,%
        rotation=0,%
        densityx=1,%
        densityy=1,%
        densityz=1,%
        dimx=8,%
        dimy=1,%
        dimz=1,%
        linefront=red!75!black,%
        linetop=red!50!black,%
        lineright=red!25!black,%
        fillfront=red!25!white,%
        filltop=red!50!white,%
        fillright=red!75!white,%
        emphedge=Y,%
        emphstyle=very thin,
    }
    \draw [->,black, line width=.35mm] (1.3cm,-5.95cm) -- (2cm,-5.95cm);

    \node (satt)[rectangle, draw=black, right of=midorig, xshift=6.1cm, minimum width=3.9cm, minimum height=7cm, fill=inpattcolor, rounded corners =.2cm] {};
    \node (satttext)[above of=satt, yshift=3cm, text width=2cm, align=center] {\large State\\ Attention};
    \node (sumatt) [circle,draw=black,right of=midorig, xshift=3.5cm, yshift=0.0cm, minimum width=.2cm,inner sep=0.01cm] {\scriptsize\texttt{mean}};
    
    \draw [->,black, line width=.35mm] (4.cm,-0.95cm) -- (5.15cm,-2.75cm) ;
    \node[text=LightGreen] at (5.2cm,-0.55cm) {\large$g(s^t_2)$};
    
    \draw [->,black, line width=.35mm] (4.cm,-1.95cm) -- (5.1cm,-2.85cm);
    \node[text=LightGreen] at (5.2cm,-1.55cm) {\large$g(s^t_3)$};
    
    \draw [->,black, line width=.35mm] (4.cm,-3.95cm) -- (5.1cm,-2.95cm);
    \node[text=LightRed] at (5.2cm,-4.35cm) {\large$g(s^t_{L-2})$};
    
    \draw [->,black, line width=.35mm] (4.cm,-4.95cm) -- (5.15cm,-3.05cm);
    \node[text=LightRed] at (5.2cm,-5.35cm) {\large$g(s^t_{L-1})$};

    \tikzcuboid{%
        shiftx=6.8cm,%
        shifty=-3cm,%
        scale=.22,%
        rotation=0,%
        densityx=1,%
        densityy=1,%
        densityz=1,%
        dimx=8,%
        dimy=1,%
        dimz=1,%
        linefront=blue!75!black,%
        linetop=blue!50!black,%
        lineright=blue!25!black,%
        fillfront=blue!25!white,%
        filltop=blue!50!white,%
        fillright=blue!75!white,%
        emphedge=Y,%
        emphstyle=very thin,
    }
    
    \draw [->,black, line width=.35mm] (5.95cm,-2.9cm) -- (6.65cm,-2.9cm);

    \draw [-,black, line width=.15mm, dotted] (4.3cm,.05cm) -- (6.15cm,.05cm);
    \draw [->,black, line width=.35mm] (4.cm,-0.95cm) -- (7.4cm,-0.95cm);
    \draw [->,black, line width=.35mm] (4.cm,-1.95cm) -- (6.8cm,-1.95cm);
    \draw [->,black, line width=.35mm] (4.cm,-3.95cm) -- (6.8cm,-3.95cm);
    \draw [->,black, line width=.35mm] (4.cm,-4.95cm) -- (7.4cm,-4.95cm);
    \draw [-,black, line width=.15mm, dotted] (4.3cm,-5.95cm) -- (6.15cm,-5.95cm);
    

    \draw [-,black, line width=.35mm] (5.6cm,-0.95cm) -- (6.15cm,-0.65cm);
    \draw [-,black, line width=.35mm] (5.6cm,-1.95cm) -- (6.15cm,-1.65cm);
    \draw [-,black, line width=.35mm] (5.6cm,-3.95cm) -- (6.15cm,-4.25cm);
    \draw [-,black, line width=.35mm] (5.6cm,-4.95cm) -- (6.15cm,-5.25cm);
    \draw [-,gray, line width=.35mm, dashed] (5.6cm,-0.95cm) -- (6.15cm,-0.65cm);
    \draw [-,gray, line width=.35mm, dashed] (5.6cm,-1.95cm) -- (6.15cm,-1.65cm);
    \draw [-,gray, line width=.35mm, dashed] (5.6cm,-3.95cm) -- (6.15cm,-4.25cm);
    \draw [-,gray, line width=.35mm, dashed] (5.6cm,-4.95cm) -- (6.15cm,-5.25cm);
    
    \node[circle,draw=black, minimum width=.1cm, fill=black, inner sep =.03cm] at (5.6cm,-0.95cm) {};
    \node[circle,draw=black, minimum width=.01cm, fill=black, inner sep =.03cm] at (5.6cm,-1.95cm) {};
    \node[circle,draw=black, minimum width=.01cm, fill=black, inner sep =.03cm] at (5.6cm,-3.95cm) {};
    \node[circle,draw=black, minimum width=.01cm, fill=black, inner sep =.03cm] at (5.6cm,-4.95cm) {};

    \node (sumatt) [circle,draw=black,right of=midorig, xshift=5.65cm, yshift=1.95cm, minimum width=.2cm,inner sep=0.1] {$+$};
    \node (sumatt) [circle,draw=black,right of=midorig, xshift=5.05cm, yshift=0.95cm, minimum width=.2cm,inner sep=0.1] {$+$};
    \node (sumatt) [circle,draw=black,right of=midorig, xshift=5.05cm, yshift=-1.05cm, minimum width=.2cm,inner sep=0.1] {$+$};
    \node (sumatt) [circle,draw=black,right of=midorig, xshift=5.65cm, yshift=-2.05cm, minimum width=.2cm,inner sep=0.1] {$+$};
    
    \draw [->,black, line width=.35mm] (7.05cm,-2.7cm) -- (7.05cm,-2.2cm);
    \draw [->,black, line width=.35mm] (7.3cm,-1.95cm) -- (9.cm,-1.95cm);

    \draw [->,black, line width=.35mm] (7.9cm,-0.95cm) -- (9.cm,-0.95cm);
    \draw [-,black, line width=.35mm] (7.65cm,-2.7cm) -- (7.65cm,-2.05cm);
    \draw[thick,black] (7.65cm,-2.05cm) arc (-90:90:.1cm);
    \draw [->,black, line width=.35mm] (7.65cm,-1.85cm) -- (7.65cm,-1.2cm);

    \draw [->,black, line width=.35mm] (7.05cm,-3.2cm) -- (7.05cm,-3.7cm);
    \draw [->,black, line width=.35mm] (7.3cm,-3.95cm) -- (9.cm,-3.95cm);

    \draw [->,black, line width=.35mm] (7.9cm,-4.95cm) -- (9.cm,-4.95cm);
    \draw [-,black, line width=.35mm] (7.65cm,-3.2cm) -- (7.65cm,-3.85cm);
    \draw[thick,black] (7.65cm,-4.05cm) arc (-90:90:.1cm);
    \draw [->,black, line width=.35mm] (7.65cm,-4.05cm) -- (7.65cm,-4.7cm);
    
    \node [ minimum width=.2cm,draw,rotate=-90,fill=white,minimum width=5cm, minimum height=.6cm] at ([xshift=8.4cm]midorig) {\texttt{softmax}};

    \node [ minimum width=.2cm,rotate=-90] at ([xshift=8.42cm,yshift=2.25cm]midorig) {\large$w^t$};
    \tikzcuboid{%
        shiftx=9.35cm,%
        shifty=-1.84cm,%
        scale=.22,%
        rotation=0,%
        densityx=1,%
        densityy=1,%
        densityz=1,%
        dimx=1,%
        dimy=2,%
        dimz=1,%
        linefront=red!75!black,%
        linetop=red!50!black,%
        lineright=red!25!black,%
        fillfront=red!25!white,%
        filltop=red!50!white,%
        fillright=red!75!white,%
        emphedge=Y,%
        emphstyle=very thin,
    }
        \tikzcuboid{%
        shiftx=9.35cm,%
        shifty=-1.4cm,%
        scale=.22,%
        rotation=0,%
        densityx=1,%
        densityy=1,%
        densityz=1,%
        dimx=1,%
        dimy=2,%
        dimz=1,%
        linefront=green!75!black,%
        linetop=green!50!black,%
        lineright=green!25!black,%
        fillfront=green!25!white,%
        filltop=green!50!white,%
        fillright=green!75!white,%
        emphedge=Y,%
        emphstyle=very thin,
    }
    
    \draw [->,black, line width=.35mm] (10.05cm,-0.65cm) -- (10.6cm,-0.95cm) -- (11.2cm,-0.95cm) -- (11.2cm,-2.6cm);
    \draw [->,black, line width=.35mm] (10.05cm,-1.65cm) -- (10.6cm,-1.95cm) -- (10.9cm,-1.95cm) -- (10.9cm,-2.6cm);
    \draw [->,black, line width=.35mm] (10.05cm,-4.25cm) -- (10.6cm,-3.95cm) -- (10.9cm,-3.95cm)-- (10.9cm,-3.2cm);
    \draw [->,black, line width=.35mm] (10.05cm,-5.25cm) -- (10.6cm,-4.95cm) -- (11.2cm,-4.95cm) -- (11.2cm,-3.2cm);
    \draw [-,gray, line width=.35mm, dashed] (10.05cm,-0.65cm) -- (10.6cm,-0.95cm);
    \draw [-,gray, line width=.35mm, dashed] (10.05cm,-1.65cm) -- (10.6cm,-1.95cm);
    \draw [-,gray, line width=.35mm, dashed] (10.05cm,-4.25cm) -- (10.6cm,-3.95cm);
    \draw [-,gray, line width=.35mm, dashed] (10.05cm,-5.25cm) -- (10.6cm,-4.95cm);
    
    \draw [->,black, line width=.35mm] (9.77cm,-2.9cm) -- (10.35cm,-2.9cm);
    
    \node [rectangle,draw=black,right of=midorig, xshift=9.2cm, yshift=0.0cm, minimum width=.5cm,inner sep=0.1cm,rounded corners=.25cm] {\tiny$\sum_l w_l^t g(s_l^t)$};

    \draw [->,black, line width=.35mm] (12.05cm,-2.9cm) -- (12.35cm,-2.9cm);

    \node [rectangle, right of=midorig, xshift=10.8cm, yshift=2.9cm, minimum width=.5cm,inner sep=0.1cm,rounded corners=.25cm] {\large$z^t_1$};

    \path (13.2cm,-1.cm) -- (13.2cm,-1.5cm) node [black, font=\Large, midway, sloped] {$\dots$};

    \node [rectangle, right of=midorig, xshift=10.8cm, yshift=0.4cm, minimum width=.5cm,inner sep=0.1cm,rounded corners=.25cm] {\large$z^t_p$};
    \path (13.2cm,-3.5cm) -- (13.2cm,-4.0cm) node [black, font=\Large, midway, sloped] {$\dots$};
    \node [rectangle, right of=midorig, xshift=10.8cm, yshift=-2.1cm, minimum width=.5cm,inner sep=0.1cm,rounded corners=.25cm] {\large$z^t_P$};

    \tikzcuboid{%
        shiftx=12.5cm,%
        shifty=-3cm,%
        scale=.22,%
        rotation=0,%
        densityx=1,%
        densityy=1,%
        densityz=1,%
        dimx=8,%
        dimy=1,%
        dimz=1,%
        linefront=blue!75!black,%
        linetop=blue!50!black,%
        lineright=blue!25!black,%
        fillfront=blue!30!red!30!white,%
        filltop=blue!40!red!30!white,%
        fillright=blue!50!red!30!white,%
        emphedge=Y,%
        emphstyle=very thin,
    }
    \tikzcuboid{%
        shiftx=12.5cm,%
        shifty=-.5cm,%
        scale=.22,%
        rotation=0,%
        densityx=1,%
        densityy=1,%
        densityz=1,%
        dimx=8,%
        dimy=1,%
        dimz=1,%
        linefront=blue!75!black,%
        linetop=blue!50!black,%
        lineright=blue!25!black,%
        fillfront=blue!30!red!30!white,%
        filltop=blue!40!red!30!white,%
        fillright=blue!50!red!30!white,%
        emphedge=Y,%
        emphstyle=very thin,
    }
    
        \tikzcuboid{%
        shiftx=12.5cm,%
        shifty=-5.5cm,%
        scale=.22,%
        rotation=0,%
        densityx=1,%
        densityy=1,%
        densityz=1,%
        dimx=8,%
        dimy=1,%
        dimz=1,%
        linefront=blue!75!black,%
        linetop=blue!50!black,%
        lineright=blue!25!black,%
        fillfront=blue!30!red!30!white,%
        filltop=blue!40!red!30!white,%
        fillright=blue!50!red!30!white,%
        emphedge=Y,%
        emphstyle=very thin,
    }
    
    \draw [-,black, line width=.2mm, dotted] (10.1cm,-.5cm) -- (12.35cm,-.5cm);
    \draw [-,black, line width=.2mm, dotted] (10.1cm,-5.5cm) -- (12.35cm,-5.5cm);
    
    \node (light1) [rectangle, draw=black, minimum width=1cm, minimum height=.4cm,inner sep=0.0cm,rounded corners=.1cm, fill=gray!30!white] at (13.7cm,-5.0cm){};
    \node (light11) [circle, draw=black, minimum width=.2cm,inner sep=0.0cm,rounded corners=.1cm, fill=red!80!black] at ([xshift=-0.8cm]light1.east){};
    \node (light12) [circle, draw=black, minimum width=.2cm,inner sep=0.0cm,rounded corners=.1cm, fill=orange!50!white, opacity=.06] at ([xshift=-0.5cm]light1.east){};
    \node (light13) [circle, draw=black, minimum width=.2cm,inner sep=0.0cm,rounded corners=.1cm, fill=green!50!black,opacity=.06] at ([xshift=-0.2cm]light1.east){};

    \node (light2) [rectangle, draw=black, minimum width=1cm, minimum height=.4cm,inner sep=0.0cm,rounded corners=.1cm, fill=gray!30!white] at (13.7cm,-.0cm){};
    \node (light21) [circle, draw=black, minimum width=.2cm,inner sep=0.0cm,rounded corners=.1cm, fill=red!80!black] at ([xshift=-0.8cm]light2.east){};
    \node (light22) [circle, draw=black, minimum width=.2cm,inner sep=0.0cm,rounded corners=.1cm, fill=orange!50!white, opacity=.06] at ([xshift=-0.5cm]light2.east){};
    \node (light23) [circle, draw=black, minimum width=.2cm,inner sep=0.0cm,rounded corners=.1cm, fill=green!50!black,opacity=.06] at ([xshift=-0.2cm]light2.east){};

    \node (light4) [rectangle, draw=black, minimum width=1cm, minimum height=.4cm,inner sep=0.0cm,rounded corners=.1cm, fill=gray!30!white] at (13.7cm,-2.5cm){};
    \node (light41) [circle, draw=black, minimum width=.2cm,inner sep=0.0cm,rounded corners=.1cm, fill=red!80!black, opacity=.06] at ([xshift=-0.8cm]light4.east){};
    \node (light42) [circle, draw=black, minimum width=.2cm,inner sep=0.0cm,rounded corners=.1cm, fill=orange!50!white, opacity=.06] at ([xshift=-0.5cm]light4.east){};
    \node (light43) [circle, draw=black, minimum width=.2cm,inner sep=0.0cm,rounded corners=.1cm, fill=green!50!black] at ([xshift=-0.2cm]light4.east){};    
    

    \node (rnn) [rect1, right of=midorig, fill=rnn2color, xshift=14.45cm, yshift=.4cm, minimum height=2.5cm, minimum width=2.5cm, rounded corners=.2cm] {};
    \node (rnntext)[above of=rnn, yshift=.5cm, text width=1.5cm, align=center] {\large RNN};

    \node [rectangle, right of=rnn, xshift=-.8cm, yshift=-.05cm, minimum height=1.cm] {\large$h^t$};
    \node [rectangle, right of=rnn, xshift=.5cm, yshift=-.05cm, minimum height=1.cm] {\large$o^t$};
    \begin{scope}[opacity=.4]
    \node [rectangle, right of=rnn, xshift=-1.0cm, yshift=.9cm, minimum height=1.cm] {\large$h^{t-1}$};
    \tikzcuboid{%
        shiftx=15.cm,%
        shifty=-2.06cm,%
        scale=.2,%
        rotation=0,%
        densityx=1,%
        densityy=1,%
        densityz=1,%
        dimx=8,%
        dimy=1,%
        dimz=1,%
        linefront=rnn1color!75!black,%
        linetop=rnn1color!50!black,%
        lineright=rnn1color!25!black,%
        fillfront=rnn1color!35!red,%
        filltop=rnn1color!50!red,%
        fillright=rnn1color!60!red,%
        emphedge=Y,%
        emphstyle=very thin
    }
    \draw [->,black, line width=.35mm] (15.0cm,-1.4cm) -- (15.2cm,-1.8cm);
    \draw [->,black, line width=.35mm] (15.35cm,-2.25cm) -- (15.55cm,-2.65cm);
    \end{scope}
    \draw [->,black, line width=.35mm] (14.35cm,-3cm) -- (15.0cm,-3cm);
    \tikzcuboid{%
        shiftx=15.2cm,%
        shifty=-3.cm,%
        scale=.22,%
        rotation=0,%
        densityx=1,%
        densityy=1,%
        densityz=1,%
        dimx=8,%
        dimy=1,%
        dimz=1,%
        linefront=rnn1color!75!black,%
        linetop=rnn1color!50!black,%
        lineright=rnn1color!25!black,%
        fillfront=rnn1color!35!red,%
        filltop=rnn1color!50!red,%
        fillright=rnn1color!60!red,%
        emphedge=Y,%
        emphstyle=very thin,
    }
    
    \node (phatt)[rectangle, draw=black, right of=midorig, xshift=17.4cm, minimum width=3.2cm, minimum height=7cm, fill=phattcolor, rounded corners =.2cm] {};
    \node (phatttext)[above of=phatt, yshift=3cm, text width=3cm, align=center] {\large Action\\ Attention};

    \node[circle,draw=black, minimum width=.01cm, fill=black, inner sep =.03cm] at (14.4cm,-3.0cm) {};
    \node[circle,draw=black, minimum width=.01cm, fill=black, inner sep =.03cm] at (19.08cm,-3.0cm) {};
    \node[circle,draw=black, minimum width=.01cm, fill=black, inner sep =.03cm] at (18.5cm,-3.0cm) {};

    \draw [->,black, line width=.35mm] (14.4cm,-3.0cm) -- (14.65cm,-4.15cm) -- (18.25cm,-4.15cm);
    \node [circle,draw=black,right of=midorig, xshift=16.5cm, yshift=-1.23cm, minimum width=.2cm,inner sep=0.1] {$+$};
    \draw [->,black, line width=.35mm] (17.02cm,-3cm) -- (18.5cm,-3cm) -- (18.5cm,-3.9cm) ;
    
    \draw [->,black, line width=.35mm] (14.35cm,-5.5cm) -- (18.8cm,-5.5cm);
    \node [circle,draw=black,right of=midorig, xshift=17.08cm, yshift=-2.6cm, minimum width=.2cm,inner sep=0.1] {$+$};
    \draw [->,black, line width=.35mm] (17.02cm,-3cm) -- (19.08cm,-3cm) -- (19.08cm,-5.23cm) ;

    \draw [->,black, line width=.35mm] (14.35cm,-.5cm) -- (18.8cm,-.5cm);
    \node [circle,draw=black,right of=midorig, xshift=17.08cm, yshift=2.38cm, minimum width=.2cm,inner sep=0.1] {$+$};
    \draw [->,black, line width=.35mm] (17.02cm,-3cm) -- (19.08cm,-3cm) -- (19.08cm,-.8cm) ;
    
    \draw [->,black, line width=.35mm]  (19.35cm,-.5cm) -- (20.1cm,-.5cm);
    \draw [->,black, line width=.35mm]  (19.35cm,-5.5cm) -- (20.1cm,-5.5cm);
    
    \draw [-,black, line width=.35mm]  (18.75cm,-4.12cm) -- (18.98cm,-4.12cm);
    \draw[thick,black] (19.18cm,-4.12cm) arc (0:180:.1cm);
    \draw [->,black, line width=.35mm]  (19.18cm,-4.12cm) -- (19.4cm,-4.12cm) -- (19.4cm, -3cm) -- (20.1cm,-3cm);
    
    \node [ minimum width=.2cm,draw,rotate=-90,fill=white,minimum width=5.5cm, minimum height=.5cm] at ([xshift=19.4cm,yshift=-.15cm]midorig) {\texttt{softmax}};
    
    \draw [->,black, line width=.35mm]  (20.7cm,-.5cm) -- (21.8cm,-.5cm);
    \draw [->,black, line width=.35mm]  (20.7cm,-3cm) -- (21.8cm,-3cm);
    \draw [->,black, line width=.35mm]  (20.7cm,-5.5cm) -- (21.8cm,-5.5cm);
    
    \tikzcuboid{%
        shiftx=22.2cm,%
        shifty=-3.cm,%
        scale=.3,%
        rotation=0,%
        densityx=1,%
        densityy=1,%
        densityz=1,%
        dimx=1,%
        dimy=1,%
        dimz=1,%
        linefront=cyan!75!black,%
        linetop=cyan!50!black,%
        lineright=cyan!25!black,%
        fillfront=cyan!35!white,%
        filltop=cyan!50!white,%
        fillright=cyan!60!white,%
        emphedge=Y,%
        emphstyle=very thin,
    }
    \tikzcuboid{%
        shiftx=22.2cm,%
        shifty=-.5cm,%
        scale=.3,%
        rotation=0,%
        densityx=1,%
        densityy=1,%
        densityz=1,%
        dimx=1,%
        dimy=1,%
        dimz=1,%
        linefront=cyan!75!black,%
        linetop=cyan!50!black,%
        lineright=cyan!25!black,%
        fillfront=cyan!35!white,%
        filltop=cyan!50!white,%
        fillright=cyan!60!white,%
        emphedge=Y,%
        emphstyle=very thin,
    }
    \tikzcuboid{%
        shiftx=22.2cm,%
        shifty=-5.5cm,%
        scale=.3,%
        rotation=0,%
        densityx=1,%
        densityy=1,%
        densityz=1,%
        dimx=1,%
        dimy=1,%
        dimz=1,%
        linefront=cyan!75!black,%
        linetop=cyan!50!black,%
        lineright=cyan!25!black,%
        fillfront=cyan!35!white,%
        filltop=cyan!50!white,%
        fillright=cyan!60!white,%
        emphedge=Y,%
        emphstyle=very thin,
    }
    
    \node [ minimum width=.2cm] at (22.3cm, 0.1) {\large$\pi^t_{1}$};
    \node [ minimum width=.2cm] at (22.3cm, -2.4) {\large$\pi^t_{p}$};
    \node [ minimum width=.2cm] at (22.3cm, -4.9) {\large$\pi^t_{P}$};
    \path (22.3cm,-1.4cm) -- (22.3cm,-1.9cm) node [black, font=\Large, midway, sloped] {$\dots$};
    \path (22.3cm,-3.8cm) -- (22.3cm,-4.3cm) node [black, font=\Large, midway, sloped] {$\dots$};

    \node (light4) [rectangle, draw=black, minimum width=1cm, minimum height=.4cm,inner sep=0.0cm,rounded corners=.1cm, fill=gray!30!white] at (22.3cm,-3.5cm){};
    \node (light41) [circle, draw=black, minimum width=.2cm,inner sep=0.0cm,rounded corners=.1cm, fill=red!80!black] at ([xshift=-0.8cm]light4.east){};
    \node (light42) [circle, draw=black, minimum width=.2cm,inner sep=0.0cm,rounded corners=.1cm, fill=orange!50!white, opacity=.06] at ([xshift=-0.5cm]light4.east){};
    \node (light43) [circle, draw=black, minimum width=.2cm,inner sep=0.0cm,rounded corners=.1cm, fill=green!50!black,opacity=.06] at ([xshift=-0.2cm]light4.east){};
    
    \node (light5) [rectangle, draw=black, minimum width=1cm, minimum height=.4cm,inner sep=0.0cm,rounded corners=.1cm, fill=gray!30!white] at (22.3cm,-6cm){};
    \node (light51) [circle, draw=black, minimum width=.2cm,inner sep=0.0cm,rounded corners=.1cm, fill=red!80!black] at ([xshift=-0.8cm]light5.east){};
    \node (light52) [circle, draw=black, minimum width=.2cm,inner sep=0.0cm,rounded corners=.1cm, fill=orange!50!white, opacity=.06] at ([xshift=-0.5cm]light5.east){};
    \node (light53) [circle, draw=black, minimum width=.2cm,inner sep=0.0cm,rounded corners=.1cm, fill=green!50!black,opacity=.06] at ([xshift=-0.2cm]light5.east){};
    
    \node (light6) [rectangle, draw=black, minimum width=1cm, minimum height=.4cm,inner sep=0.0cm,rounded corners=.1cm, fill=gray!30!white] at (22.3cm,-1cm){};
    \node (light61) [circle, draw=black, minimum width=.2cm,inner sep=0.0cm,rounded corners=.1cm, fill=red!80!black, opacity=.06] at ([xshift=-0.8cm]light6.east){};
    \node (light62) [circle, draw=black, minimum width=.2cm,inner sep=0.0cm,rounded corners=.1cm, fill=orange!50!white, opacity=.06] at ([xshift=-0.5cm]light6.east){};
    \node (light63) [circle, draw=black, minimum width=.2cm,inner sep=0.0cm,rounded corners=.1cm, fill=green!50!black] at ([xshift=-0.2cm]light6.east){};

\end{tikzpicture}


%% file: intersection1_noinout.tex
\definecolor{LightGreen}{HTML}{CEFBCE}
\definecolor{LightRed}{HTML}{FFDDDD}

\input{tikz_utils}

\begin{tikzpicture}[]
\node [bg, minimum height =7cm, minimum width=12cm] at (.0, 1.5) {} ;
\draw[road2 node] (-5cm,.7cm)  -- (5cm,.7cm);
\draw[road2 node] (-5cm,-.7cm)  -- (5cm,-.7cm);
\def\shiftr{18.5pt}
\draw[road2 node] (-.7cm, .7cm+\shiftr ) -- (-.7,4cm +\shiftr);
\draw[road2 node] (.7cm, .7cm+\shiftr) -- (.7cm,4cm+ +\shiftr);
\node[fill=way, minimum height=2.7cm, minimum width=2.66cm] at (0,0.05) {};

\leftturn(-2.1,-0.5){0};
\straight(-2.1,-1){0};
\straight(2.1,0.4){180};
\rightturn(2.1,.9){180};
\leftturn(-0.5, 2.1){270};
\rightturn(-.9,2.1){270};


share
\end{tikzpicture}

%% file: intersection2.tex
\definecolor{LightGreen}{HTML}{CEFBCE}
\definecolor{LightRed}{HTML}{FFDDDD}

\input{tikz_utils}

\begin{tikzpicture}[]
\node [bg, minimum height =11cm, minimum width=12cm] at (.0, 0) {} ;
\draw[road2 node] (-5cm,.7cm)  -- (5cm,.7cm);
\draw[road2 node] (-5cm,-.7cm)  -- (5cm,-.7cm);
\def\shiftr{18.5pt}
\draw[road2 node] (-.7cm, -5.2cm+\shiftr ) -- (-.7,4cm +\shiftr);
\draw[road2 node] (.7cm, -5.2cm+\shiftr) -- (.7cm,4cm+ +\shiftr);
\node[fill=way, minimum height=2.7cm, minimum width=2.8cm] at (0,0) {};

\leftturn(-2.1,-0.5){0};
\straightrightturn(-2.1,-.9){0};

\leftturn(2.1,0.5){180};
\straightrightturn(2.1,.9){180};

\leftturn(-0.5, 2.){270};
\straightrightturn(-.9,2.){270};

\leftturn(0.5, -2.){90};
\straightrightturn(.9,-2.){90};



share
\end{tikzpicture}

%% file: intersection5.tex
\definecolor{LightGreen}{HTML}{CEFBCE}
\definecolor{LightRed}{HTML}{FFDDDD}

\input{tikz_utils}

\begin{tikzpicture}[]
\node [bg, minimum height =12cm, minimum width=12cm] at (.0, 0) {} ;
\draw[road2 node] (-5cm,.7cm)  -- (5cm,.7cm);
\draw[road2 node] (-5cm,-.7cm)  -- (0cm,-.7cm);
\draw[road2 node] (-0cm,-.7cm)  -- (5cm,-.7cm);
\def\shiftr{18.5pt}
\draw[road2 node] (-.0cm, .705cm+\shiftr ) -- (-.0,4cm +\shiftr);
\draw[road2 node] (-.0cm, -5.2cm+\shiftr ) -- (-.0,-1.305cm +\shiftr);
\draw[road2 node] (-0cm,-.7cm)  -- (5cm,-.7cm);

\node[fill=way, minimum height=2.75cm, minimum width=1.35cm] at (0,0) {};

\straight(-1.3,-.3){0};
\rightturn(-1.3,-.9){0};

\leftturn(1.3,0.5){180};
\straight(1.3,.9){180};

\leftturn(0.3, 2.){270};
\straight(-.2,2.){270};



share
\end{tikzpicture}

%% file: intersection6.tex
\definecolor{LightGreen}{HTML}{CEFBCE}
\definecolor{LightRed}{HTML}{FFDDDD}

\input{tikz_utils}

\begin{tikzpicture}[]
\node [bg, minimum height =12cm, minimum width=12cm] at (.0, 0) {} ;
\draw[road3 node] (-5cm,1.02cm)  -- (5cm,1.02cm);
\draw[road3 node] (-5cm,-1.02cm)  -- (5cm,-1.02cm);
\def\shiftr{18.5pt}
\draw[road2 node] (-.72cm, -5cm ) -- (-.72,5cm );
\draw[road2 node] (0.72cm, -5cm) -- (.72cm,5cm);
\node[fill=way, minimum height=4.cm, minimum width=2.8cm] at (0,0) {};

\leftturn(-2.1,-0.5){0};
\straight(-2.1,-1){0};
\rightturn(-2.1,-1.5){0};

\leftturn(2.1,0.5){180};
\straight(2.1,1){180};
\rightturn(2.1,1.5){180};

\leftturn(0.5,-2.7){90};
\straightrightturn(0.9,-2.7){90};

\leftturn(-0.5,2.7){270};
\straightrightturn(-.9,2.7){270};






share
\end{tikzpicture}

%% file: intersection1_nolabel_app.tex
\definecolor{LightGreen}{HTML}{CEFBCE}
\definecolor{LightRed}{HTML}{FFDDDD}

\input{tikz_utils}

\begin{tikzpicture}[]
\node [bg, minimum height =7cm, minimum width=12cm] at (.0, 1.5) {} ;
\draw[road2 node] (-5cm,.7cm)  -- (5cm,.7cm);
\draw[road2 node] (-5cm,-.7cm)  -- (5cm,-.7cm);
\def\shiftr{18.5pt}
\draw[road2 node] (-.7cm, .7cm+\shiftr ) -- (-.7,4cm +\shiftr);
\draw[road2 node] (.7cm, .7cm+\shiftr) -- (.7cm,4cm+ +\shiftr);
\node[fill=way, minimum height=2.7cm, minimum width=2.66cm] at (0,0.05) {};

\leftturn(-2.1,-0.5){0};
\straight(-2.1,-1){0};
\straight(2.1,0.4){180};
\rightturn(2.1,.9){180};
\leftturn(-0.5, 2.1){270};
\rightturn(-.9,2.1){270};

share
\end{tikzpicture}

%% file: intersection1_traffic_3roads.tex
\definecolor{LightGreen}{HTML}{04300B}
\definecolor{LightRed}{HTML}{580101}

\input{phase_utils}

\begin{tikzpicture}[]
\node[] (ph1_txt) at (-0,2.0) {\footnotesize Phase-1};
\node [wbg, minimum height =1.2cm, minimum width=1.2cm,rounded corners=0.2cm] (ph1_sign) at (.0, 1.2) {} ;
\leftturn(-.10, 1.50cm){270};

\node[]  (ph2_txt) at (-.0,.05) {\footnotesize Phase-2};
\node [wbg, minimum height =1.2cm, minimum width=1.2cm,rounded corners=0.2cm] (ph2_sign) at (.0, -0.8) {} ;
\straight(.3,-0.75){180};

\node[] (ph3_txt) at (-0,-1.95) {\footnotesize Phase-3};
\node (ph3_sign) [wbg, minimum height =1.2cm, minimum width=1.2cm,rounded corners=0.2cm] at (.0, -2.8) {} ;
\leftturn(-.320, -2.75cm){0};
\straight(-.320, -3.05cm){0};





share
\end{tikzpicture}

%% file: intersection4.tex
\definecolor{LightGreen}{HTML}{CEFBCE}
\definecolor{LightRed}{HTML}{FFDDDD}

\input{tikz_utils}

\begin{tikzpicture}[]
\node [bg, minimum height =8cm, minimum width=12cm] at (.0, 1.5) {} ;
\draw[road2 node] (-5cm,.7cm)  -- (5cm,.7cm);
\draw[road2 node] (-5cm,-.7cm)  -- (5cm,-.7cm);
\def\shiftr{18.5pt}
\draw[road2 node] (-.0cm, .7cm+\shiftr ) -- (-.0,4cm +\shiftr);
\node[fill=way, minimum height=2.7cm, minimum width=1.25cm] at (0,0.05) {};

\leftturn(-1.3,-0.5){0};
\straight(-1.3,-1){0};
\straight(1.3,0.4){180};
\straightrightturn(1.3,.9){180};



share
\end{tikzpicture}

%% file: intersection4_phases.tex
\definecolor{LightGreen}{HTML}{258500}
\definecolor{LightRed}{HTML}{BA0000}

\input{phase_utils}

\begin{tikzpicture}[]
\node[] at (0,0.9) {\small Phase-1};
\node [wbg, minimum height =1.2cm, minimum width=1.2cm,rounded corners=0.2cm] at (.0, .0) {} ;
\leftturn(-.3, .10cm){0};
\straight(-.3, -.2cm){0};
\node[] at (0,-1.1) {\small Phase-2};
\node [wbg, minimum height =1.2cm, minimum width=1.2cm,rounded corners=0.2cm] at (.0, -2) {} ;
\straight(.3,-1.7){180};
\straight(.3,-2.0){180};
\straight(-.3,-2.3){0};

share
\end{tikzpicture}

%% file: intersection2_phases_4.tex
\definecolor{LightGreen}{HTML}{258500}
\definecolor{LightRed}{HTML}{BA0000}

\input{phase_utils}

\begin{tikzpicture}[]
background
\node[] at (0.0,.9) {\small Phase-1};
\node [wbg, minimum height =1.2cm, minimum width=1.2cm,rounded corners=0.2cm] at (.0, .0) {} ;

\leftturn(.10, .30cm){270};
\leftturn(-.1, -.30cm){90};
\node[] at (0.0,-1.1) {\small Phase-2};
\node [wbg, minimum height =1.2cm, minimum width=1.2cm,rounded corners=0.2cm] at (.0, -2.0) {} ;
\leftturn(-.3,-1.9){0};
\leftturn(.3,-2.1){180};

\node[] at (0.0,-3.1) {\small Phase-3};
\node [wbg, minimum height =1.2cm, minimum width=1.2cm,rounded corners=0.2cm] at (.0, -4) {} ;
\straight(.2, -4.3cm){90};
\straight(-.2, -3.7cm){270};

\node[] at (0.0,-5.1) {\small Phase-4};
\node [wbg, minimum height =1.2cm, minimum width=1.2cm,rounded corners=0.2cm] at (.0, -6) {} ;
\straight(-.3, -6.2cm){0};
\straight(.3, -5.8cm){180};

share
\end{tikzpicture}

%% file: intersection2_phases_8.tex
\definecolor{LightGreen}{HTML}{258500}
\definecolor{LightRed}{HTML}{BA0000}

\input{phase_utils}

\begin{tikzpicture}[]
background
\node[] at (0.0,.9) {\small Phase-1};
\node [wbg, minimum height =1.2cm, minimum width=1.2cm,rounded corners=0.2cm] at (.0, .0) {} ;

\leftturn(.10, .30cm){270};
\leftturn(-.1, -.30cm){90};
\node[] at (0.0,-1.1) {\small Phase-2};
\node [wbg, minimum height =1.2cm, minimum width=1.2cm,rounded corners=0.2cm] at (.0, -2.0) {} ;
\leftturn(-.3,-1.9){0};
\leftturn(.3,-2.1){180};

\node[] at (0.0,-3.1) {\small Phase-3};
\node [wbg, minimum height =1.2cm, minimum width=1.2cm,rounded corners=0.2cm] at (.0, -4) {} ;
\straight(.2, -4.3cm){90};
\straight(-.2, -3.7cm){270};

\node[] at (0.0,-5.1) {\small Phase-4};
\node [wbg, minimum height =1.2cm, minimum width=1.2cm,rounded corners=0.2cm] at (.0, -6) {} ;
\straight(-.3, -6.2cm){0};
\straight(.3, -5.8cm){180};

\node[] at (1.5,0.9) {\small Phase-5};
\node [wbg, minimum height =1.2cm, minimum width=1.2cm,rounded corners=0.2cm] at (1.5, .0) {} ;
\leftturn(1.55, .30cm){270};
\straight(1.25, .30cm){270}
\node[] at (1.5,-1.1) {\small Phase-6};
\node [wbg, minimum height =1.2cm, minimum width=1.2cm,rounded corners=0.2cm] at (1.5, -2.) {} ;
\leftturn(1.2,-1.95){0};
\straight(1.2, -2.25cm){0};

\node[] at (1.5,-3.1) {\small Phase-7};
\node [wbg, minimum height =1.2cm, minimum width=1.2cm,rounded corners=0.2cm] at (1.5, -4.) {} ;
\straight(1.75, -4.3cm){90};
\leftturn(1.45,-4.3){90};
\node[] at (1.5,-5.1) {\small Phase-8};
\node [wbg, minimum height =1.2cm, minimum width=1.2cm,rounded corners=0.2cm] at (1.5, -6) {} ;
\leftturn(1.8, -6.05cm){180};
\straight(1.8, -5.75cm){180};

share
\end{tikzpicture}

%% file: intersection3.tex
\definecolor{LightGreen}{HTML}{CEFBCE}
\definecolor{LightRed}{HTML}{FFDDDD}

\input{tikz_utils}

\begin{tikzpicture}[]
\node [bg, minimum height =12cm, minimum width=12cm] at (.0, 1.5) {} ;
\draw[road3 node] (-5cm,2.52cm)  -- (5cm,2.52cm);
\draw[road3 node] (-5cm,0.48cm)  -- (5cm,0.48cm);
\def\shiftr{18.5pt}
\draw[road3 node] (-1.02cm, -3.5cm ) -- (-1.02,6.5cm );
\draw[road3 node] (1.02cm, -3.5cm) -- (1.02cm,6.5cm);
\node[fill=way, minimum height=4.cm, minimum width=4.1cm] at (0,1.5) {};

\leftturn(-2.7,1.0){0};
\straight(-2.7,0.5){0};
\rightturn(-2.7,0.0){0};

\leftturn(2.7,2.0){180};
\straight(2.7,2.5){180};
\rightturn(2.7,3.0){180};

\leftturn(0.5,-1.2){90};
\straight(1.0,-1.2){90};
\rightturn(1.5,-1.2){90};

\leftturn(-0.5,4.2){270};
\straight(-1,4.2){270};
\rightturn(-1.5,4.2){270};

share
\end{tikzpicture}

%% file: intersection3_phases.tex
\definecolor{LightGreen}{HTML}{258500}
\definecolor{LightRed}{HTML}{BA0000}

\input{phase_utils}

\begin{tikzpicture}[]
background
\node[] at (0.0,0.9) {\small Phase-1};
\node [wbg, minimum height =1.2cm, minimum width=1.2cm,rounded corners=0.2cm] at (.0, .0) {} ;
\straight(-.28, .20cm){0};
\straight(.28, -.20cm){180};
\node[] at (0.0,-1.1) {\small Phase-2};
\node [wbg, minimum height =1.2cm, minimum width=1.2cm,rounded corners=0.2cm] at (.0, -2) {} ;
\straight(-.2,-2.3){90};
\straight(.2,-1.7){270};

\node[] at (0.0,-3.1) {\small Phase-3};
\node [wbg, minimum height =1.2cm, minimum width=1.2cm,rounded corners=0.2cm] at (.0, -4) {} ;
\leftturn(.3, -4.1cm){180};
\leftturn(-.3, -3.9cm){0};

\node[] at (0.0,-5.1) {\small Phase-4};
\node [wbg, minimum height =1.2cm, minimum width=1.2cm,rounded corners=0.2cm] at (.0, -6) {} ;
\leftturn(-.1, -6.3cm){90};
\leftturn(.1, -5.7cm){270};

\node[] at (1.5,0.9) {\small Phase-5};
\node [wbg, minimum height =1.2cm, minimum width=1.2cm,rounded corners=0.2cm] at (1.5, .0) {} ;
\leftturn(1.2, .1cm){0};
\straight(1.2, -.2cm){0}
\node[] at (1.5,-1.1) {\small Phase-6};
\node [wbg, minimum height =1.2cm, minimum width=1.2cm,rounded corners=0.2cm] at (1.5, -2) {} ;
\straight(1.8, -1.8cm){180};
\leftturn(1.8,-2.1){180};
\node[] at (1.5,-3.10) {\small Phase-7};
\node [wbg, minimum height =1.2cm, minimum width=1.2cm,rounded corners=0.2cm] at (1.5, -4) {} ;
\leftturn(1.6, -3.8cm){270};
\straight(1.3,-3.8){270};
\node[] at (1.5,-5.10) {\small Phase-8};
\node [wbg, minimum height =1.2cm, minimum width=1.2cm,rounded corners=0.2cm] at (1.5, -6) {} ;
\leftturn(1.4, -6.2cm){90};
\straight(1.7, -6.2cm){90};

share
\end{tikzpicture}

%% file: intersection5_phases.tex
\definecolor{LightGreen}{HTML}{258500}
\definecolor{LightRed}{HTML}{BA0000}

\input{phase_utils}

\begin{tikzpicture}[]
background
\node[] at (0.0,0.9) {\small Phase-1};
\node [wbg, minimum height =1.2cm, minimum width=1.2cm,rounded corners=0.2cm] at (.0, .0) {} ;
\straight(-.28, .2cm){0};
\straight(.28, -.2cm){180};
\node[] at (0.0,-1.1) {\small Phase-2};
\node [wbg, minimum height =1.2cm, minimum width=1.2cm,rounded corners=0.2cm] at (.0, -2) {} ;
\straight(-.2,-1.72){270};
\leftturn(.1,-1.72){270};

\node[] at (0.0,-3.10) {\small Phase-3};
\node [wbg, minimum height =1.2cm, minimum width=1.2cm,rounded corners=0.2cm] at (.0, -4) {} ;
\leftturn(.3, -4.1cm){180};
\straight(.3, -3.8cm){180};

share
\end{tikzpicture}

%% file: intersection1_phases.tex
\definecolor{LightGreen}{HTML}{04300B}
\definecolor{LightRed}{HTML}{580101}

\input{phase_utils}

\begin{tikzpicture}[]
\node[] (ph1_txt) at (-0,2.0) {\footnotesize Phase-1};
\node [wbg, minimum height =1.2cm, minimum width=1.2cm,rounded corners=0.2cm] (ph1_sign) at (.0, 1.2) {} ;
\leftturn(-.10, 1.50cm){270};

\node[]  (ph2_txt) at (-.0,.05) {\footnotesize Phase-2};
\node [wbg, minimum height =1.2cm, minimum width=1.2cm,rounded corners=0.2cm] (ph2_sign) at (.0, -0.8) {} ;
\straight(.3,-0.68){180};
\straight(-.3,-1.03){0};

\node[] (ph3_txt) at (-0,-1.95) {\footnotesize Phase-3};
\node (ph3_sign) [wbg, minimum height =1.2cm, minimum width=1.2cm,rounded corners=0.2cm] at (.0, -2.8) {} ;
\leftturn(-.320, -2.75cm){0};
\straight(-.320, -3.05cm){0};





share
\end{tikzpicture}

%% file: intersection9.tex
\definecolor{LightGreen}{HTML}{CEFBCE}
\definecolor{LightRed}{HTML}{FFDDDD}

\input{tikz_utils}

\begin{tikzpicture}[]
\node [bg, minimum height =9cm, minimum width=12cm] at (.0, 1.5) {} ;
\draw[road3 node] (-5cm,.38cm)  -- (0cm,.38cm);
\draw[road2 node] (0cm,.7cm)  -- (5cm,.7cm);
\draw[road2 node] (-5cm,-1.35cm)  -- (0cm,-1.35cm);
\draw[road3 node] (-0cm,-1.035cm)  -- (5cm,-1.035cm);

\def\shiftr{18.5pt}
\draw[road1 node] (-.4cm, .7cm+\shiftr ) -- (-.4,4cm +\shiftr);
\draw[road1 node] (.4cm, .7cm+\shiftr) -- (.4cm,4cm+ +\shiftr);
\node[fill=way, minimum height=3.34cm, minimum width=1.4cm] at (0,-0.292) {};

\leftturn(-1.6,-1.15){0};
\straight(-1.6,-1.65){0};
\straight(1.6,0.4){180};
\rightturn(1.6,0.9){180};
\lefttrightturn(-.4,2.1){270};

share
\end{tikzpicture}

%% file: intersection9_phases.tex
\definecolor{LightGreen}{HTML}{258500}
\definecolor{LightRed}{HTML}{BA0000}

\input{phase_utils}

\begin{tikzpicture}[]
background
\node[] at (0.0,.9) {\small Phase-1};
\node [wbg, minimum height =1.2cm, minimum width=1.2cm,rounded corners=0.2cm] at (.0, .0) {} ;

\straight(-.3, -.2cm){0};
\straight(.3, .2cm){180};

\node[] at (0.0,-1.1) {\small Phase-2};
\node [wbg, minimum height =1.2cm, minimum width=1.2cm,rounded corners=0.2cm] at (.0, -2.0) {} ;

\straight(-.30, -2.3cm){0};
\leftturn(-.3, -1.95cm){0};

\node[] at (0.0,-3.1) {\small Phase-3};
\node [wbg, minimum height =1.2cm, minimum width=1.2cm,rounded corners=0.2cm] at (.0, -4) {} ;
\leftturn(-.10, -3.70cm){270};

share
\end{tikzpicture}

%% file: intersection8.tex
\definecolor{LightGreen}{HTML}{CEFBCE}
\definecolor{LightRed}{HTML}{FFDDDD}

\input{tikz_utils}

\begin{tikzpicture}[]
\node [bg, minimum height =9cm, minimum width=12cm] at (.0, 1.5) {} ;
\draw[road3 node] (-5cm,.38cm)  -- (0cm,.38cm);
\draw[road2 node] (0cm,.7cm)  -- (5cm,.7cm);
\draw[road2 node] (-5cm,-1.35cm)  -- (0cm,-1.35cm);
\draw[road3 node] (-0cm,-1.035cm)  -- (5cm,-1.035cm);

\def\shiftr{18.5pt}
\draw[road2 node] (-.7cm, .7cm+\shiftr ) -- (-.7,4cm +\shiftr);
\draw[road2 node] (.7cm, .7cm+\shiftr) -- (.7cm,4cm+ +\shiftr);
\node[fill=way, minimum height=3.34cm, minimum width=2.66cm] at (0,-0.29) {};

\leftturn(-2.6,-1.15){0};
\straight(-2.6,-1.65){0};
\straight(2.1,0.4){180};
\rightturn(2.1,0.9){180};
\leftturn(-0.5, 2.1){270};
\rightturn(-0.9,2.1){270};

share
\end{tikzpicture}

%% file: intersection8_phases.tex
\definecolor{LightGreen}{HTML}{258500}
\definecolor{LightRed}{HTML}{BA0000}

\input{phase_utils}

\begin{tikzpicture}[]
background
\node[] at (0.0,.9) {\small Phase-1};
\node [wbg, minimum height =1.2cm, minimum width=1.2cm,rounded corners=0.2cm] at (.0, .0) {} ;

\straight(-.3, -.2cm){0};
\straight(.3, .2cm){180};

\node[] at (0.0,-1.1) {\small Phase-2};
\node [wbg, minimum height =1.2cm, minimum width=1.2cm,rounded corners=0.2cm] at (.0, -2.0) {} ;

\straight(-.30, -2.3cm){0};
\leftturn(-.3, -1.95cm){0};

\node[] at (0.0,-3.1) {\small Phase-3};
\node [wbg, minimum height =1.2cm, minimum width=1.2cm,rounded corners=0.2cm] at (.0, -4) {} ;
\leftturn(-.10, -3.70cm){270};

share
\end{tikzpicture}

%% file: intersection7.tex
\definecolor{LightGreen}{HTML}{CEFBCE}
\definecolor{LightRed}{HTML}{FFDDDD}

\input{tikz_utils}

\begin{tikzpicture}[]
\node [bg, minimum height =9cm, minimum width=12cm] at (.0, 1.5) {} ;
\draw[road3 node] (-5cm,.38cm)  -- (0cm,.38cm);
\draw[road2 node] (0cm,.7cm)  -- (5cm,.7cm);
\draw[road2 node] (-5cm,-1.35cm)  -- (0cm,-1.35cm);
\draw[road3 node] (-0cm,-1.035cm)  -- (5cm,-1.035cm);

\def\shiftr{18.5pt}
\draw[road3 node] (-0.95cm, .7cm+\shiftr ) -- (-0.95,4cm +\shiftr);
\draw[road2 node] (.8cm, .7cm+\shiftr) -- (.8cm,4cm+ +\shiftr);
\node[fill=way, minimum height=3.34cm, minimum width=3.33cm] at (-.23,-0.29) {};

\leftturn(-2.6,-1.15){0};
\straight(-2.6,-1.65){0};
\straight(2.1,0.4){180};
\rightturn(2.1,0.9){180};
\leftturn(-0.45, 2.1){270};
\leftturn(-1.05, 2.1){270};
\rightturn(-1.45,2.1){270};

share
\end{tikzpicture}

%% file: intersection7_phases.tex
\definecolor{LightGreen}{HTML}{258500}
\definecolor{LightRed}{HTML}{BA0000}

\input{phase_utils}

\begin{tikzpicture}[]
background
\node[] at (0.0,.9) {\small Phase-1};
\node [wbg, minimum height =1.2cm, minimum width=1.2cm,rounded corners=0.2cm] at (.0, .0) {} ;

\straight(-.3, -.2cm){0};
\straight(.3, .2cm){180};

\node[] at (0.0,-1.1) {\small Phase-2};
\node [wbg, minimum height =1.2cm, minimum width=1.2cm,rounded corners=0.2cm] at (.0, -2.0) {} ;

\straight(-.30, -2.3cm){0};
\leftturn(-.3, -1.95cm){0};

\node[] at (0.0,-3.1) {\small Phase-3};
\node [wbg, minimum height =1.2cm, minimum width=1.2cm,rounded corners=0.2cm] at (.0, -4) {} ;
\leftturn(.10, -3.70cm){270};
\leftturn(-.3, -3.70cm){270};

share
\end{tikzpicture}

%% file: intersection11.tex
\definecolor{LightGreen}{HTML}{CEFBCE}
\definecolor{LightRed}{HTML}{FFDDDD}

\input{tikz_utils}

\begin{tikzpicture}[]
\node [bg, minimum height =11cm, minimum width=12cm] at (.0, 0) {} ;
\draw[road3 node] (-5cm,1.02cm)  -- (5cm,1.02cm);
\draw[road3 node] (-5cm,-1.02cm)  -- (5cm,-1.02cm);
\def\shiftr{18.5pt}
\draw[road2 node] (-.72cm, -5cm ) -- (-.72,5cm );
\draw[road2 node] (0.72cm, -5cm) -- (.72cm,5cm);
\node[fill=way, minimum height=4.cm, minimum width=2.8cm] at (0,0) {};

\straight(-2.1,-0.4){0};
\straight(-2.1,-1){0};
\rightturn(-2.1,-1.5){0};

\straight(2.1,0.4){180};
\straight(2.1,1){180};
\rightturn(2.1,1.5){180};

\leftturn(0.5,-2.7){90};
\straightrightturn(0.9,-2.7){90};

\leftturn(-0.5,2.7){270};
\straightrightturn(-.9,2.7){270};






share
\end{tikzpicture}

%% file: intersection11_phases_3.tex
\definecolor{LightGreen}{HTML}{258500}
\definecolor{LightRed}{HTML}{BA0000}

\input{phase_utils}

\begin{tikzpicture}[]
background
\node[] at (0.0,.9) {\small Phase-1};
\node [wbg, minimum height =1.2cm, minimum width=1.2cm,rounded corners=0.2cm] at (.0, .0) {} ;

\straight(-.3, -.2cm){0};
\straight(.3, .2cm){180};

\node[] at (0.0,-1.1) {\small Phase-2};
\node [wbg, minimum height =1.2cm, minimum width=1.2cm,rounded corners=0.2cm] at (.0, -2.0) {} ;

\straight(.20, -1.7cm){270};
\straight(-.2, -2.30cm){90};

\node[] at (0.0,-3.1) {\small Phase-3};
\node [wbg, minimum height =1.2cm, minimum width=1.2cm,rounded corners=0.2cm] at (.0, -4) {} ;
\leftturn(.10, -3.70cm){270};
\leftturn(-.1, -4.30cm){90};

share
\end{tikzpicture}

%% file: intersection11_phases_5.tex
\definecolor{LightGreen}{HTML}{258500}
\definecolor{LightRed}{HTML}{BA0000}

\input{phase_utils}

\begin{tikzpicture}[]
background
\node[] at (0.0,.9) {\small Phase-1};
\node [wbg, minimum height =1.2cm, minimum width=1.2cm,rounded corners=0.2cm] at (.0, .0) {} ;

\straight(-.3, -.2cm){0};
\straight(.3, .2cm){180};

\node[] at (0.0,-1.1) {\small Phase-2};
\node [wbg, minimum height =1.2cm, minimum width=1.2cm,rounded corners=0.2cm] at (.0, -2.0) {} ;

\straight(.20, -1.7cm){270};
\straight(-.2, -2.30cm){90};

\node[] at (0.0,-3.1) {\small Phase-3};
\node [wbg, minimum height =1.2cm, minimum width=1.2cm,rounded corners=0.2cm] at (.0, -4) {} ;
\leftturn(.10, -3.70cm){270};
\leftturn(-.1, -4.30cm){90};

\node[] at (1.5,0.9) {\small Phase-4};
\node [wbg, minimum height =1.2cm, minimum width=1.2cm,rounded corners=0.2cm] at (1.5, .0) {} ;
\leftturn(1.4, -.30cm){90};
\straight(1.7, -.30cm){90}
\node[] at (1.5,-1.1) {\small Phase-5};
\node [wbg, minimum height =1.2cm, minimum width=1.2cm,rounded corners=0.2cm] at (1.5, -2.) {} ;
\leftturn(1.6,-1.7){270};
\straight(1.3, -1.7cm){270};

share
\end{tikzpicture}